\documentclass[10pt,onecolumn,letterpaper]{article}

\usepackage[top=1.5in,bottom=1.5in,left=1.2in,right=1.2in,columnsep=0.35in]{geometry}

\usepackage{indentfirst}
\usepackage{xcolor}
\usepackage{hyperref}
\usepackage{enumitem}
\usepackage{setspace}
\usepackage{algorithm}
\usepackage{algorithmic}
\usepackage{amsmath}
\usepackage{amsthm}
\usepackage{amssymb}
\usepackage{natbib}
\usepackage{stmaryrd}
\usepackage{hyperref}
\usepackage{tabularx}
\usepackage{booktabs} 
\usepackage{color}
\usepackage{graphicx}
\usepackage{caption}
\usepackage{subcaption}

\usepackage{authblk}

\newtheorem{lemma}{Lemma}[section]
\newtheorem{theorem}{Theorem}[section]
\newtheorem{corollary}{Corollary}[section]
\newtheorem{example}{Example}
\newtheorem{assumption}{Assumption}
\theoremstyle{definition}
\newtheorem{definition}{Definition}
\newtheorem{remark}{Remark}[section]

\newcommand{\vertiii}[1]{{\left\vert\kern-0.25ex\left\vert\kern-0.25ex\left\vert #1 
    \right\vert\kern-0.25ex\right\vert\kern-0.25ex\right\vert}}
\newcommand{\vertangl}[1]{{\langle\kern-0.3ex\langle #1 
    \rangle\kern-0.3ex\rangle}}

\author[1]{Wenjie Li}
\author[2]{Adarsh Barik}
\author[2]{Jean Honorio}
\affil[1]{Department of Statistics, Purdue University}
\affil[2]{Department of Computer Science, Purdue University}
\date{}

\begin{document}

\title{A Simple Unified Framework for High Dimensional Bandit Problems}
\maketitle

\begin{abstract}
Stochastic high dimensional bandit problems with low dimensional structures are useful in different applications such as online advertising and drug discovery. In this work, we propose a simple unified algorithm for such problems and present a general analysis framework for the regret upper bound of our algorithm. We show that under some mild unified assumptions, our algorithm can be applied to different high dimensional bandit problems. Our framework utilizes the low dimensional structure to guide the parameter estimation in the problem, therefore our algorithm achieves the comparable regret bounds in the LASSO bandit, as well as novel bounds in the low-rank matrix bandit, the group sparse matrix bandit, and in a new problem: the multi-agent LASSO bandit.
\end{abstract}

\section{Introduction}
\label{sec: introduction}

Stochastic multiarmed contextual bandits are useful models in various application domains, such as recommendation systems, online advertising, and personalized healthcare \citep{Auer2002Using, Chu2011Contextual, Abbasi-Yadkori2011Online}. Under this setting, the agent chooses one specific arm  at each round and observes a reward, which is modeled as a function of an unknown parameter and the context of the arm. In practice, such problems are often high-dimensional, but the unknown parameter is typically assumed to have low-dimensional structure, which in turns implies a succinct representation of the final reward.

For example, in high dimensional sparse linear bandits, also known as the LASSO bandit problem \citep{Bastani2015online}, both the contexts and the unknown parameter are high-dimensional vectors, while the parameter is assumed to be sparse with limited nonzero elements. There has been a line of research on LASSO bandits, such as \citet{Abbasi-Yadkori2011Online}, \citet{Carpentier2012Bandit}, \citet{Bastani2015online}, \citet{lattimore2015linear}, \citet{wang2018minimax}, \citet{Kim2019Doubly}, \citet{hao2020highdimensional}, and \citet{oh2020sparsityagnostic}. Different algorithms are proposed for this problem, and different regret analyses based on various assumptions are provided.

When the unknown parameter becomes a matrix, some recent research have worked on low-rank matrix bandits. For example, \citet{katariya2017bernoulli, katariya2017stochastic} and \citet{trinh2020solving} considered the rank-1 matrix bandit problems, but their results cannot be extended to higher ranks. \citet{jun2019bilinear} studied the bilinear bandit problems where the reward is modelled as a bilinear product of the left arm, the parameter matrix, and the right arm. \citet{kveton2017stochastic} first studied the low-rank matrix bandit, but required strong assumptions on the mean reward matrix. \citet{lu2020lowrank} extended the work by \citet{jun2019bilinear} to a more generalized low-rank matrix bandit problem but their action set was fixed. \citet{hao2020lowrank} further studied the problem of low-rank tensor bandits.

Despite the recent progress in all the above high dimensional bandit problems, both experimentally and theoretically, prior works are scattered and different algorithms with different assumptions are proposed for these problems. Although in LASSO Bandits, people have already given logarithmic dependence on the high dimension, we are surprised that such dependence has not been shown for other problems and polynomial dependence on the high dimensions still exists. An interesting question here is: does there exist a unified algorithm that works in all the high dimensional bandit problems, and if so, does such an algorithm hold a desirable regret bound in the different settings in terms of the dimensional dependence? In this work, we provide affirmative answers to these questions. 

Our work is inspired by the literature on traditional high dimensional statistical analysis \citep{Negahban2012A} and modern high dimensional bandit algorithms \citep{Kim2019Doubly, oh2020sparsityagnostic}. The only similar prior work we are aware of is the framework by \citet{johnson2016structured}, but they analyzed a completely different problem setting. Their algorithm is not only much more complicated than ours, but also requires very strong assumptions on the arm sets as well as low-dimensional structural information of the unknown parameter.

In particular, we want to highlight the following contributions of our paper:
{
\begin{itemize}
    \item We present a simple and unified algorithm framework named Explore-the-Structure-Then-Commit (ESTC) for high dimensional stochastic bandit problems and provide a problem-independent regret analysis framework for our algorithm. We show that to ensure a desirable regret, one simply needs to ensure that the two special events defined in Section \ref{sec: main_results} happen with high probability after the exploration stage under mild assumptions.
    \item We demonstrate the usefulness of our framework by applying it to different high dimensional bandit problems. We show that under a mild unified assumption, our algorithm can achieve desirable regret bounds. {We first prove that our algorithm achieves comparable regret bounds in the LASSO bandit problem as a sanity check; we then provide novel bounds in the low-rank matrix bandit and the group sparse matrix bandit which also depend logarithmicly on the dimensions. We also show that a simple extension of our algorithm can achieve group sparsity with a desirable regret bound in a new problem: the multi-agent LASSO bandit.} We summarize our results and compare them with the most relevant existing works in Table \ref{tab: regret_bounds}. 
    \footnote{For the first regret upper bound by \citet{hao2020highdimensional}, i.e. $\mathcal{O}(\tau s^{2/3}T^{2/3} \sqrt{{\log (dT)}})$, we have contacted the authors personally and confirmed that there is a mistake in the proof of Theorem 4.2 in their paper. The $\sqrt{{\log (dT)}}$ term is missing in the original theorem. Explanations are provided in Appendix \ref{sec: appendixMain}. }
    \item Our novel framework provides new proofs for different high dimensional bandit problems. Although prior works such as \citet{hao2020highdimensional} has proved similar results in the LASSO Bandit case, the proof techniques for low-rank matrix bandit and group-sparse matrix bandit are highly different and novel, which are also part of the contributions in this paper, e.g., proofs for Lemma \ref{lem: matrix_lambda_condition}, \ref{lem: group_sparse_lambda_condition}, \ref{lem: matrix_RSC_condition}, \ref{lem: matrix_RSC_difference}, and \ref{lem: matrix_sigma_distance}. Interested readers can refer to the Appendix for more details.
\end{itemize}
}

\begin{table*}
\centering
    \footnotesize
    \caption{\footnotesize Summary of regret bounds by different research works and their assumptions. \textbf{Parameters: }$d, (d_1, d_2)$ are the dimension sizes for vectors and matrices respectively; $K$ denotes the number of arms when the arm set is finite; $s$ denotes the support size of a vector; $r$ denotes the rank of a matrix; $\tau$ is a paper-dependent constant. \textbf{Arm Set:} ``fixed'' represents that the arm set is a fixed set $\mathcal{A}$; ``CBL'' represents that different arms have different parameters. ``finite'' represents that the number of arms is finite; ``contextual'' represents that $K$ different contexts are generated from a distribution at each round. $\widetilde{\mathcal{O}}(\cdot)$ hides the non-dominating terms. }
    \vspace{2pt}
\begin{tabular}{l  l l  l }
\hline
\sc{{LASSO}}  & \sc{Regret Bound} & \sc{Arm Set} & \sc{Assumptions} \\ \\[-1em]
 \hline
 \citet{lattimore2015linear} & $\mathcal{O}(s\sqrt{n})$ & fixed & action set is a hypercube  \\ \\[-1em]
 \citet{Bastani2015online} & $\mathcal{O}(\tau Ks^2\log^2 T)$ & CBL & compatibility, margin, optimality  \\ \\[-1em]
 \citet{wang2018minimax} & $\mathcal{O}(\tau K s^3 \log T)$ & CBL & compatibility, margin, optimality  \\ \\[-1em]
 \citet{Kim2019Doubly}  &  $\mathcal{O}(\tau s \sqrt{T} \log (dT))$  & contextual & compatibility  \\ \\[-1em]
 \citet{oh2020sparsityagnostic}& $\mathcal{O}(\tau s \sqrt{T \log (dT)})$ & contextual & compatibility, symmetry, covariance\\ \\[-1em]
 \citet{oh2020sparsityagnostic}  & $\mathcal{O}(\tau \sqrt{s T \log (dT)})$ & contextual & RE condition, symmetry, covariance  \\ \\[-1em]
\citet{hao2020highdimensional} & $\mathcal{O}(s^{2/3}T^{2/3} \sqrt{{\log (dT)}})$   & fixed & action set spans $\mathbb{R}^d$\\ \\[-1em]
 \citet{hao2020highdimensional}  & $\mathcal{O}(\sqrt{s T \log (KT)})$ & finite, fixed & action set spans $\mathbb{R}^d$, minimum signal \\ \\[-1em]
 \citet{hao2020highdimensional}  &  $\Omega(s^{1/3}T^{2/3})$ & finite, fixed & action set spans $\mathbb{R}^d$ \\ \\[-1em]
\textbf{This paper (sanity check)}  &   $\mathcal{O} (s^{1/3}T^{2/3}\sqrt{{\log (dT)}} ) $ & \textbf{contextual} & \textbf{RE condition} \\ \\[-1em]
\hline
\sc{Low-rank} & & &  \\ \\[-1em]
\hline
\citet{jun2019bilinear} & $\widetilde{\mathcal{O}}((d_1 + d_2)^{3/2}\sqrt{rT} )$ & fixed & the reward function is bilinear\\
\citet{johnson2016structured} &  $\widetilde{\mathcal{O}}((d_1 + d_2)^{3/2}\sqrt{rT} )$ & fixed  & prior knowledge of parameter \\ \\[-1em]
\citet{lu2020lowrank} &  $\widetilde{\mathcal{O}}((d_1 + d_2)^{3/2}\sqrt{rT} )$ & fixed & sub-Gaussian sampling distribution \\ \\[-1em]
\textbf{{This paper}} &  $\widetilde{\mathcal{O}}( r^{1/3} T^{2/3} \log ((d_1 + d_2))$   & \textbf{contextual} &\textbf{RE condition} \\ \\[-1em]
\hline
 \sc{Group-sparse} & & & \\ \\[-1em]
\hline
\citet{johnson2016structured}   &  $\mathcal{O}(\sqrt{s(d_2+\log K)d_1 T})$  & fixed  &  prior knowledge of parameter \\ \\[-1em]
\textbf{This paper}   &  $\widetilde{\mathcal{O}}(s^{1/3}\sqrt{d_2}T^{2/3} + s^{1/3}T^{2/3} \sqrt{\log d_1})$ & \textbf{contextual } & \textbf{RE condition} \\ \\[-1em]
\hline
\sc{Multi-agent} & & &  \\ \\[-1em]
\hline 
\textbf{This paper}   & ${\mathcal{O}(d_2 s^{1/3} T^{2/3}\sqrt{\log (d_1T)})}$  & \textbf{contextual} & \textbf{RE condition}\\
\hline
\end{tabular}
\label{tab: regret_bounds}
\end{table*}

\section{Preliminaries}
\label{sec: prelim}

In this section, we establish some important preliminary notations and definitions in this paper.

\textbf{Notations.} Given a subspace $\mathcal{M}$ of $\mathbb{R}^p$, its orthogonal complement ${\mathcal{M}}^{\bot}$ is defined as ${\mathcal{M}}^{\bot} := \{ v \in \mathbb{R}^p| \langle u, v\rangle = 0 \text{ for all } u \in {\mathcal{M}}\}$. The matrix inner product for two matrices $A, B$ of the same size are defined as $\vertangl{ A, B}:=\operatorname{trace}\left(A^{T} B\right) .$  We use $\|\cdot\|$ for vector norms and $\vertiii{\cdot}$ for matrix norms. Given a norm $\|\cdot\|$, we use the notation $\theta_{\mathcal{M}}$ to represent the projection of a vector $\theta$ onto $\mathcal{M}$, i.e., $\theta_{\mathcal{M}} = \text{argmin}_{\theta' \in \mathcal{M}}\|\theta - \theta'\|$, and similarly for $ \theta_{{\mathcal{M}}^{\bot}}$. The dual of a norm $\|\cdot\|$ is defined to be $\|u\|_* := \text{sup}_{\|v\| \leq 1} \langle u, v\rangle$. For the regularization norm $R$, we denote its dual to be $R^*$. We use $\widetilde{\mathcal{O}}(\cdot)$ to hide the non-dominating factors in big-$\mathcal{O}$ notations. We frequently use the notation $[T]$ for $T \in \mathbb{N}$ to denote the set of integers $\{1, 2, \cdots, T\}$. For the reader's reference, a complete list of all the norms and their duals used in this paper is provided in Appendix \ref{sec: appendixMain}.

\textbf{Multiarmed Contextual Bandits}. In modern multiarmed contextual bandit problems, a set of contexts $\{x_{t, a_i}\}_{i=1}^K$ for each arm is generated at every round $t$, and then the agent chooses an action $a_t$ from the $K$ arms. The contexts are assumed to be sampled i.i.d from a distribution $\mathcal{P}_{X}$ with respect to $t$, but the contexts for different arms can be correlated \citep{Chu2011Contextual}. After the action is selected, a reward 
$y_t=f(x_{t, a_t}, \theta^*)+ \epsilon_t$ for the chosen action is received, where $f$ is a deterministic function, $\theta^*$ is an unknown parameter, and $\epsilon_t$ is a zero-mean random noise term which is often assumed to be sub-Gaussian or even conditional sub-Gaussian in a few cases. In this paper, we focus on bandit problems where $\theta^*$ is high dimensional but with low-dimensional structure such as a sparse vector, a low-rank matrix, and a group-sparse matrix. 

Let $a_t^* = \text{argmax}_{i\in[K]} f(x_{t,a_i}, \theta^*)$ denote the optimal action at each round. We measure the performance of all algorithms by the expectation of the regret, denoted as
\begin{equation}
        \nonumber
        \mathbb{E}[\text{Regret} (T)]= \mathbb{E}\left[\sum_{t=1}^T f(x_{t,a_t^*}, \theta^*) -  f(x_{t,a_t}, \theta^*)\right]
\end{equation}
The goal for all the bandit algorithms studied in this paper is to ensure a sublinear expectation of the regret with respect to $T$, so that the average regret converges, thus making the chosen actions nearly optimal. Moreover, we require the regret bounds to depend on the low-dimensional structure constraint of the parameter, for example, the number of non-zero elements for a sparse vector,  rather than polynomially depend on the high dimension(s), so that the algorithm utilizes the low-dimensional structure of the parameter to reduce the regret.

\textbf{Optimization Problem}. We use the shorthand notations $\mathbf{X}_t = \{x_{i, a_i}\}_{i=1}^t$ and $\mathbf{Y}_t = \{y_i\}_{i=1}^t$ to represent all the contexts of the chosen actions  and the rewards received up to time $t$. Many algorithms designed for multiarmed bandit problems involve solving an online optimization problem with a loss function $L_t(\theta; \mathbf{X}_t, \mathbf{Y}_t)$ and a regularization norm $R(\theta)$, i.e.,
\begin{equation}
    \begin{aligned}
    \label{eqn: optimization_problem}
   \theta_t \in \text{argmin}_{\theta  \in \mathbf{\Theta} } \Big\{L_t(\theta; \mathbf{X}_t, \mathbf{Y}_t) + \lambda_t R(\theta) \Big\}
    \end{aligned}
\end{equation}
where $\lambda_t$ is the regularization parameter chosen differently in different algorithms and $\mathbf{\Theta}$ is the parameter domain. We also often use the notation $L_t(\theta)$ to represent $L_t(\theta; \mathbf{X}_t, \mathbf{Y}_t)$. The solution $\theta_t$ in Eqn. (\ref{eqn: optimization_problem}) is a regularized estimate of $\theta^*$ based on the currently available data. The reason to use the regularization $R(\theta)$ is that it can hopefully identify the low-dimensional structure of $\theta^*$ so that $\theta_t$ can converge to $\theta^*$ fast and the action chosen by the algorithm becomes optimal after a few rounds. 

{
\textbf{High Dimensional Statistics}. Our general theory applies to ${M}$-estimators, i.e., the loss has the form $L_t(\theta) = \frac{1}{t} \sum_{\tau=1}^t l_{\tau}(\theta)$, where each $l_{\tau}(\theta)$ is a convex and differentiable loss function and $R(\theta)$ is a decomposable norm. ${M}$-estimators and decomposable norms are well-accepted concepts in the high dimensional statistics literature \citep{Negahban2012A}. We first establish the definition of a decomposable norm.
}
\begin{definition}
\label{def: decomposible_R}
    Given a pair of subspaces $\mathcal{M} \subseteq \overline{\mathcal{M}}$, the norm $R$ is said to be decomposable with respect to the subspace pair $(\mathcal{M}, \overline{\mathcal{M}}^{\bot})$ iff
$
     R(\theta + \gamma) = R(\theta) + R(\gamma), \text{ for all } \theta \in \mathcal{M}, \gamma \in \overline{\mathcal{M}}^{\bot}
$
\end{definition} 

We present a few examples of the pair of subspaces and the decomposable regularization, each corresponding to one application of our general theory in Section \ref{sec: applications}.

\begin{example}
\label{example: sparse} \textbf{(Sparse Vectors and the $l_1$ Norm)}. Let $\theta^* \in \mathbb{R}^d$ be a sparse vector with $s \ll d$ nonzero entries, we denote $S(\theta^*)$ to be the set of non-zero indices of $\theta^*$ (i.e., the support). The pair of subspaces $\mathcal{M} \subseteq \overline{\mathcal{M}}$ are chosen as 
$
      \mathcal{M} =\overline{\mathcal{M}} = \{\theta \in \mathbb{R}^d \mid \theta_j = 0, \text{ for all } j \notin S(\theta^*)\}
$
.Then $l_1$ norm is decomposable with respect to $(\mathcal{M}, \overline{\mathcal{M}}^{\bot})$ i.e., $\|\theta + \gamma\|_1 = \|\theta\|_1 + \|\gamma\|_1$ for all $\theta \in \mathcal{M}, \gamma \in \overline{\mathcal{M}}^{\bot}$, since $\theta$ and $\gamma$ have non-zero elements on different entries.
\end{example}

\begin{example}
\label{example: low_rank} \textbf{(Low Rank Matrices and the Nuclear Norm)}. Let $\Theta^* \in \mathbb{R}^{d_1 \times d_2}$ be a low rank matrix with rank $r \ll \min \{d_1, d_2\}$. We define the pair of subspace $(\mathcal{M}, \overline{\mathcal{M}}^\bot)$ as 
\begin{equation}
        \begin{aligned}
        \nonumber
  \mathcal{M} &:=\{\Theta \in \mathbb{R}^{d_1 \times d_2} | \text{ row}(\Theta) \subseteq V, \text{ col} (\Theta) \subseteq U\} \\ 
  \overline{\mathcal{M}}^\bot &:=\{\Theta \in \mathbb{R}^{d_1 \times d_2} | \text{ row}(\Theta) \subseteq V^\bot, \text{ col} (\Theta) \subseteq U^\bot\}, 
    \end{aligned}
    \end{equation}
where $U$ and $V$ represent the space of the left and right singular vectors of the target matrix $\Theta^*$. Note that in this case $\mathcal{M} \subsetneq \overline{\mathcal{M}}$. Then the nuclear norm $\vertiii{\cdot}_{nuc}$ is decomposable with respect to $(\mathcal{M}, \overline{\mathcal{M}}^\bot)$, i.e., $\vertiii{\Theta + \Gamma}_{nuc} = \vertiii{\Theta}_{nuc} + \vertiii{\Gamma}_{nuc}$  for all $\Theta \in \mathcal{M}, \Gamma \in \overline{\mathcal{M}}^{\bot}$.
\end{example}

\begin{example}
\label{example: group_sparse} \textbf{(Group-sparse Matrices and the $l_{1,q}$ Norm)}. Let $\Theta^* \in \mathbb{R}^{d_1 \times d_2}$ be a matrix with group sparse rows, i.e., each row $\Theta^*_i$ is nonzero only if $i \in S(\Theta^*)$, and $|S(\Theta^*)| = s\ll d_1$. Similar to Example \ref{example: sparse}, we define the pair of subspace $(\mathcal{M}, \overline{\mathcal{M}}^\bot)$ as 
\begin{equation}
        \begin{aligned}
        \nonumber
      \mathcal{M} =\overline{\mathcal{M}} = \{\Theta \in \mathbb{R}^{d_1\times d_2} \mid \Theta_i = 0, \text{ for all } i \notin S(\Theta^*)\}
    \end{aligned}
    \end{equation}
The orthogonal complement can be defined with respect to the matrix inner product. Then the $l_{1,q}$ norm, defined as $\vertiii{\Theta}_{1,q} = \sum_{i=1}^{d_1}[\sum_{i=1}^{d_1} |\Theta_{ij}|^q]^{1/q}$, is decomposable with respect to $(\mathcal{M}, \overline{\mathcal{M}}^\bot)$.
\end{example}

Given the decomposable regularization $R$, it is shown in \citet{Negahban2012A} that when the regularization parameter $\lambda_t \geq 2R^*(\nabla L_t(\theta^*))$, the error $\theta_t - \theta^*$ belongs to a constraint set $\mathbb{C}$ defined as follows. 

{
\begin{definition}
\label{def: constraint_set}
    For any decomposable $R$ on the subspace pair $(\mathcal{M}, \overline{\mathcal{M}}^\bot)$, the constraint set $\mathbb{C}$ is defined as
\begin{equation}
    \begin{aligned}
    \nonumber
  \mathbb{C} := \Big\{ \Delta \mid  R(\Delta_{ \overline{\mathcal{M}}^\bot}) \leq 3R(\Delta_{\overline{\mathcal{M}}}) + 4R ( \theta^*_{\mathcal{M^{\bot}}}))\Big\}\\ 
    \end{aligned} 
\end{equation}
\end{definition} 
}

Such a set restrains the behavior of the error when the regularization parameter is correctly chosen. Next we present the definition of restricted strong convexity on the loss function $L_t(\theta)$.

\begin{definition}
\label{def: restricted_strong_convexity}
    The loss function $L_t(\theta)$ is said to be restricted strongly convex (RSC) around $\theta^*$ with respect to the norm $\|\cdot\|$ with curvature $\alpha > 0$ and tolerance function $Z_t(\theta^*)$ if
    \begin{equation}
        \begin{aligned}
        \nonumber
        B_t(\theta, \theta^*) &:= L_t(\theta) - L_t(\theta^*) - \langle \nabla L_t(\theta^*), \theta - \theta^* \rangle \\
        & \geq \alpha \Big\| \theta - \theta^* \Big\|^2 - Z_t(\theta^*)
    \end{aligned}
    \end{equation}
\end{definition} 

In the high dimensional statistics literature, restricted strong convexity is often ensured by a sufficient number of samples for some specific distributions of $\mathbf{X}_{t}$ such as the Gaussian distribution \citep{Negahban2012A}. In the online case, we need some special assumptions to guarantee that restricted strong convexity holds after a number of rounds. In addition, the following subspace compatibility constant plays a key role in restricting the distance between the true parameter and its estimate by the low dimensional structural constraint, hence generating a desirable regret bound.

\begin{definition}
\label{def: subspace_compatibility_constant}
    For a subspace $\overline{\mathcal{M}}$, the subspace compatibility constant with respect to the pair $(\|\cdot\|, R)$ is given by $\phi := \text{sup}_{u \in \mathcal{\overline{M}} \backslash \{0\}} \left( {R(u)}/{\|u\|} \right)$
\end{definition} 

For instance, $\phi = \sqrt{s}$ for the $(\|\cdot\|_2, \|\cdot\|_1)$ norm pair and $\mathcal{M}$ defined as in Example \ref{example: sparse} because $\sqrt{s} \|u\|_2 \geq \|u\|_1$ for $u \in \mathcal{M}$ by the Cauchy-Schwarz inequality.

\section{Main results}
\label{sec: main_results}

In this section, we present our simple and general algorithm, as well as its analysis framework.  Let $\alpha > 0$ be a constant to be specified later, we define the following two probability events  $\mathcal{A}_t$ and $\mathcal{E}_t$
\begin{equation}
    \begin{aligned}
    \nonumber
         &\mathcal{A}_t :=  \Big\{ \lambda_t \geq 2 R^*(\nabla L_t(\theta^*)) \Big\},\quad \\
         &\mathcal{E}_t :=  \Big\{ B_t(\theta, \theta^*) \geq \alpha \Big\| \theta - \theta^* \Big\|^2 - Z_t(\theta^*) \Big\}
    \end{aligned} 
\end{equation}
where $\mathcal{A}_t$ represents a correctly chosen regularization parameter and $\mathcal{E}_t$ means that $L_t(\theta)$ is RSC with respect to the undefined norm $\|\cdot\|$ with curvature $\alpha$ and tolerance $Z_t(\theta^*)$ at round $t$. As we will show below, ensuring that these two events happen with high probability is of vital importance.

\subsection{Oracle inequality}
We first present a general oracle inequality between the estimate and the true parameter, which is extended from the results in \citet{Negahban2012A}. 

\begin{lemma}
\label{lem: oracle_inequality}\textbf{\upshape (Oracle Inequality)} If at round $t$, the probability event $\mathcal{A}_t$ holds and the event $\mathcal{E}_t$ holds for all $\theta-\theta^* \in \mathbb{C}$, then for the solution $\theta_t$ of Eqn. \eqref{eqn: optimization_problem}, the difference $\theta_t - \theta^*$ satisfies the bound
\begin{equation}
\nonumber
    \Big\|\theta_t - \theta^*\Big\|^2 \leq 9 \dfrac{\lambda_t^2}{\alpha^2} \phi^2 + \dfrac{1}{\alpha}\Big[2Z_t(\theta^*) + 4 \lambda_t R(\theta^*_{\mathcal{M}^{\bot}}) \Big]
\end{equation}
\end{lemma}

{
\begin{remark}
The proof of {Lemma \ref{lem: oracle_inequality}} is provided in Appendix \ref{sec: appendixOracle} for completeness. {Lemma \ref{lem: oracle_inequality}} states an oracle inequality for each choice of the pair of norms $(\|\cdot\|, R)$, and the corresponding subspaces where $R$ is decomposable. The difference converges to zero if both $\lambda_t$ and $Z_t(\theta^*)$ decreases when $t$ increases, and therefore the estimate $\theta_t$ becomes more and more accurate as rounds progress.
\end{remark} 
}

\subsection{The general algorithm}

{
Now we present our Explore-the-Structure-Then-Commit (ESTC) algorithm for high dimensional contextual bandit problems. Our algorithm consists of two stages: the exploration stage where arms are randomly picked and the commit stage where the best arm explored is chosen. Algorithm \ref{alg: general} shows our procedure in detail. 
}
\begin{algorithm}[ht]
   \caption{Explore-the-Structure-Then-Commit}
   \label{alg: general}
\begin{algorithmic}[1]
   \STATE \textbf{Input:} $\lambda_{T_0}, K \in \mathbb{N}, L_t(\theta), R(\theta), f(x, \theta), \theta_0, T_0$
   \STATE Initialize $\mathbf{X}_0, \mathbf{Y}_0 = (\emptyset, \emptyset), \theta_t = \theta_0$
   \FOR{$ t= 1$ to $T_0$}
   \STATE Observe $K$ contexts, $x_{t,1}, x_{t,2}, \cdots, x_{t, K}$ 
    \STATE Choose action $a_t$ uniformly randomly 
    \STATE Receive reward $y_{t} = f(x_{t, a_t}, \theta^*) + \epsilon_{t}$
    \STATE $\mathbf{X}_{t} = \mathbf{X}_{t-1} \cup \{x_{t, a_t}\}, \mathbf{Y}_{t} = \mathbf{Y}_{t-1} \cup \{y_{a_t}\}$
        \ENDFOR
    \STATE Compute the estimator $\theta_{T_0}$: 
      \qquad $$\theta_{T_0} \in \text{argmin}_{\theta \in \mathbf{\Theta}} \left \{L_{T_0}(\theta; \mathbf{X}_{T_0}, \mathbf{Y}_{T_0}) + \lambda_{T_0} R(\theta) \right \}$$
      \vspace{-10pt}
    \FOR{$ t=  T_0 +1$ to $T$}
      \STATE Choose action $a_t = \text{argmax}_a f(x_{t,a}, \theta_{T_0})$ 
   \ENDFOR
\end{algorithmic}
\end{algorithm}

\begin{remark}

Similar algorithms have been proposed in some recent papers. For example, \citet{hao2020highdimensional} proposed an Explore-the-Sparsity-Then-Commit algorithm, but it was only designed and analyzed for fixed action sets and for LASSO bandits. \citet{oh2020sparsityagnostic} proposed a sparsity-agnostic algorithm in the LASSO bandit problem, which does not need the sparsity parameter $s$, but they required much more assumptions than ours in the multi-arm case $(K\geq 3)$. \citet{lu2020lowrank} proposed a Low-ESTR algorithm that solved the same optimization problem as our Section \ref{subsec: matrix_bandit} in the low-rank matrix bandit, but they require the lowOFUL algorithm by \citet{jun2019bilinear} and the algorithm is designed for fixed action sets. 
\end{remark}
{
\begin{remark}
Our algorithm generalizes over the prior efforts on different high dimensional bandit problems.
The advantages of our algorithm are that \textbf{ (1).} it is very simple,\textbf{ (2).} it does not require strong assumptions, and \textbf{(3). }it can be applied to different problems while it remains unclear whether prior works can be extended to all the problems we discuss here. Furthermore, we provide comparable regret bounds for existing problems and novel results for new problems (see Table \ref{tab: regret_bounds}, Section \ref{sec: applications}, and \ref{sec: discussions}).
\end{remark}
}

\subsection{Regret analysis}

To analyze the regret of Algorithm \ref{alg: general}, we impose some weak assumptions on the reward function and the norm of the context. The two assumptions are listed below for the general theorem, and we will demonstrate how they can be easily guaranteed in the specific applications in Section \ref{sec: applications}.

\begin{assumption}
\label{ass: normalization}
 $x$ is normalized with respect to the norm $\|\cdot\|$., i.e. $\|x\| \leq k_1$ for some constant $k_1$ .
\end{assumption}

\begin{assumption}
\label{ass: lipschitz}
$f(x, \theta)$ is $C_1$-Lipschitz over $x$ and $C_2$-Lipschitz over $\theta$ with respect to $\| \cdot \|$. i.e.,
\begin{equation}
    \begin{aligned}
    \nonumber
        &f(x_1, \theta) - f(x_2, \theta) \leq C_1 \Big\|x_1 - x_2\Big\|, \quad \\
        &f(x, \theta_1) - f(x, \theta_2) \leq C_2 \Big\|\theta_1 - \theta_2 \Big\|
    \end{aligned} 
\end{equation}
\end{assumption}

\begin{remark}
Assumption \ref{ass: normalization} is very standard in the contextual bandit literature, see e.g., \citet{Chu2011Contextual, Bastani2015online, Kim2019Doubly, lu2020lowrank}, where the $l_2$ norm or the Frobenius norm of the contexts is assumed to be either bounded by $1$ or a constant $x_{max}$. In the case where $f$ is normalization-invariant (e.g., the linear case), Assumption \ref{ass: normalization} can be achieved without loss of generality through normalization of the contexts and the rewards. Assumption \ref{ass: lipschitz} can also be easily guaranteed with the help of Assumption \ref{ass: normalization} for linear models, as well as for some generalized linear models whose link functions has bounded derivatives such as the logistic model. Given Assumptions \ref{ass: normalization} and \ref{ass: lipschitz}, we present our novel problem-independent theorem of the regret bound.
\end{remark}

\begin{theorem}
\label{thm: regret_bound}\textbf{\upshape(Problem Independent Regret Bound)}
Suppose that Assumption \ref{ass: normalization} and \ref{ass: lipschitz} hold. Then the expected cumulative regret of Algorithm \ref{alg: general} satisfies the bound
\begin{equation}
    \begin{aligned}
    \nonumber
        &\mathbb{E}[\text{Regret}(T) ] \leq 
        \underbrace{2C_1k_1T_0  \vphantom{\sum_{t = T_0}^T } }_{(a)} + 
        \underbrace{ 2C_1 k_1 \sum_{t = T_0}^T \left[  \mathbb{P} \left( \mathcal{A}_{T_0}^c \right) + \mathbb{P}(\mathcal{E}_{T_0}^c)\right]}_{(b)} \\
        & +\underbrace{ 2C_2 \sum_{t = T_0}^T \sqrt{9 \dfrac{\lambda_{T_0}^2}{\alpha^2} \phi^2 + \dfrac{1}{\alpha}[2Z_{T_0}(\theta^*) + 4 \lambda_{T_0} R(\theta^*_{\mathcal{M}^{\bot}})]} }_{(c)}
    \end{aligned} 
\end{equation}
\end{theorem}

\textbf{Proof Sketch}. Instead of assuming a particular form of the model $f$ or the shape of the parameter $\theta$, our proof techniques utilizes the property of all high dimensional bandit problems solved by Algorithm \ref{alg: general} and thus contributes to a general bound. As far as we are aware, no such general analysis have been proved before. The proof starts by showing the instant regret at round $t$ is bounded almost surely.
\begin{equation}
\nonumber
f(x_{t, a_t^*}, \theta^*) - f(x_{t, a_t}, \theta^*)  \leq 2C_1 k_1
\end{equation}
which can be used to bound the regret in $T_0$ and when $\mathcal{A}_{T_0}$ and $\mathcal{E}_{T_0}$ do not happen. Then notice that when $t\geq T_0$, our choice of action in the commit phase implies that $f(x_{t,a_t}, \theta_{T_0}) \geq  f(x_{t,a_t^*}, \theta_{T_0})$, we can therefore transform the problem of bounding the regret into a problem of bounding the distance between $\theta_{T_0}$ and $\theta^*$ by proving that under event $\mathcal{A}_{T_0}$ and $\mathcal{E}_{T_0}$, for $t\geq T_0$
\begin{equation}
    \begin{aligned}
    \nonumber
 f(x_{t,a_t^*}, \theta^*)- f(x_{t, a_t}, \theta^*) \leq  2 C_2 \|\theta_{T_0} - \theta^*\|  \\
    \end{aligned} 
\end{equation}
Therefore for any constant $v>0$, either
\begin{equation}
    \begin{aligned}
    \nonumber
 v\leq f(x_{t,a_t^*}, \theta^*)- f(x_{t, a_t}, \theta^*) \leq  2 C_2 \|\theta_{T_0} - \theta^*\|
    \end{aligned} 
\end{equation}
or 
$f(x_{t,a_t^*}, \theta^*)- f(x_{t, a_t}, \theta^*) \leq v$, which means that either the instantaneous regret is bounded by $v$, or the distance $\|\theta_{T_0} - \theta^*\|$ is large. The proof is finished by taking $v$ to be the upper bound in Lemma \ref{lem: oracle_inequality} multiplied by a constant, and taking expectations and summations over $t$. \hfill $\square$

\begin{remark}

The full proof is provided in Appendix \ref{sec: appendixMain}. The above regret bound may seem to be complicated at first sight, but we can interpret it in the following way.  In the initial $T_0$ exploration rounds, since we pull arms randomly to collect more samples,  we have to consider the worst case scenario and bound the regret linearly in $(a)$. After enough exploration at $T_0$,  when the two events $\mathcal{A}_{T_0}$ or $\mathcal{E}_{T_0}$ do not happen, no conclusions can be made on the distance between the estimator $\theta_{T_0}$ and the target $\theta^*$, contributing to the second term $(b)$. When both events happen, $\theta_{T_0}$ and $\theta^*$ are close enough and we can carefully bound the regret with the help of Lemma \ref{lem: oracle_inequality}, which generates the third term $(c)$. Theorem \ref{thm: regret_bound} indicates that the expected regret upper bound of {Algorithm \ref{alg: general}} depends on the probability of $\mathcal{A}_{T_0}$, $\mathcal{E}_{T_0}$ after a chosen round $T_0$, as well as the choice of $\lambda_{T_0}$ and $Z_{T_0}(\theta^*)$. Therefore, to obtain a sublinear regret, we only need to ensure that the following two things happen in the specific applications. 
\begin{enumerate}
    \item $\mathcal{A}_{T_0}, \mathcal{E}_{T_0}$ are high probability events.
    \item $\lambda_{T_0}, Z_{T_0}(\theta^*)$ are carefully chosen so that term $(c)$ is sublinear
\end{enumerate}

For instance, suppose that term (b) is finite, $\lambda_{T_0} = \mathcal{O}(T^{-1/3})$, and $Z_{T_0}(\theta^*) = R(\theta^*_{\mathcal{M}^{\bot}}) = 0$, then the expected regret is of size $\mathcal{O}({\phi} T^{2/3})$ by simple algebra. Such a result is desirable since the regret bound is sublinear with respect to $T$, and it depends on the subspace compatibility constant instead of the dimension size, so we utilize the low-dimensional structure in $\theta^*$. The final regret bound in Theorem \ref{thm: regret_bound} will become clearer when we discuss its specific applications. 
\end{remark}
\subsection{{High probability events}}

Finally, we address the two high probability events as they are needed to prove a final regret bound. As we will show in Section \ref{sec: applications}, the probability of the event $\mathcal{A}_t$ is often decided by the choice of $\lambda_t$ and the model structure. No further assumptions are needed for $\mathcal{A}_t$ to hold with high probability in all the problems in this paper. However, $\mathcal{E}_t$ does not necessarily hold with high probability even after a large number of rounds, and we need another assumption to guarantee its validity. 

{
\begin{assumption} \textbf{\upshape (Restricted Eigenvalue Condition)}
\label{ass: alpha_covariance}
Let $\mathbf{X}$ denote the matrix where each row is a context vector from an arm. The population Gram matrix $\Sigma = \frac{1}{K}\mathbb{E}[\mathbf{X}^T\mathbf{X}]$ satisfies that there exists some constant $\alpha_0 > 0$ such that $\beta^T \Sigma \beta \geq \alpha_0 \|\beta\|^2$, for all $\beta \in \mathbb{C}$.
\end{assumption}

\begin{remark}
In the case where the contexts are matrices, Assumption \ref{ass: alpha_covariance} represents that the population Gram matrix built from the vectorized contexts satisfies the aforementioned condition. To perform vetorization, we simply need to stack the columns of the matrices and then use the vectorized contexts to obtain the matrix $\mathbf{X}$. We emphasize that Assumption \ref{ass: alpha_covariance} is very general and mild because it is only a requirement on the population, which is satisfied for many distributions, for example, the uniform distribution on a Euclidean unit ball. Apart from Assumption \ref{ass: alpha_covariance}, prior works either need many more assumptions such as symmetric distribution and balanced co-variance in \citep{oh2020sparsityagnostic}, or very complicated algorithms to make $\mathcal{E}_t$ a high probability event \citep{Kim2019Doubly}. Our algorithm, however, is simple and general and it can be applied to different problems with only Assumption \ref{ass: alpha_covariance}.
\end{remark}

\begin{remark}

Another popular assumption in the LASSO bandit problem is the compatibility condition \citep{Bastani2015online, Kim2019Doubly}, which would replace the norm in Assumption \ref{ass: alpha_covariance} by the $l_1$ norm (the regularization norm $R$ in the LASSO bandit problem). Although the compatibility condition is slightly weaker than the restricted eigenvalue condition, we can easily replace the norms in all our arguments by the $l_1$ norm and still obtain the favorable properties similar to Lemma \ref{lem: oracle_inequality}. The final regret bound may slightly differ from our results in Section \ref{subsec: lasso_bandit}  in terms of multiplicative constants if we use the compatibility condition, but the proof idea is the same. We use the RE condition in our paper because it is currently unknown whether the compatibility condition can be extended to the matrix bandit case. We leave it as a future work direction.
\end{remark}

}

\section{Applications on existing problems}
\label{sec: applications}

In this section, we present some specific applications of our general framework. Each subsection is organized in the following way. We first clarify all the unspecified notations in the framework, such as the loss, the regularization norm, the compatibility constant, and so on. Then we present a lemma to give the proper choice of the regularization parameter $\lambda_t$ in each problem. Given the lemma, we derive corollaries of Theorem \ref{thm: regret_bound} to present the final regret bound of the corresponding algorithm. We emphasize that even though we focus on linear models in all examples for clarity, it is easy to extend our results to nonlinear models that satisfy Assumption \ref{ass: normalization} and \ref{ass: lipschitz}. For example, results for generalized linear models whose link functions are Lipschitz can be easily obtained. {Validation experiments that support the correctness of our regret upper bounds are provided in Appendix \ref{sec: appendixExp} }

\subsection{LASSO bandit}
\label{subsec: lasso_bandit}
We first consider the LASSO bandit problem. In this case, the reward is assumed to be a linear function of the context of the chosen action $x_{t, a_t} \in \mathbb{R}^d$ and the unknown parameter $\theta^* \in \mathbb{R}^d$ , i.e., $y_t = \langle x_{t, a_t}, \theta^* \rangle + \epsilon_t$, where $\epsilon_t$ is a (conditional) sub-Gaussian noise. In the LASSO bandit problem, the unknown parameter $\theta^*$ is assumed to be sparse with only $s$ non-zero elements and $s \ll d$. This naturally leads to the use of $l_1$ regularization.

To fit the problem into our framework so that we can get a regret bound, we first clarify the corresponding notations. The loss function and the regularization are defined to be
\begin{equation}
    \nonumber
    L_t(\theta) = \frac{1}{2t} \sum_{i=1}^t (y_i - x_{i, a_i}^T \theta)^2, \quad R(\theta) = \|\theta\|_1
\end{equation}
In this case, we let $(\mathcal{M}, \overline{\mathcal{M}}^\bot)$ be defined as in Example \ref{example: sparse}, then $l_1$ regularization is decomposable with respect to this pair of spaces. We choose the norm $\|\cdot\|$ in $ \mathcal{E}_t$ as the $l_2$ norm, and thus the compatibility constant $\phi = \sqrt{s}$. Following \citet{Chu2011Contextual}, we assume that the contexts $x_{t, a_i}$ and the parameter $\theta^*$ are all normalized so that $\|x_{t, a_i}\|_2 \leq 1$, $\|\theta^*\|_2 \leq 1$. Therefore Assumption \ref{ass: normalization} and \ref{ass: lipschitz} are satisfied automatically. Now, we present the following lemma, which provides the choice of $\lambda_t$ in the LASSO Bandit problem. 

\begin{lemma}
\label{lem: lasso_lambda_condition}
Suppose the noise $\epsilon_t$ is conditional $\sigma$-sub-Gaussian. For any $\delta \in (0,1)$, use 
\begin{equation}
\nonumber
    \lambda_{T_0} =  \frac{2\sigma}{\sqrt{{T_0}}}\sqrt{{2 \log (2d/\delta)}}
\end{equation} in Algorithm \ref{alg: general}, then with probability at least $1-\delta$, we have $\lambda_{T_0} \geq R^*(\nabla L_{T_0}(\theta^*))$
\end{lemma}

Given the above lemma, it is easy to apply Theorem \ref{thm: regret_bound} to get a specific regret bound in the LASSO bandit. Corollary \ref{cor: lasso_regret} follows from taking $\delta = 1/T^2 $ in Lemma \ref{lem: lasso_lambda_condition} and $T_0 = {\Theta}(s^{1/3}T^{2/3})$. 

{
\begin{corollary}
\label{cor: lasso_regret}
The expected cumulative regret of the Algorithm \ref{alg: general} in the LASSO bandit problem is upper bounded by \begin{equation}
\nonumber
    \mathbb{E}[ \text{Regret}(T) ] = \mathcal{O} \left(s^{1/3}T^{2/3}\sqrt{{\log (dT)}} \right) 
\end{equation}
\end{corollary}
}

\begin{remark}
The proofs and the specific algorithm are provided in Appendix \ref{sec: appendixLASSO}. Note that most of the regret bound depends polynomially on the size of the support $s$ and only logarithmicly on the high dimension $d$. Therefore, Algorithm \ref{alg: general} converges much faster than directly applying linear bandit algorithms such as LinUCB \citep{Chu2011Contextual} to the sparse setting, which satisfies our requirement.  Corollary \ref{cor: lasso_regret} indicates that the regret bound is of size $\widetilde{\mathcal{O}}(s^{1/3}T^{2/3})$, which matches the regret upper bound proved by \citet{hao2020highdimensional} on the fixed action set and the minimax lower bound up to a logarithmic factor \citep{hao2020highdimensional}. 
\end{remark}

{

\begin{remark}
Some recent papers have proved better dependence in terms of the total number of rounds $T$ with more complexly-designed algorithms, e.g. \citet{Kim2019Doubly} or more assumptions, e.g. \citet{oh2020sparsityagnostic}. Specifically, they have provided $\widetilde{\mathcal{O}}(\sqrt{T})$ regret bounds. However, as mentioned by \citet{hao2020highdimensional}, these works all need $T$ to be sufficiently large as an additional requirement. In the so-called data-poor regime where $T$ is limited, an $\Omega(T^{2/3})$ regret is unavoidable and thus our regret is optimal \citep{hao2020highdimensional}. Moreover, since vectors are 1-column matrices, such $\Omega(T^{2/3})$ lower bounds are also unavoidable in the low-rank and group-sparse matrix bandit problems. 
Our result on LASSO bandit serves as a sanity check of the logarithmic dependence on the dimension and our main goal is to show such dependence in the other problems. 
\end{remark}
}

\subsection{Low-Rank matrix bandit}
\label{subsec: matrix_bandit}

Next, we consider the implications of our general theory in the low-rank matrix bandit problems. In this case, the reward is assumed to be a linear function of the low rank matrix $\Theta^* \in \mathbb{R}^{d_1 \times d_2}$ and the context matrix $X_{i, a_i} \in \mathbb{R}^{d_1 \times d_2}$, i.e., $Y_{t}=\vertangl{X_{t, a_t}, \Theta^{*}}+\epsilon_{t}$, where $\text{rank}(\Theta^*) = r \ll \min\{d_1, d_2\}$. We specify the loss function and the regularization norm to be
\begin{equation}
    \nonumber
    L_t(\Theta) =  \frac{1}{2t} \sum_{i=1}^{t}(y_{i}- \vertangl{X_{i, a_i}, \Theta} )^{2}, \quad R(\Theta) = \vertiii{\Theta}_{nuc}
\end{equation}
By Example \ref{example: low_rank}, the nuclear norm regularization is decomposable on $(\mathcal{M}, \overline{\mathcal{M}}^\bot)$ defined by the left and right singular vector spaces of $\Theta^*$. Note that in this case $\phi = \sqrt{2r}$ because the space $\overline{\mathcal{M}}$ contains matrices with rank at most $2r$. The norm we choose in the event $\mathcal{E}_t$ is the Frobenious norm. Also, we assume without loss of generality that the matrices are normalized, so that $\vertiii{X_{t, a_i}}_F \leq 1$, $\vertiii{\Theta^*}_F \leq 1$. Therefore, Assumption \ref{ass: normalization} and \ref{ass: lipschitz} are automatically satisfied. Similar to {Lemma \ref{lem: lasso_lambda_condition}}, the following lemma provides a good choice of $\lambda_t$ for the low-rank matrix bandit.

\begin{lemma}
\label{lem: matrix_lambda_condition}
 Suppose the noise $\epsilon_t$ is $\sigma$-sub-Gaussian.  At each round $t$, for any $\delta \in (0,1)$, use $$\lambda_{T_0} =  \frac{4\sigma}{\sqrt{{T_0}}} \sqrt{{3\log(2{T_0}/\delta) \log^2(2(d_1 + d_2)/\delta)}}$$ in Algorithm \ref{alg: general}, then with probability at least $1-\delta$, we have $\lambda_{T_0} \geq R^*(\nabla L_{T_0}(\Theta^*))$
\end{lemma}

Similarly, we can obtain the regret bound by setting $\delta = 1/T^2$ in Lemma \ref{lem: matrix_lambda_condition} and $T_0 = \Theta(r^{1/3} T^{2/3})$.

{

\begin{corollary}
\label{cor: matrix_regret}
The expected cumulative regret of the Algorithm \ref{alg: general} in the low-rank matrix bandit problem is upper bounded by 
\begin{equation}
\nonumber
    \mathbb{E}[ \text{Regret}(T) ] = \mathcal{O} \left ( r^{1/3} T^{2/3}\sqrt{\log(T)} \log((d_1+d_2)T) \right)
\end{equation}
\end{corollary}
}

{
\begin{remark}
\label{remark: low-rank}
The proofs and the corresponding matrix bandit algorithm are provided in Appendix \ref{sec: appendixMatrix}. The regret bound in Corollary \ref{cor: matrix_regret} is of size $\widetilde{\mathcal{O}}(r^{1/3} T^{2/3} \log((d_1+d_2)) )$, which polynomially depends on the low rank structure but only logarithmicly on the dimensions and is thus better than directly applying linear bandit algorithms to the problem, which is again desirable. We emphasize that our result cannot be directly compared with some prior results, because papers such as \citet{lu2020lowrank, jun2019bilinear} consider a fixed set of the context matrices $X_{t,a_t}$ instead of a generating distribution $P_\mathcal{X}$, which is different from our setting. Although our dependence on the total number of rounds $T$ is $\widetilde{\mathcal{O}}(T^{2/3})$, we have logarithmic dependence on the sum of the dimensions  $(d_1 + d_2)$ and thus it is more desirable than bounds that have polynomial dependence on $d$ when the dimensions $d_1, d_2$ are large.
\end{remark}
}

\subsection{Group-sparse matrix bandit}
\label{subsec: group_sparse_bandit}

Finally, we also apply our general framework to the group-sparse matrix bandit problem. Similar to the low-rank matrix bandit case, we assume that the expected reward is a linear function of the context matrix $X_{t, a_t}  \in \mathbb{R}^{d_1 \times d_2} $ and the optimal parameter $\Theta^*  \in \mathbb{R}^{d_1 \times d_2} $, i.e., $Y_{t}=\vertangl{X_{t, a_t}, \Theta^{*}}+\epsilon_{t}$. Define $S(\Theta^*) = \{i \in [d_1] \mid \Theta^{*}_i = 0\}$ to be the set of non-zero rows in $\Theta^*$ and assume that $|S(\Theta^*)| = s \ll d_1$, so the matrix is group sparse. We again specify the sum of squared errors to be the loss, but use the $l_{1,q}$ norm as the regularization norm.
\begin{equation}
    \nonumber
    L_t(\Theta) =  \frac{1}{2t} \sum_{i=1}^{t}(y_{i}- \vertangl{X_{i, a_i}, \Theta} )^{2}, \quad R(\Theta) = \vertiii{\Theta}_{1,q}
\end{equation}
Note that if $q =1$, the above problem is equivalent to the LASSO bandit case, because we only need to vectorize all the matrices by stacking their columns. Then $L_t(\Theta)$ becomes the mean squared error and $R(\Theta)$ becomes the $l_1$ regularization, thus we only consider $q > 1$ here. Define the subspaces using $S(\Theta^*)$ as in Example \ref{example: group_sparse}, then the $l_{1,q}$ norm regularization is decomposable.

Similarly, we assume without loss of generality that the matrices are normalized with respect to the Frobenious norm, so that Assumption \ref{ass: normalization} and \ref{ass: lipschitz} are satisfied automatically. We define the function $\eta(d_2, m) = \max\{1,  d_2^{m}\}$. This notation is used because we need slightly different choices of $\lambda_t$ for $q \in (1,2]$ and $q > 2$. Also, the compatibility constant $\phi = \eta(d_2, \frac{1}{q} - \frac{1}{2})\sqrt{s}$, which varies for different $q$. Similarly, we provide the following lemma for a good choice of the parameter $\lambda_t$ in group-sparse matrix bandits.

\begin{lemma}
\label{lem: group_sparse_lambda_condition}
Suppose that $\epsilon_t$ is $\sigma$-sub-Gaussian. At each round $t$, for any $\delta \in (0,1)$, use
$$\lambda_{T_0} =   \frac{2 \sigma}{\sqrt{{T_0}}} {d_2^{1-{1}/{q}}} + \frac{\sigma \eta(d_2, \frac{1}{2} - \frac{1}{q})}{\sqrt{{T_0}}} \sqrt{2 \log ({2d_1}/{\delta})}$$
 in Algorithm \ref{alg: general}, then with probability at least $1-\delta$, we have $\lambda_{T_0} \geq R^*(\nabla L_{T_0}(\Theta^*))$.
\end{lemma}

Given the above lemma, we can similarly derive a desirable regret bound.

{
\begin{corollary}
\label{cor: group_sparse_regret}
With the two functions of $d_2$
\begin{equation}
\begin{aligned}
\nonumber
    &C_1 (d_2) = d_2^{1-\frac{1}{q}} {  \eta(d_2, {\frac{1}{q} - \frac{1}{2}})}, \\ 
    &C_2 (d_2) = {\eta(d_2, \frac{1}{q} - \frac{1}{2})  \eta(d_2, \frac{1}{2} - \frac{1}{q})} 
    \end{aligned}
\end{equation} the expected cumulative regret of the Algorithm \ref{alg: general} in the group-sparse matrix bandit problem is upper bounded by 
\begin{equation}
\begin{aligned}
\nonumber
    &\mathbb{E}[ \text{Regret}(T)] \\
    &= \mathcal{O} \left(C_1(d_2) s^{1/3} T^{2/3} + C_2(d_2) s^{1/3} T^{2/3} \sqrt{\log (d_1 T)} \right)
\end{aligned}
\end{equation} 
\end{corollary}
}

{
\begin{remark}
\label{remark: group-sparse}
The proofs and the algorithm are provided in Appendix \ref{sec: appendixGroupSparse}. The regret bound in Corollary \ref{cor: group_sparse_regret} depends on the choice of the regularization norm. For example, if $q=2$, then $R(\Theta)$ is the group-LASSO regularization and the final regret bound is $\mathcal{O}(\sqrt{d_2}s^{1/3} T^{2/3} + s^{1/3} T^{2/3}\sqrt{\log(d_1T)})$. Besides, the regret bound only has logarithmic dependence on $d_1$, so it is  desirable since we assume that the matrix is group sparse with $s \ll d_1$. Again, we emphasize that the regret bound is not directly comparable to \citet{johnson2016structured} since they consider a fixed action set. Although we have $\widetilde{\mathcal{O}}(T^{2/3})$ dependence on $T$, our regret bound is more desirable when the dimension $d_1$ is high and our logarithmic dependence on $d_1$ is preferred.
\end{remark}
}

\section{Application on a new problem: the multi-agent LASSO bandit}
\label{sec: discussions}

Apart from direct applications, our general analysis framework can also inspire less-direct use cases. Here we briefly present a novel application where our framework guides a new algorithm and a new regret bound.

Suppose there are $d_2$ agents solving LASSO bandit problems synchronously, and each agent $k$ receives a stochastic reward $y_{t}^{(k)} = x_{t, a_t}^{(k)T} \theta^{(k)*} + \epsilon_t^{(k)}$ at every round $t$. Here, the contexts for each problem are similar and thus the parameters are required to be group-sparse instead of just sparse. The major challenges in this problem are that, first, there are multiple actions and multiple rewards at each round. Second, the Lipschitzness assumption may only be guaranteed for each agent, but not for the whole problem. Therefore, we propose a variant of our algorithm (shown in Algorithm \ref{alg: multi-agent} in Appendix \ref{sec: appendixMultiAgent} where the agents jointly solve the following optimization problem.
\begin{equation}
\nonumber
    \text{argmin}\{\frac{1}{2T_0} \sum_{i=1}^{T_0} \sum_{k=1}^{d_2} (y_{i}^{(k)}- x_{i,a_i}^{(k)T}\theta^{(k)*})^2+ \lambda_{T_0} \vertiii{\Theta}_{1,2} \}
\end{equation}

Note that the above problem also has the form of Eqn. (\ref{eqn: optimization_problem}). Therefore, following the same logic in our analysis framework, we prove two high probability events that correspond to the correctly chosen regularization parameter and the restricted strong convexity in this multi-agent problem, and then an $\mathcal{O}(d_2 s^{1/3}T^{2/3} \sqrt{\log (d_1T)})$ regret bound is proved in Appendix \ref{sec: appendixMultiAgent}.
\begin{theorem}
\label{thm: regret_bound_multi_agent}
Suppose that Assumption \ref{ass: normalization} and Assumption \ref{ass: lipschitz} hold for each agent (with respect to the $\|\cdot\|_2$ norm). Also suppose that Assumption \ref{ass: alpha_covariance} is satisfied. Then the expected cumulative regret of Algorithm \ref{alg: multi-agent} is upper bounded by 
\begin{equation}
    \begin{aligned}
    \nonumber
        \mathbb{E}[\text{Regret}(T) ]
        & = \mathcal{O}\left( d_2  s^{1/3} T^{2/3} \sqrt{\log 2d_1T}\right)
    \end{aligned} 
\end{equation}
\end{theorem}
\begin{remark}
Although the bound in Theorem \ref{thm: regret_bound_multi_agent} is the same as applying the LASSO bandit algorithm to each agent independently, our regularization can ensure group-sparsity in the parameter, which cannot be guaranteed in the former case. Again, such a regret bound is desirable since we only have logarithmic dependence on $d_1$, and $s\ll d_1$.
\end{remark}

\bibliography{my_bib}
\bibliographystyle{dinat}

\clearpage
\onecolumn
\appendix

\onecolumn
\appendix

\def\cpar{\hss\egroup\line\bgroup\hss}

\centerline{\textbf{\LARGE Supplementary Materials to "A Simple Unified }}
\vspace{5pt}
\centerline{\textbf{\LARGE  Framework for High Dimensional Bandit Problems"}}

\section{Notations and the Main Theorem}
\label{sec: appendixMain}

\subsection{List of Norms}
For a vector $v \in \mathbb{R}^d$, 
\begin{itemize}
    \item $\|v\|_2$ is the $l_2$ norm of $v$, i.e., $\|v\|_2 = (v_1^2 + v_2^2 + \cdots + v_d^2)^{1/2}$, which is self-dual.
    \item $\|v\|_1$ is the $l_1$ norm of $v$, i.e., $\|v\|_1 = |v_1| + |v_2| + \cdots + |v_d|$, whose dual norm is the $l_\infty$ norm.
    \item $\|v\|_\infty$ is the $l_\infty$ norm of $v$, i.e., $\|v\|_{\infty} = \max_{i=1,2,\cdots d}|v_i|$, whose dual norm is the $l_1$ norm.
\end{itemize}

For a matrix $M \in \mathbb{R}^{d_1 \times d_2}$ 
\begin{itemize}
    \item $\vertiii{M}_F$ is the Frobenius norm of $M$, i.e., $\vertiii{M}_F = (\sum_{i=1}^{d_1} \sum_{j=1}^{d_2} M_{ij}^2)^{1/2}$, which is self-dual.
    \item $\vertiii{M}_{nuc}$ is the nuclear norm of $M$, which is defined as $\vertiii{M}_{nuc} = \sum_{i=1}^{\min \{d_1,d_2\}} \sigma_i$, where $\{\sigma_i\}$'s are the singular values of $M$. The dual norm of the nuclear norm is the $l_2$ induced operator norm.
    \item $\vertiii{M}_{op}$ is the operator norm of $M$ induced by the vector norm $\|\cdot\|_2$, i.e., $\vertiii{M}_{op} = \sup_{\|x\|_2 =1} \|Mx\|_2$. An important property of $\vertiii{\cdot}_{op}$ is that $\vertiii{M}_{op} = \max_{i=1}^{\min \{d_1,d_2\}} \sigma_i$, where $\{\sigma_i\}$'s are the singular values of $M$. The dual norm of the induced operator norm is the nuclear norm.
    \item $\vertiii{M}_{1,q} = \sum_{i=1}^{d_1}[\sum_{j=1}^{d_2} |M_{ij}|^q]^{1/q}$ is the $l_{1,q}$ norm of $M$. For example if $q = 2$, this corresponds to the group lasso regularization. The dual norm of $l_{1,q}$ norm is the $l_{\infty,p}$ norm, defined as $\vertiii{M}_{\infty,p} = \max_{i=1}^{d_1}[\sum_{j=1}^{d_2} |M_{ij}|^p]^{1/p}$, with the relationship $1/p + 1/q = 1$.
    \item $\vertiii{M}_{max}$ is the element-wise maximum norm of $M$, which is defined as $\vertiii{M}_{max} = \max_{i=1}^{d_1} \max_{j=1}^{d_2} |M_{ij}|$.
\end{itemize}

\subsection{Explanations for the Regret Bound by \citet{hao2020highdimensional}}

For the results in this subsection, we have personally contacted the authors and confirmed that the logarithmic term is missing in their proof. For clarity, we follow their notations where they have used $n$ instead of $T$ for the total number of rounds and $n_1$ instead of $T_0$ for the exploration rounds. The other notations are either the same as ours (e.g. sparsity level of the target vector is denoted by $s$, dimension size is denoted by $d$) or just constants (e.g., $R_{\max}$ is the maximum reward, which is 1 in our case). Please refer to the original paper for more detailed notations. Now in their ``Section B.3: Proof of Theorem 4.2: regret upper bound'', on page 17, the authors have proved that the regret is upper bounded by the following terms in ``Step 3: optimize the length of exploration''. 

With probability at least $1-\delta$. 
\begin{equation}
\begin{aligned}
R_{\theta}(n)
& \leq n_{1} R_{\max }+\left(n-n_{1}\right) \frac{4}{C_{\min }} \sqrt{\frac{2 s^{2}\left(\log (2 d)+\log \left(n_{1}\right)\right)}{n_{1}}} +3 n R_{\max } \exp \left(-c_{1} n_{1}\right)
\end{aligned}
\end{equation}

In the final step, \citet{hao2020highdimensional} have taken $n_1 = n^{2/3} (s^2 \log(2d))^{1/3} R_{\max}^{-2/3} (2/C_{\min}^2)^{1/3} = {\Theta}(n^{2/3})$ and compute the final regret bound. However, they have ignored the $\log(n_1)$ term on the numerator of the fractional under the square root sign. In other words, when taking $n_1 = n^{2/3} (s^2 \log(2d))^{1/3} R_{\max}^{-2/3} (2/C_{\min}^2)^{1/3} = {\Theta}(n^{2/3})$, we get

\begin{equation}
\begin{aligned}
R_{\theta}(n)
& \leq n^{2/3} (s^2 \log(2d))^{1/3} (2/C_{\min}^2)^{1/3} R_{\max}^{1/3}+ n^{2/3} s^{2/3} R_{\max}^{1/3} \frac{4 \cdot 2^{1/3}}{C_{\min}^{2/3}} \sqrt{\frac{\left(\log (2 d)+ \frac{2}{3}\log \left(n\right) + \alpha\right)}{(\log(2d))^{1/3}}} \\
&\qquad +3 n R_{\max } \exp \left(-c_{1} n_{1}\right) \\
&= \mathcal{O}\left(s^{2/3} n^{2/3} \sqrt{\log (dn)} \right)
\end{aligned}
\end{equation}

where $\alpha = \log \left((s^2 \log(2d))^{1/3} R_{\max}^{-2/3} (2/C_{\min}^2)^{1/3} \right)$ is a constant. Therefore, the regret bound in \citet{hao2020highdimensional} has the same order as our regret bound in Corollary \ref{cor: lasso_regret}.

\subsection{Proof of Theorem \ref{thm: regret_bound}}
\textbf{Proof}. By the Lipschitzness of $f$ over $x$ with respect to $\|\cdot\|$, and the boundedness of $\|x\|$, we have

\begin{equation}
    \begin{aligned}
    \nonumber
        f(x_{t, a_t^*}, \theta^*) - f(x_{t, a_t}, \theta^*) 
                      \leq C_1 \|x_{t, a_t^*} - x_{t, a_t}\| \ \leq 2C_1 k_1
    \end{aligned} 
\end{equation}

Then we can decompose the one-step regret from round $t$ into three parts as follows,

\begin{equation}
    \begin{aligned}
    \nonumber
        & f(x_{t, a_t^*}, \theta^*) - f(x_{t, a_t}, \theta^*) \\
                      &=  \left [f(x_{t, a_t^*}, \theta^*) - f(x_{t, a_t}, \theta^*)  \right] \mathbb{I}(t \leq T_0) + \left [f(x_{t, a_t^*}, \theta^*) - f(x_{t, a_t}, \theta^*)   \right] \mathbb{I}(t > T_0, \mathcal{E}_{T_0}) \\
                     & \qquad  + \left [f(x_{t, a_t^*}, \theta^*) - f(x_{t, a_t}, \theta^*)  \right] \mathbb{I}(t > T_0, \mathcal{E}_{T_0}^c) \\
                     &\leq 2C_1 k_1 \mathbb{I}(t \leq T_0) +   \left [f(x_{t, a_t^*}, \theta^*) - f(x_{t, a_t}, \theta^*)  \right]  \mathbb{I}(t >  T_0, \mathcal{E}_{T_0}) + 2C_1 k_1\mathbb{I}(t > T_0, \mathcal{E}_{T_0}^c) \\
                    &= 2C_1 k_1 \mathbb{I}(t \leq T_0) +   \left [f(x_{t, a_t^*}, \theta^*) - f(x_{t, a_t}, \theta^*)  \right]  \mathbb{I}(t >  T_0, f(x_{t,a_t}, \theta_{T_0}) \geq f(x_{t, a_t^*}, \theta_{T_0}), \mathcal{E}_{T_0}) + 2C_1 k_1\mathbb{I}(t > T_0, \mathcal{E}_{T_0}^c)
    \end{aligned} 
\end{equation}

where $\mathbb{I}(\cdot)$ is the indicator function. The last equality is due to the choice of $a_t =  \text{argmax}_a f(x_{t,a}, \theta_{T_0})$, and thus we know that $f(x_{t,a_t}, \theta_{T_0}) \geq  f(x_{t,a_t^*}, \theta_{T_0})$. We focus on the second indicator function now. By the Lipschitzness of $f$ over $\theta$, we have

\begin{equation}
    \begin{aligned}
    \nonumber
        &\mathbb{I}\left(t >  T_0, \mathcal{E}_{T_0}\right)  \\
        &= \mathbb{I}\left(t >  T_0, f(x_{t,a_t}, \theta_{T_0}) \geq f(x_{t,a_t^*}, \theta_{T_0}), \mathcal{E}_{T_0}\right)  \\
        &= \mathbb{I}\left(t >  T_0, f(x_{t,a_t}, \theta_{T_0})   - f(x_{t,a_t^*}, \theta_{T_0}) + f(x_{t,a_t^*}, \theta^*) - f(x_{t,a_t}, \theta^*) \geq   f(x_{t,a_t^*}, \theta^*) - f(x_{t,a_t}, \theta^*), \mathcal{E}_{T_0}\right) \\
        &= \mathbb{I}\left(t >  T_0, \left [f(x_{t,a_t}, \theta_{T_0})   - f(x_{t,a_t}, \theta^*) \right] + \left[f(x_{t,a_t^*}, \theta^*) - f(x_{t,a_t^*}, \theta_{T_0})\right] \geq   f(x_{t,a_t^*}, \theta^*) - f(x_{t,a_t}, \theta^*), \mathcal{E}_{T_0}\right) \\
        & \leq \mathbb{I}\left(t >  T_0, 2 C_2 \|\theta_{T_0} - \theta^*\| \geq   f(x_{t,a_t^*}, \theta^*)- f(x_{t,a_t}, \theta^*), \mathcal{E}_{T_0} \right) \\
    \end{aligned} 
\end{equation}

Substitute the above inequality back and take expectation on both sides of the one-step regret from round $t$, we get

\begin{equation}
    \begin{aligned}
    \nonumber
        \mathbb{E}\left[f(x_{t, a_t^*}, \theta^*) - f(x_{t, a_t}, \theta^*) \right] \leq 2C_1 k_1 \text{ \hfill  for }t \leq T_0
    \end{aligned} 
\end{equation}

For $t > T_0$ and any constant $v \in \mathbb{R}$, the expectation is bounded by

\begin{equation}
    \begin{aligned}
    \nonumber
        &\mathbb{E}\left[f(x_{t, a_t^*}, \theta^*) - f(x_{t, a_t}, \theta^*) \right] \\
        &\leq \mathbb{E} \left [ \left(f(x_{t, a_t^*}, \theta^*) - f(x_{t, a_t}, \theta^*) \right)\mathbb{I}\left( 2 C_2 \|\theta_{T_0} - \theta^*\| \geq   f(x_{t,a_t^*}, \theta^*)- f(x_{t, a_t}, \theta^*), \mathcal{E}_{T_0} \right) \right] + 2C_1 k_1 \mathbb{P}(\mathcal{E}_{T_0}^c) \\
        & \leq \mathbb{E} \left [ \left(f(x_{t, a_t^*}, \theta^*) - f(x_{t, a_t}, \theta^*) \right)\mathbb{I}\left( 2 C_2 \|\theta_{T_0} - \theta^*\| \geq   f(x_{t,a_t^*}, \theta^*)- f(x_{t,a_t}, \theta^*) > 2C_2 v, \mathcal{E}_{T_0} \right) \right] + 2C_1 k_1 \mathbb{P}(\mathcal{E}_{T_0}^c) \\
        & \quad + \mathbb{E} \left [ \left(f(x_{t, a_t^*}, \theta^*) - f(x_{t, a_t}, \theta^*) \right) \cdot \right.\\
        &\qquad \left. \mathbb{I}\left( 2 C_2 \|\theta_{T_0} - \theta^*\| \geq   f(x_{t, a_t^*}, \theta^*)- f(x_{t, a_t}, \theta^*),  f(x_{t, a_t^*}, \theta^*)- f(x_{t, a_t}, \theta^*) < 2C_2 v ,  \mathcal{E}_{T_0} \right) \right]  \\
        & \leq \mathbb{E} \left [ \left(f(x_{t, a_t^*}, \theta^*) - f(x_{t, a_t}, \theta^*) \right)\mathbb{I}\left( 2 C_2 \|\theta_{T_0} - \theta^*\| \geq   f(x_{t, a_t^*}, \theta^*)- f(x_{t, a_t}, \theta^*) > 2C_2 v, \mathcal{E}_{T_0} \right) \right] \\
        &\quad + 2C_1 k_1 \mathbb{P}(\mathcal{E}_{T_0}^c) + 2C_2v \\ 
        & \leq 2C_1 k_1 \mathbb{P} \left( \|\theta_{T_0} - \theta^*\| >  v, \mathcal{E}_{T_0} \right) + 2C_1 k_1 \mathbb{P}(\mathcal{E}_{T_0}^c) + 2C_2v \\  \\
    \end{aligned} 
\end{equation}

Now take $v$ to be the upper bound of $\|\theta_{T_0} - \theta^*\|$ in Lemma \ref{lem: oracle_inequality}, i.e., $v^2 = 9 \dfrac{\lambda_{T_0}^2}{\alpha^2} \phi^2 + \dfrac{1}{\alpha}[2Z_{T_0}(\theta^*) + 4 \lambda_{T_0} R(\theta^*_{\mathcal{M}^{\bot}})]$. We know by Lemma \ref{lem: oracle_inequality} that $\mathbb{P}(\|\theta_{T_0} - \theta^*\| >  v, \mathcal{E}_{T_0}) \leq \mathbb{P}(\mathcal{A}_{T_0}^c, \mathcal{E}_{T_0})$. Then the expected cumulative regret becomes

\begin{equation}
\tag*{$\square$}
    \begin{aligned}
       \mathbb{E}[ \text{Regret}(T) ]
        &\leq 2C_1k_1T_0 + \sum_{t=T_0}^T \left[ 2C_1 k_1 \mathbb{P} \left( \mathcal{A}_{T_0}^c, \mathcal{E}_{T_0} \right) + 2C_1 k_1 \mathbb{P}(\mathcal{E}_{T_0}^c) + 2C_2v \right] \\
        &\leq 2C_1k_1T_0 + 2C_1 k_1 \sum_{t=T_0}^T \left[  \mathbb{P} \left( \mathcal{A}_{T_0}^c \right) + \mathbb{P}(\mathcal{E}_{T_0}^c)\right] +  2C_2 \sum_{t=T_0}^T \sqrt{9 \dfrac{\lambda_{T_0}^2}{\alpha^2} \phi^2 + \dfrac{1}{\alpha}[2Z_{T_0}(\theta^*) + 4 \lambda_{T_0} R(\theta^*_{\mathcal{M}^{\bot}})]}
    \end{aligned} 
\end{equation}

\subsection{Regret Bound with the Regularization Norm}

In this subsection, we provide a regret bound when Assumption \ref{ass: normalization} and \ref{ass: lipschitz} are assumed base on the regularization norm $R$, which would correspond to, for example, using $l_1$ norm and the compatibility condition in LASSO bandit. The regret bound we derive in Corollary \ref{cor: regret_bound_regularization} can be easily extended to the LASSO bandit, the low-rank matrix bandit and the group-sparse matrix bandit problems. We first provide the second oracle inequality \citep{Negahban2012A}.

\begin{lemma}
\label{lem: oracle_inequality_regularizer_norm}\textbf{\upshape (Oracle Inequality)} If the probability event $\mathcal{A}_t$ holds and $\mathcal{E}_t$ holds for $\theta-\theta^* \in \mathbb{C} \cap \mathbb{B}(r_t)$, where $r_t^2 \geq 9 \frac{\lambda_t^2}{\alpha^2} \phi^2 + \frac{\lambda_t}{\alpha}(2Z_t(\theta^*) + 4R(\theta^*_{\mathcal{M}^{\bot}}))$. Further suppose that $R(\theta^*_{\mathcal{M}^{\bot}}) = 0$, then the difference $\theta_t - \theta^*$ satisfies the bound
\begin{equation}
\nonumber
   R(\theta_t - \theta^*) \leq 4\phi \sqrt{9 \dfrac{\lambda_t^2}{\alpha^2} \phi^2 + \dfrac{2\lambda_t}{\alpha} Z_t(\theta^*)} \\
\end{equation}
\end{lemma}

\begin{assumption}
\label{ass: normalization_R}
We assume $x$ is normalized with respect to the norm $R$., i.e., $R(x) \leq k_1$ for some constant $k_1$ .
\end{assumption}

\begin{assumption}
\label{ass: lipschitz_R}
Here we assume similar Lipschitzness conditions on $f$ and boundedness on $R(x)$, as what we have done in Section \ref{sec: main_results}. 
We assume $f(x, \theta)$ is $C_1$-Lipschitz over $x$ and $C_2$-Lipschitz over $\theta$ with respect to the norm $R$. i.e.,

\begin{equation}
    \begin{aligned}
    \nonumber
        f(x_1, \theta) - f(x_2, \theta) &\leq C_1  R (x_1 - x_2) \\
        f(x, \theta_1) - f(x, \theta_2) &\leq C_2 R (\theta_1 - \theta_2)
    \end{aligned} 
\end{equation}
\end{assumption}

We show that based on such conditions, we can get a similar result as in Theorem \ref{thm: regret_bound}. 

\begin{corollary}
\label{cor: regret_bound_regularization}\textbf{\upshape(Regret Bound)}
Let $T_0 \in \mathbb{N}$ be a constant. Suppose that Assumption \ref{ass: normalization_R} and \ref{ass: lipschitz_R} hold. Also suppose that  $\theta^* \in \mathcal{M}$, then the expected cumulative regret satisfies the bound
\begin{equation}
    \begin{aligned}
    \nonumber
        \mathbb{E}[\text{Regret}(T) ]
        &\leq  2C_1k_1T_0 + 2C_1 k_1 \sum_{t=T_0}^T \left[  \mathbb{P} \left( \mathcal{A}_{T_0}^c \right) + \mathbb{P}(\mathcal{E}_{T_0}^c)\right] +  8C_2 \phi \sum_{t=T_0}^T \sqrt{9 \dfrac{\lambda_{T_0}^2}{\alpha^2} \phi^2 + \dfrac{2\lambda_{T_0}}{\alpha} Z_{T_0}(\theta^*)}
    \end{aligned} 
\end{equation}
\end{corollary}
\textbf{Proof}.
By the Lipschitzness of $f$ over $x$ with respect to $R(\cdot)$, and the boundedness of $R(x)$, we have

\begin{equation}
    \begin{aligned}
    \nonumber
        f(x_{t, a_t^*}, \theta^*) - f(x_{t, a_t}, \theta^*) 
                      \leq C_1 R(x_{t, a_t^*} - x_{t, a_t}) \ \leq 2C_1 k_1
    \end{aligned} 
\end{equation}

Then we can decompose the one-step regret from round $t$ into three parts as follows

\begin{equation}
    \begin{aligned}
    \nonumber
        & f(x_{t, a_t^*}, \theta^*) - f(x_{t, a_t}, \theta^*) \\
                      &=  \left [f(x_{t, a_t^*}, \theta^*) - f(x_{t, a_t}, \theta^*)  \right] \mathbb{I}(t \leq T_0) + \left [f(x_{t, a_t^*}, \theta^*) - f(x_{t, a_t}, \theta^*)   \right] \mathbb{I}(t > T_0, \mathcal{E}_{T_0}) \\
                     & \qquad  + \left [f(x_{t, a_t^*}, \theta^*) - f(x_{t, a_t}, \theta^*)  \right] \mathbb{I}(t \geq T_0, \mathcal{E}_{T_0}^c) \\
                     &\leq 2C_1 k_1 \mathbb{I}(t \leq T_0) +   \left [f(x_{t, a_t^*}, \theta^*) - f(x_{t, a_t}, \theta^*)  \right]  \mathbb{I}(t >  T_0, \mathcal{E}_{T_0}) + 2C_1 k_1\mathbb{I}(t > T_0, \mathcal{E}_{T_0}^c) \\
                    &= 2C_1 k_1 \mathbb{I}(t \leq T_0) +   \left [f(x_{t, a_t^*}, \theta^*) - f(x_{t, a_t}, \theta^*)  \right]  \mathbb{I}(t >  T_0, f(x_{t, a_t}, \theta_{T_0}) \geq f(x_{t, a_t^*}, \theta_{T_0}), \mathcal{E}_{T_0}) + 2C_1 k_1\mathbb{I}(t > T_0, \mathcal{E}_{T_0}^c)
    \end{aligned} 
\end{equation}

where $\mathbb{I}(\cdot)$ is the indicator function. The last equality is due to the choice of $a_t =  \text{argmax}_a f(x_{t,a}, \theta_{T_0})$, and thus we know that $f(x_{t,a_t}, \theta_{T_0}) \geq  f(x_{t,a_t^*}, \theta_{T_0})$. We focus on the second indicator function now. By the Lipschitzness of $f$ over $\theta$, we have

\begin{equation}
    \begin{aligned}
    \nonumber
        & \mathbb{I}\left(t >  T_0, \mathcal{E}_{T_0}\right)  \\
        &= \mathbb{I}\left(t >  T_0, f(x_{t,a_t}, \theta_{T_0}) \geq f(x_{t,a_t^*}, \theta_{T_0}), \mathcal{E}_{T_0}\right)  \\
        &= \mathbb{I}\left(t >  T_0, f(x_{t,a_t}, \theta_{T_0})   - f(x_{t,a_t^*}, \theta_{T_0}) + f(x_{t,a_t^*}, \theta^*) - f(x_{t,a_t}, \theta^*) \geq   f(x_{t,a_t^*}, \theta^*) - f(x_{t,a_t}, \theta^*), \mathcal{E}_{T_0}\right) \\
        &= \mathbb{I}\left(t >  T_0, \left [f(x_{t,a_t}, \theta_{T_0})   - f(x_{t,a_t}, \theta^*) \right] + \left[f(x_{t,a_t^*}, \theta^*) - f(x_{t,a_t^*}, \theta_{T_0})\right] \geq   f(x_{t,a_t^*}, \theta^*) - f(x_{t,a_t}, \theta^*), \mathcal{E}_{T_0}\right) \\
        & \leq \mathbb{I}\left(t >  T_0, 2 C_2 R(\theta_t - \theta^*) \geq   f(x_{t,a_t^*}, \theta^*)- f(x_{t,a_t}, \theta^*), \mathcal{E}_{T_0} \right) \\
    \end{aligned} 
\end{equation}

Substitute the above inequality back and take expectation on both sides of the one-step regret from round $t$, we get

\begin{equation}
    \begin{aligned}
    \nonumber
        \mathbb{E}\left[f(x_{t, a_t^*}, \theta^*) - f(x_{t, a_t}, \theta^*) \right] \leq 2C_1 k_1 \text{ \hfill  for }t \leq T_0
    \end{aligned} 
\end{equation}

For $t > T_0$ and any constant $v \in \mathbb{R}$, the expectation is bounded by

\begin{equation}
    \begin{aligned}
    \nonumber
        &\mathbb{E}\left[f(x_{t, a_t^*}, \theta^*) - f(x_{t, a_t}, \theta^*) \right] \\
        &\leq \mathbb{E} \left [ \left(f(x_{t, a_t^*}, \theta^*) - f(x_{t, a_t}, \theta^*) \right)\mathbb{I}\left( 2 C_2 R(\theta_{T_0} - \theta^*) \geq   f(x_{t, a_t^*}, \theta^*)- f(x_{t, a_t}, \theta^*), \mathcal{E}_{T_0} \right) \right] + 2C_1 k_1 \mathbb{P}(\mathcal{E}_{T_0}^c) \\
        & \leq \mathbb{E} \left [ \left(f(x_{t, a_t^*}, \theta^*) - f(x_{t, a_t}, \theta^*) \right)\mathbb{I}\left( 2 C_2 R(\theta_{T_0} - \theta^*) \geq   f(x_{t, a_t^*}, \theta^*)- f(x_{t, a_t}, \theta^*) > 2C_2 v, \mathcal{E}_{T_0} \right) \right] + 2C_1 k_1 \mathbb{P}(\mathcal{E}_{T_0}^c) \\
        &+ \quad \mathbb{E} \left [ \left(f(x_{t, a_t^*}, \theta^*) - f(x_{t, a_t}, \theta^*) \right) \cdot \right. \\
        &\left. \qquad  \mathbb{I}\left( 2 C_2 R(\theta_{T_0} - \theta^*) \geq   f(x_{t, a_t^*}, \theta^*)- f(x_{t, a_t}, \theta^*),  f(x_{t,a_t^*}, \theta^*)- f(x_{t, a_t}, \theta^*) < 2C_2 v ,  \mathcal{E}_{T_0} \right) \right]  \\
        & \leq \mathbb{E} \left [ \left(f(x_{t, a_t^*}, \theta^*) - f(x_{t, a_t}, \theta^*) \right)\mathbb{I}\left( 2 C_2 R(\theta_{T_0} - \theta^*) \geq   f(x_{t,a_t^*}, \theta^*)- f(x_{t,a_t}, \theta^*) > 2C_2 v, \mathcal{E}_{T_0} \right) \right] \\
        &\quad + 2C_1 k_1 \mathbb{P}(\mathcal{E}_{T_0}^c) + 2C_2v \\ 
        & \leq 2C_1 k_1 \mathbb{P} \left( R(\theta_{T_0} - \theta^*) >  v, \mathcal{E}_{T_0} \right) + 2C_1 k_1 \mathbb{P}(\mathcal{E}_{T_0}^c) + 2C_2v \\  \\
    \end{aligned} 
\end{equation}

Now take $v$ to be the upper bound of $R(\theta_{T_0} - \theta^*)$ in Lemma \ref{lem: oracle_inequality_regularizer_norm}, i.e., $v =4\phi \sqrt{9 \dfrac{\lambda_{T_0}^2}{\alpha^2} \phi^2 + \dfrac{2\lambda_{T_0}}{\alpha} Z_{T_0}(\theta^*)}$. We know by Lemma \ref{lem: oracle_inequality_regularizer_norm} that $\mathbb{P}(R(\theta_{T_0} - \theta^*) >  v, \mathcal{E}_{T_0}) \leq \mathbb{P}(\mathcal{A}_{T_0}^c, \mathcal{E}_{T_0})$. Then the expected cumulative regret becomes

\begin{equation}
\tag*{$\square$}
    \begin{aligned}
       \mathbb{E}[ \text{Regret}(T) ]
        &\leq 2C_1k_1T_0 + \sum_{t=T_0}^T \left[ 2C_1 k_1 \mathbb{P} \left( \mathcal{A}_{T_0}^c, \mathcal{E}_{T_0} \right) + 2C_1 k_1 \mathbb{P}(\mathcal{E}_{T_0}^c) + 2C_2v \right] \\
        &\leq 2C_1k_1T_0 + 2C_1 k_1 \sum_{t=T_0}^T \left[  \mathbb{P} \left( \mathcal{A}_{T_0}^c \right) + \mathbb{P}(\mathcal{E}_{T_0}^c)\right] +  8C_2 \phi \sum_{t=T_0}^T \sqrt{9 \dfrac{\lambda_{T_0}^2}{\alpha^2} \phi^2 + \dfrac{2\lambda_{T_0}}{\alpha} Z_{T_0}(\theta^*)}
    \end{aligned} 
\end{equation}

\section{Proof for the LASSO Bandit}
\label{sec: appendixLASSO}

\subsection{Notations and Algorithm}

\begin{algorithm}[ht]
   \caption{The ESTC Algorithm for LASSO Bandit}
   \label{alg: lasso}
\begin{algorithmic}[1]
   \STATE \textbf{Input:} $\lambda_{T_0}, K \in \mathbb{N}, L_t(\theta), R(\theta), f(x, \theta), T_0$
   \FOR{$ t= 1$ to $T_0$}
   \STATE Observe $K$ contexts, $x_{t,1}, x_{t,2}, \cdots, x_{t, K}$ 
    \STATE Choose action $a_t$ uniformly randomly 
    \STATE Receive reward $y_{t} = \langle x_{t,a_t}, \theta^* \rangle + \epsilon_{t}$
        \ENDFOR
    \STATE Compute the estimator $\theta_{T_0}$: 
      \qquad $$\theta_{T_0} \in \text{argmin}_{\theta \in \mathbf{\Theta}} \left \{\frac{1}{2{T_0}} \sum_{i=1}^{T_0} (y_i - x_{i, a_i}^T \theta)^2 + \lambda_{T_0} \|\theta\|_1 \right \}$$
      \vspace{-10pt}
    \FOR{$ t=  T_0 +1$ to $T$}
      \STATE Choose action $a_t = \text{argmax}_a \langle x_{t,a_t}, \theta_{T_0} \rangle$ 
   \ENDFOR
\end{algorithmic}
\end{algorithm}

We first clarify the notations we use throughout the proof for the LASSO bandit. We use the notations $X_t \in \mathbb{R}^{t \times d}, Y_t$, $e_t \in \mathbb{R}^t$ to represent the context matrix, the reward and the error vectors respectively, i.e., $[X_t]_i = x_{i, a_i} \in \mathbb{R}^d, [Y_t]_i = y_{i}, [e_t]_i = \epsilon_i, \forall i \in [t]$. The loss function now becomes $L_t(\theta) = \frac{1}{2t}\|Y_t -X_t \theta\|_2^2$. The derivative of $L_t(\theta)$ evaluated at $\theta^*$ can be computed as 

\begin{equation}
    \begin{aligned}
    \nonumber
    \nabla L_t(\theta^*) 
    = - \frac{1}{t} X_t^T Y_t + X_t^TX_t \theta^* 
    = -\frac{1}{t} X_t^T e_t
    \end{aligned}
\end{equation}

The Bregman divergence $ B_t(\theta, \theta^*) $ can therefore be computed as 

\begin{equation}
    \begin{aligned}
    \nonumber
    B_t(\theta, \theta^*)  
    &= L_t(\theta) - L_t(\theta^*) - \langle \nabla L_t(\theta^*), \theta - \theta^* \rangle \\
    &= \frac{1}{2t}\|Y_t - X_t\theta\|_2^2 -  \frac{1}{2t}\|Y_t - X_t\theta^*\|_2^2 - \frac{1}{t} \langle  -X_t^T e_t, \theta - \theta^* \rangle \\
    &= \frac{1}{2t}(\theta - \theta^*)^TX_t^TX_t(\theta - \theta^*)
    \end{aligned}
\end{equation}

Therefore the event $\mathcal{A}_t$ is equivalent to $\{\lambda_t \geq 2\|\nabla L_t(\theta^*)\|_{\infty}  = \frac{2}{t}\|X_t^T e_t\|_{\infty}\}$. The event $\mathcal{E}_t$ (RSC condition) is equivalent to 

\begin{equation}
    \begin{aligned}
    \nonumber
    \frac{1}{2t}(\theta - \theta^*)^TX_t^TX_t(\theta - \theta^*) \geq \alpha \|\theta - \theta^*\|_2^2 - Z_t(\theta^*)
    \end{aligned}
\end{equation}

Based on the above Bregman divergence, we define the matrix $\hat{\Sigma}_t \in \mathbb{R}^{d \times d}$  as follows . 

\begin{equation}
    \begin{aligned}
    \nonumber
    \hat{\Sigma}_t  = \frac{1}{t}\sum_{\tau=1}^t x_{\tau, a_\tau}x_{\tau, a_\tau}^T
    \end{aligned}
\end{equation}

\subsection{Technical Lemmas}
The following two Bernstein-type inequalities are very useful in our analysis.

\begin{lemma}
\label{lem: bernstein_concentration}\textbf{\upshape (Lemma EC.1 of \citet{Bastani2015online})}
Let $\{D_k, \mathcal{F}_k\}_{k=1}^\infty$ be a martingale difference sequence, and suppose that $D_k$ is $\sigma$-sub-Gaussian in an adapted sense, i.e., for all $\alpha \in \mathbb{R}, \mathbb{E}\left[e^{\alpha D_k}|\mathcal{F}_{k-1}\right] \leq e^{\alpha^2\sigma^2/2} $ almost surely. Then, for all $t \geq 0$, we have

\begin{equation*}
\mathbb{P} \left(|\sum_{k=1}^n D_k| \geq t \right) 
\leq 2 e^{ - { t^2}/(2n\sigma^2)}
\end{equation*}

\end{lemma}

\begin{lemma}
\label{lem: T0_bernstein}
 \textbf{\upshape (Lemma EC.4. of \citet{Bastani2015online})}
Let $x_{1}, x_{2}, \cdots, x_{t}$ be i.i.d. random vectors in $\mathbb{R}^{d}$ with $\left\|x_{\tau}\right\|_{\infty} \leq 1$ for all $\tau .$ Let $\Sigma=\mathbb{E}\left[x_{\tau} x_{\tau}^{T}\right]$ and $\hat{\Sigma}_{t}=\frac{1}{t} \sum_{\tau=1}^{t} x_{\tau} x_{\tau}^{T} .$. Then for any $w > 0$, we have
$$
\mathbb{P}\left[ \vertiii{\hat{\Sigma}_t  - \Sigma\|}_{\max}\geq 2\left(w + \sqrt{2w} + \sqrt{\frac{2\log(d^2+d)}{t}} +  {\frac{\log(d^2+d)}{t}} \right) \right] \leq \exp \left(-tw\right)
$$

\end{lemma}

\subsection{Proof of Lemma \ref{lem: lasso_lambda_condition}}
\textbf{Proof.} Denote  $X_t^{(r)}$ to be the $r$-th column of the matrix $X_t$. Since we assume that each $\|x_t\|_{2} \leq 1$, then the matrix $X_t$ is column normalized, i.e., $\|X_t^{(r)}\|_2 \leq \sqrt{t}$. From the union bound, we know that for any constant $c \in \mathbb{R}$, we have

\begin{equation}
    \begin{aligned}
    \nonumber
  \mathbb{P}\left(\frac{2}{t}\|X_t^T e_t\|_{\infty} \leq c \right) 
  &=  \mathbb{P}\left(\forall r \in [d], |\langle e_t, X_t^{(r)}  \rangle| \leq \frac{ct}{2} \right) \\
  &\geq 1 - \sum_{r=1}^d \mathbb{P} \left( |\langle e_t, X_t^{(r)}  \rangle| \geq \frac{ct}{2} \right)\\
    \end{aligned} 
\end{equation}

Note that $\{\langle e_t, X_t^{(r)}  \rangle\}_{\tau=1}^t$ is a martingale difference sequence adapted to the filtration $\mathcal{F}_1 \subset \mathcal{F}_2 \subset \cdots \subset \mathcal{F_\tau} $, because $\mathbb{E}[\langle e_t, X_t^{(r)}  \rangle\mid \mathcal{F}_{t-1 }] = 0$. Also note that each $\epsilon_t$ is $\sigma^2$-sub-Gaussian, therefore

\begin{equation}
    \begin{aligned}
    \nonumber
   \mathbb{E}\left[\text{exp}\left(\alpha \langle e_t, X_t^{(r)}  \rangle \right)|\mathcal{F}_{k-1} \right]
   &=  \mathbb{E}\left[\text{exp}\left(\alpha \sum_{i=1}^t X_{t,i}^{(r)}  \epsilon_i\right)|\mathcal{F}_{k-1} \right] \\
   &= \mathbb{E}_{X_t^{(r)} }\left[\mathbb{E}_{\epsilon_t|x_{\tau}}\left [\text{exp}\left( \alpha \sum_{i=1}^t X_{t,i}^{(r)} \epsilon_i \right) \mid \mathcal{F}_{k-1}, x_{\tau}\right]\right] \\
   &\leq \mathbb{E}_{X_t^{(r)} }\left[\text{exp}\left( \sum_{i=1}^t \alpha^2 X_{t,i}^{2(r)} \sigma^2 /2 \right)\mid \mathcal{F}_{k-1}  \right] \\
   &\leq \text{exp} \left({\alpha^2 \sigma^2 t /2} \right)
    \end{aligned} 
\end{equation}

Hence we can use {Lemma \ref{lem: bernstein_concentration}} and thus

\begin{equation}
    \begin{aligned}
    \nonumber
    \mathbb{P}\left(\frac{2}{t}  \big \lVert X_t^T e_t \rVert_{\infty} \leq c \right)
    &\geq 1 - \sum_{\tau=1}^d \mathbb{P}\left( |\langle e_t, X_t^{(r)}  \rangle|\geq \frac{ct}{2} \right) \\
    &\geq 1 - 2d e^{-{c^2 t}/(8\sigma^2)}
    \end{aligned} 
\end{equation}

If we take $\lambda_t = c = \sqrt{(8 \sigma^2 \log \frac{2d}{\delta})/t}$, then with probability $1-\delta$, we conclude that

\begin{equation}
\tag*{$\square$}
    \begin{aligned}
    \nonumber
    \mathbb{P}\left(\frac{2}{t}\|X_t^T e_t\|_{\infty} \leq \lambda_t \right) \geq 1 - \delta
    \end{aligned} 
\end{equation}

\subsection{ Proof of Lemma \ref{lem: lasso_RSC_condition}}

{
\begin{lemma}
\label{lem: lasso_RSC_condition}
Suppose Assumption \ref{ass: alpha_covariance} is satisfied, then with probability at least $1 -  \text{exp}\left(-\frac{T_0 C(s)}{4}\right) $, we have $B_{{T_0}}(\theta, \theta^*) \geq \frac{\alpha_0}{4} \|\theta - \theta^*\|_2^2$, for all $ T_0 \geq 2\log(d^2+d)/C(s)$ and $\theta - \theta^* \in \mathbb{C}$, where $C(s) > 0$ is a constant that depends on $s$.
\end{lemma}
}

First, we introduce the following two technical lemmas.

\begin{lemma}
\label{lem: lasso_RSC_difference}
(\textbf{ \upshape Lemma 9 of \citet{oh2020sparsityagnostic}})
Suppose that $\Sigma_1, \Sigma_2 \in \mathbb{R}^{d\times d}$, $\beta \in \mathbb{C} \cap \mathbb{R}^{d}$, and that the matrix $\Sigma_1$ satisfies the restricted eigenvalue condition $ \beta^T \Sigma_1\beta \geq \alpha  \|\beta\|_{2}^2$  with $\alpha>0$. Moreover, suppose the two matrices are close enough such that $\vertiii{\Sigma_{2}-\Sigma_{1}}_{\max} \leq \delta,$ where $32 s \delta \leq \alpha .$ Then $$\beta^T \Sigma_2 \beta \geq \frac{\alpha}{2}  \|{\beta}\|_{2}^2 $$
\end{lemma}
\textbf{Proof}. The proof here modifies the proof by \citet{oh2020sparsityagnostic}, and we provide it for completeness. By the Cauchy-Schwartz inequality, we have

\begin{equation}
    \begin{aligned}
    \nonumber
       \left | \beta^T \Sigma_1 \beta - \beta^T \Sigma_2 \beta  \right |
      &= \left |\beta^T (\Sigma_1 - \Sigma_2) \beta \right |\\
      &\leq  \|(\Sigma_1 - \Sigma_2) \beta\|_\infty \|\beta\|_1 \\
      &\leq  \vertiii{\Sigma_1 - \Sigma_2}_{\max} \|\beta\|_1^2 \\
      &\leq \delta \|\beta\|_1^2\\
    \end{aligned}
\end{equation}

For $\beta \in \mathbb{C}$, we have the inequality $\|\beta_{\overline{M}^{\bot}}\|_1 \leq 3 \|\beta_{\overline{M}}\|_1$. Thus

\begin{equation}
    \begin{aligned}
    \nonumber
    \|\beta\|_1 \leq 4\|\beta_{\overline{M}}\|_1 \leq  4\phi\|\beta_{\overline{M}}\|_2 \leq 4\phi\|\beta\|_2 \leq 4\phi\sqrt{\frac{1}{\alpha} \beta^T \Sigma_1\beta}
    \end{aligned}
\end{equation}

Therefore since we assume that $32s \delta  \leq \alpha $, we have

\begin{equation}
    \begin{aligned}
    \nonumber
       \left | \beta^T \Sigma_1 \beta - \beta^T \Sigma_2 \beta  \right |
       \leq \delta \|\beta\|_1^2 
       \leq  \frac{16\phi^2\delta}{\alpha } \beta^T \Sigma_1\beta 
       \leq  \frac{1}{2} \beta^T \Sigma_1 \beta\\
    \end{aligned}
\end{equation}

By some simple algebra on the above inequality, we know that

\begin{equation}
\tag*{$\square$}
    \begin{aligned}
    \beta^T \Sigma_2 \beta \geq \frac{\alpha}{2}  \|{\beta}\|_{2}^2
    \end{aligned}
\end{equation}

\begin{lemma}
\label{lem: lasso_sigma_distance}
\textbf{\upshape (Distance Between Two Matrices)}
Define $C(s) = (\sqrt{ \frac{\alpha_0}{64 s} + 1} - 1)^2 = \Theta(s^{-1})$, where $\alpha_0$ is defined in Assumption \ref{ass: alpha_covariance}. Then for all $T_0 \geq 2\log(d^2+d)/C(s)$,  we have
\begin{equation}
\nonumber
    \begin{aligned}
    \mathbb{P}\left( \vertiii{\Sigma - \hat{\Sigma}_{T_0}}_{\max} \geq \frac{\alpha_0}{32s} \right)
    \leq \exp \left(-\frac{T_0 C(s)}{2}\right) 
    \end{aligned}
\end{equation}
\end{lemma}

\textbf{Proof}. Since we sample uniformly in the exploration stage, we know that the contexts $\{x_{\tau, a_{\tau}}\}_{\tau=1}^{T_0}$ are i.i.d., and thus for any constant $w$,  by Lemma \ref{lem: T0_bernstein} we have

\begin{equation*}
    \mathbb{P}\left(\frac{1}{2}\vertiii{\hat{\Sigma}_{T_0} - \Sigma}_{\max} 
    \geq w + \sqrt{2w} + \sqrt{\frac{2\log(d^2+d)}{T_0}} +  {\frac{\log(d^2+d)}{T_0}}  \right) \leq \exp \left(-wT_0\right)
\end{equation*}

Now our choices of $w$ and $T_0$ are to let both terms concerning $w$ and $T_0$ to be small, i.e.,

$$ w + \sqrt{2w} \leq \frac{\alpha_0}{128s} 
\quad \text{ and } \quad
\sqrt{\frac{2 \log \left( d^{2} + d\right)}{T_0}}+\frac{ \log \left( d^{2} + d\right)}{T_0} \leq \frac{\alpha_0}{128s}$$

Solving the two inequalities leads to
$$w = \frac{1}{2} (\sqrt{ \frac{\alpha_0}{64s} + 1} - 1)^2
\quad \text{ and } \quad 
T_0 \geq 2\log(d^2+d) / (\sqrt{ \frac{\alpha_0 }{64s} + 1} - 1)^2$$
then we have the following inequality

\begin{equation*}
\tag*{$\square$}
\begin{aligned}
    \mathbb{P}\left(\frac{1}{2}\vertiii{ \hat{\Sigma}_{T_0} - \Sigma }_{\max} \geq \frac{\alpha_0}{64s} \right) 
    \leq \exp \left(-w T_0\right)
    = \exp \left(-\frac{T_0 C(s)}{2}\right)
\end{aligned}
\end{equation*}

Finally, we provide the proof of Lemma \ref{lem: lasso_RSC_condition}.

\textbf{Proof.} The proof follows from combining Lemma \ref{lem: lasso_RSC_difference} and Lemma \ref{lem: lasso_sigma_distance}. Since we have that the restricted eigenvalue condition holds for $\Sigma$ by Assumption \ref{ass: alpha_covariance} , i.e., 

\begin{equation}
    \begin{aligned}
    \nonumber
 (\theta - \theta^* )^T {\Sigma} (\theta - \theta^* ) \geq \alpha_0 \|\theta - \theta^*\|_2^2, \forall \theta - \theta^* \in \mathbb{C}
    \end{aligned} 
\end{equation}

Also, the two matrices  $\Sigma$,  $\hat{\Sigma}_{T_0}$ are close enough when $T_0$ is large by Lemma \ref{lem: lasso_sigma_distance},

\begin{equation}
\nonumber
    \begin{aligned}
    \mathbb{P}\left( \vertiii{\Sigma - \hat{\Sigma}_{T_0}}_{\max} \geq \frac{\alpha_0}{32s} \right)
    \leq \exp \left(-\frac{T_0 C(s)}{2}\right) 
    \end{aligned}
\end{equation}
where $C(s)$ is a constant defined as in  Lemma \ref{lem: lasso_sigma_distance}. By  Lemma \ref{lem: lasso_RSC_difference} we can claim that

\begin{equation}
    \begin{aligned}
    \nonumber
 (\theta - \theta^*)^T \hat{\Sigma}_{T_0} (\theta - \theta^*) \geq \frac{\alpha_0}{2} \|\theta - \theta^*\|_2^2, \forall \theta - \theta^* \in \mathbb{C}
    \end{aligned} 
\end{equation}

with high probability $1 - \text{exp}(-\frac{{T_0}C(s)}{2})$ when $T_0 \geq 2\log(d^2+d)/C(s)$. Therefore

\begin{equation}
\nonumber
    \begin{aligned}
    B_{T_0}(\theta, \theta^*) = \frac{1}{2} (\theta - \theta^*)^T \hat{\Sigma}_{T_0} (\theta - \theta^*) \geq \frac{\alpha_0}{4} \|\theta - \theta^*\|_2^2, \forall \theta - \theta^* \in \mathbb{C}
    \end{aligned} 
\end{equation}

\subsection{ Proof of Corollary \ref{cor: lasso_regret}}
Now given the results in Lemma \ref{lem: lasso_lambda_condition}, Lemma \ref{lem: lasso_RSC_condition} and Theorem \ref{thm: regret_bound}, we can easily derive an upper bound for our LASSO bandit algorithm. We specify all the constants here. For example, the Lipschitz constants $C_1 = C_2 = 1$, the bound of the norm $k_1 = 1$, and we set $T_0 \geq 2\log(d^2+d)/C(s) $, where $C(s) $ is a constant defined in Lemma \ref{lem: lasso_sigma_distance}.  The restricted strong convexity holds with $\alpha = \alpha_0/4$. The compatibility constant $\phi = \sqrt{s}$. Also, $\theta^*_{\mathcal{M}^{\bot}} = 0$ so $R(\theta^*_{\mathcal{M}^{\bot}}) = 0$.

\textbf{Proof.} Using Theorem \ref{thm: regret_bound}, we can get the following upper bound

\begin{equation*}
    \begin{aligned}
       \mathbb{E}[ \text{Regret}(T) ]
        &\leq 2 T_0 + 2  \sum_{t = T_0}^T\left[  \mathbb{P} \left( \mathcal{A}_{T_0}^c \right) + \mathbb{P}(\mathcal{E}_{T_0}^c)\right] +  2  \sum_{t = T_0}^T\sqrt{ \dfrac{144\lambda_{T_0}^2}{\alpha_0^2} s+ \dfrac{8\lambda_{T_0}}{\alpha_0} Z_{T_0}(\theta^*)} \\
    \end{aligned} 
\end{equation*}

Note that $Z_{T_0}(\theta^*) = 0$. Since we take $\delta = 1/T^2$ in Lemma \ref{lem: lasso_lambda_condition}, $P(\mathcal{A}_{T_0}^c) \leq {1}/{T^2}$. Furthermore,

\begin{equation*}
    \begin{aligned}
       \sum_{t = T_0}^T \mathbb{P} \left( \mathcal{A}_{T_0}^c \right)
        \leq  \sum_{t = T_0}^T \frac{1}{T^2}
        \leq \sum_{t = 1}^{T} \frac{1}{T^2} 
        \leq \frac{1}{T}
    \end{aligned} 
\end{equation*}

Also by our choice of $T_0 = \Theta(s^{1/3}T^{2/3})$, we know that

\begin{equation*}
    \begin{aligned}
        \sum_{t = T_0}^T \mathbb{P} \left( \mathcal{E}_{T_0}^c \right)
        \leq \sum_{t = T_0}^T  \text{ exp }( - \frac{T_0 C(s)}{2} )
        \leq T\text{ exp }( - \frac{T_0 C(s)}{2} ) = \mathcal{O}(1)
    \end{aligned} 
\end{equation*}

where the last equality is by the fact that $ x^{3/2} e^{-ax}$ is bounded above by a constant $(\frac{3}{2a})^3 e^{-\frac{9}{4a}}$ when $ a > 0$. Therefore we arrive to the following bound

\begin{equation*}
    \begin{aligned}
       \mathbb{E}[ \text{Regret}(T) ]
        & \leq 2 T_0 + 2 \sum_{t = T_0}^T \left[  \mathbb{P} \left( \mathcal{A}_{T_0}^c \right) + \mathbb{P}(\mathcal{E}_{T_0}^c)\right] +  2 \sum_{t = T_0}^T \sqrt{ \dfrac{144\lambda_{T_0}^2}{\alpha_0^2} s + \dfrac{8\lambda_{T_0}}{\alpha_0} Z_{T_0}(\theta^*)} \\
        & \leq 2 T_0 + 2 \sum_{t = T_0}^T \left[  \mathbb{P} \left( \mathcal{A}_{T_0}^c \right) + \mathbb{P}(\mathcal{E}_{T_0}^c)\right] +  24(T - T_0 ) \frac{\lambda_{T_0}}{\alpha_0} \sqrt{s}\\
        & \leq 2 T_0 + 2 \sum_{t = T_0}^T \left[  \mathbb{P} \left( \mathcal{A}_{T_0}^c \right) + \mathbb{P}(\mathcal{E}_{T_0}^c)\right] +   \frac{T}{\sqrt{T_0}} \frac{48\sigma \sqrt{s}}{\alpha_0}  \sqrt{{2 \log (2d T )}} \\
    \end{aligned} 
\end{equation*}

Matching the two dominating terms, we get the choice of $T_0 = s^{1/3}(\frac{24 \sigma}{\alpha_0})^{2/3} T^{2/3}$ and the regret upper bound

\begin{equation*}
    \begin{aligned}
       \mathbb{E}[ \text{Regret}(T) ]
        & \leq  4 s^{1/3}(\frac{24\sigma}{\alpha_0})^{2/3} T^{2/3} \sqrt{2\log(2dT)} + 2 \sum_{t = T_0}^T \left[  \mathbb{P} \left( \mathcal{A}_{T_0}^c \right) + \mathbb{P}(\mathcal{E}_{T_0}^c)\right] \\
        & = \mathcal{O} \left( s^{1/3} T^{2/3} \sqrt{\log(dT)} \right)\\
    \end{aligned} 
\end{equation*}

\hfill $\square$

\section{Proof for the Low-Rank Matrix Bandit}
\label{sec: appendixMatrix}

\subsection{Notations and Algorithm}

\begin{algorithm}[ht]
   \caption{The ESTC Algorithm for Low Rank Matrix Bandit}
   \label{alg: low_rank}
\begin{algorithmic}[1]
   \STATE \textbf{Input:} $\lambda_{T_0}, K \in \mathbb{N}, L_t(\theta), R(\theta), f(x, \theta),  T_0$
   \FOR{$ t= 1$ to $T_0$}
   \STATE Observe $K$ contexts, $X_{t,1}, X_{t,2}, \cdots, X_{t, K}$ 
    \STATE Choose action $a_t$ uniformly randomly 
    \STATE Receive reward $y_{t} = \vertangl{ X_{t,a_t}, \Theta^* } + \epsilon_{t} $
        \ENDFOR
    \STATE Compute the estimator $\Theta_{T_0}$: 
      \qquad $$\Theta_{T_0} \in \text{argmin}_{\Theta \in \mathbf{\Theta}} \left \{\frac{1}{2T_0} \sum_{i=1}^{T_0}(y_{i}- \vertangl{X_{i, a_i}, \Theta} )^{2} + \lambda_t \vertiii{\Theta}_{nuc} \right \}$$
      \vspace{-10pt}
    \FOR{$ t=  T_0 +1$ to $T$}
      \STATE Choose action $a_t = \text{argmax}_a  \vertangl{ X_{t,a_t}, \Theta_{T_0} }$ 
   \ENDFOR
\end{algorithmic}
\end{algorithm}

We first establish the notations and the corresponding algorithm in low rank matrix bandit problems.  We use the shorthand vector notations $\mathcal{X}_t(\Theta^*), Y_t, e_t \in \mathbb{R}^t$ such that $[\mathcal{X}_t(\Theta^*)]_i = \vertangl{X_{i, a_i}, \Theta^*}$, $[Y_t]_i = y_i,$ and $[e_t]_i = \epsilon_i$. For any matrix $A$ , $vec(A)$ denotes the vectorization of the matrix by stacking all rows of the matrix into a vector, i.e., if $A$ has entries

\begin{equation}
    \begin{aligned}
    \nonumber
    \begin{bmatrix}
        A_{11} & A_{12}& \cdots & A_{1n} \\
        A_{21} & A_{22}& \cdots & A_{2n} \\
        \cdots \\
        A_{m1} & A_{m2}& \cdots & A_{mn} \\
    \end{bmatrix}\\
    \end{aligned}
\end{equation}

then $vec(A) = (A_{11}, A_{12}, \cdots A_{1n}, A_{21}, A_{22}, \cdots, A_{2n}, \cdots A_{m1}, A_{m2}, \cdots, A_{mn})^T$, and $\vertangl{A, B} = \text{tr }(A^TB) = {vec}(A)^T {vec}(B)$.
 
The loss function now becomes $L_t(\Theta) = \frac{1}{2t}\|Y_t - \mathcal{X}_t(\Theta)\|_2^2$, and the derivative with respect to $\Theta^*$ can be computed as 

\begin{equation}
    \begin{aligned}
    \nonumber
\nabla L_t(\Theta^*) 
&= \dfrac{\partial}{\partial \Theta^*} \frac{1}{2t}(Y_t - \mathcal{X}_t(\Theta^*))^T (Y_t - \mathcal{X}_t(\Theta^*)) \\
&= \frac{1}{2t} \dfrac{\partial}{\partial \Theta^*} ( - 2\mathcal{X}_t(\Theta^*)^T Y_t + \mathcal{X}_t(\Theta^*)^T \mathcal{X}_t(\Theta^*)) \\
&= - \frac{1}{t} \sum_{i=1}^t X_{i, a_i} \epsilon_i \\
    \end{aligned} 
\end{equation}

where the differentiation is based on the chain rule of matrix calculus and the following facts

\begin{equation}
    \begin{aligned}
    \nonumber
\dfrac{\partial(- 2\mathcal{X}_t(\Theta^*)^T Y_t + \mathcal{X}_t(\Theta^*)^T \mathcal{X}_t(\Theta^*))}{\partial \mathcal{X}_t(\Theta^*)}
&= - 2 e_t \\
\dfrac{\partial [\mathcal{X}_t(\Theta^*)]_i }{\partial \Theta^*_{j,k}} = \dfrac{\partial \text{ tr}(X_{i, a_i}^T \Theta^*)}{\partial \Theta^*_{j,k}}
&= X_{i, a_i; j,k} \\
    \end{aligned} 
\end{equation}

Therefore the Bregman divergence $B_t(\Theta, \Theta^*)$ in the low rank matrix setting is 

\begin{equation}
    \begin{aligned}
    \nonumber
    &B_t(\Theta, \Theta^*) \\
    &= L_t(\Theta) - L_t(\Theta^*) - \langle \nabla L_t(\Theta^*), \Theta - \Theta^* \rangle \\
    &= \frac{1}{2t} \|Y_t - \mathcal{X}_t(\Theta)\|_2^2 -  \frac{1}{2t} \|Y_t - \mathcal{X}_t(\Theta^*)\|_2^2 +  \vertangl{\frac{1}{t} \sum_{i=1}^t X_{i, a_i} \epsilon_i , \Theta - \Theta^*}\\
    &= \frac{1}{2t} [ - 2\mathcal{X}_t(\Theta)^T Y_t + \mathcal{X}_t(\Theta)^T\mathcal{X}_t(\Theta) + 2\mathcal{X}_t(\Theta^*)^T Y_t - \mathcal{X}_t(\Theta^*)^T\mathcal{X}_t(\Theta^*)] +  \vertangl{\frac{1}{t} \sum_{i=1}^t X_{i, a_i} \epsilon_i , \Theta - \Theta^*} \\
    &= \frac{1}{2t} [ - 2\mathcal{X}_t(\Theta)^T  e_t  - 2\mathcal{X}_t(\Theta)^T\mathcal{X}_t(\Theta^*) + \mathcal{X}_t(\Theta)^T\mathcal{X}_t(\Theta) + 2\mathcal{X}_t(\Theta^*)^T e_t + \mathcal{X}_t(\Theta^*)^T\mathcal{X}_t(\Theta^*)] +  \vertangl{\frac{1}{t} \sum_{i=1}^t X_{i, a_i} \epsilon_i , \Theta - \Theta^*} \\
    &= \frac{1}{2t}[\mathcal{X}_t(\Theta)^T\mathcal{X}_t(\Theta)  - 2\mathcal{X}_t(\Theta)^T\mathcal{X}_t(\Theta^*) + \mathcal{X}_t(\Theta^*)^T\mathcal{X}_t(\Theta^*)] \\
    &= \frac{1}{2t}[(\mathcal{X}_t(\Theta) - \mathcal{X}_t(\Theta^*))^T(\mathcal{X}_t(\Theta) - \mathcal{X}_t(\Theta^*))] \\
    &= \frac{1}{2t} \sum_{i=1}^t \vertangl{X_{i, a_i}, \Theta - \Theta^*}^2 \\
    \end{aligned}
\end{equation}

Note that one important property of the above Bregman divergence is 
\begin{equation}
    \begin{aligned}
    \nonumber
    B_t(\Theta, \Theta^*)  
    &= \frac{1}{2t} \sum_{i=1}^t \vertangl{X_{i, a_i}, \Theta - \Theta^*}^2 \\
    &=  \frac{1}{2t} \sum_{i=1}^t [vec(X_{i, a_i})^T vec(\Theta - \Theta^*)]^2  \\ &=  \frac{1}{2t} \sum_{i=1}^t vec(\Theta - \Theta^*)^T [vec(X_{i, a_i}) vec(X_{i, a_i})^T] vec(\Theta - \Theta^*)  \\ 
    \end{aligned}
\end{equation}

Based on the above discussions, we define the matrix $\hat{\Sigma}_t \in \mathbb{R}^{d_1 d_2 \times d_1 d_2}$ as follows

\begin{equation}
    \begin{aligned}
    \nonumber
    \hat{\Sigma}_t  = \frac{1}{t}\sum_{\tau=1}^t vec(X_{\tau, a_\tau})vec(X_{\tau, a_\tau})^T
    \end{aligned}
\end{equation}

Also by Assumption \ref{ass: alpha_covariance}, we know that for any $B \in \mathbb{C}$
\begin{equation}
    \begin{aligned}
    \nonumber
 vec(B)^T  \Sigma vec(B) \geq \alpha_0 \|vec({B})\|_2^2 = \alpha_0 \vertiii{B}_F^2
    \end{aligned}
\end{equation}

\subsection{Technical Lemmas}
The following matrix version of the Bernstein inequality is useful in our analysis of matrix bandits.

\begin{lemma}
\label{lem: matrix_bernstein}
\textbf{\upshape (Theorem 3.2 of \citet{Recht2011A})}
Let $X_{1}, \ldots, X_{L}$ be independent zero-mean random matrices of dimension $d_{1} \times d_{2} .$ Suppose $\rho_{k}^{2}=\max \left\{\vertiii{\mathbb{E}[X_{k} X_{k}^{T}]}_{op} ,\vertiii{\mathbb{E}\left[X_{k}^{T} X_{k}\right]}_{op}\right\}$ and $\vertiii{X_{k}}_{op} \leq$
M almost surely for all $k .$ Then for any $\tau>0$, we have
$$
\mathbb{P}\left[\vertiii{\sum_{k=1}^{L} X_{k}}_{op}>\tau\right] \leq\left(d_{1}+d_{2}\right) \exp \left(\frac{-\tau^{2} / 2}{\sum_{k=1}^{L} \rho_{k}^{2}+M \tau / 3}\right)
$$
\end{lemma}

\subsection{Proof of Lemma \ref{lem: matrix_lambda_condition}}
\textbf{Proof.} 
The requirement on $\lambda_t$ in terms of the spectral norm of $\nabla L_t(\Theta^*)$ is

\begin{equation}
    \begin{aligned}
    \nonumber
\mathbb{P} \left(\lambda_t \geq 2\vertiii{\frac{1}{t} \sum_{i=1}^t X_{i, a_i} \epsilon_i}_{op} \right) 
=  1 - \mathbb{P} \left(\lambda_t \leq 2\vertiii{\frac{1}{t} \sum_{i=1}^t X_{i, a_i} \epsilon_i}_{op} \right) \\
    \end{aligned} 
\end{equation}

Here we define the high probability event $\mathcal{B}:=\left\{\max _{i=1, \ldots, t}\left|\epsilon_{i}\right| < v \right\}$, and we first concentrate on the probability of the event conditioned on fixed matrices $\{X_{i, a_i}\}_{i=1}^T$. By the definition of sub-Gaussian random variables.

\begin{equation}
    \begin{aligned}
    \nonumber
\mathbb{P} (\mathcal{B}^c) 
&= \mathbb{P}(\max _{i=1, \ldots, t}\left|\epsilon_{i}\right|> v) \leq \sum_{i=1}^t \mathbb{P}(\left|\epsilon_{i}\right|> v) \leq t \text{ exp}(- \frac{v^2}{2\sigma^2})
    \end{aligned} 
\end{equation}

where the first inequality is due to the union bound. Hence if we take $v = \sigma \sqrt{2 \log (2t/\delta)}$, then $\mathbb{P}(\mathcal{B}^c) 
\leq \delta / 2$. Under the event $\mathcal{B}$, the operator norm of each $X_{i, a_i}\epsilon_i$ can be bounded by

\begin{equation}
    \begin{aligned}
    \nonumber
\vertiii{X_{i, a_i}\epsilon_i}_{op} \leq \vertiii{X_{i, a_i}\epsilon_i}_{F} =  \sqrt{\sum_{j=1}^{d_1}\sum_{k=1}^{d_2} X_{a_i; j, k}^2\epsilon_i^2 } \leq v \vertiii{X_{i, a_i}}_{F} =  \sigma \sqrt{2 \log (2t/\delta)}
    \end{aligned} 
\end{equation}

Moreover, under the event $\mathcal{B}$, we have the following bounds

\begin{equation}
    \begin{aligned}
    \nonumber
&\max \left\{ \vertiii{\mathbb{E} [\epsilon_{i}^{2} X_{i, a_i} X_{i, a_i}^{T}  \mid \{X_{i, a_i}\}_{i=1}^t]}_{o p}, \vertiii{\mathbb{E} [\epsilon_{i}^{2} X_{i, a_i}^T X_{i, a_i} \mid \{X_{i, a_i}\}_{i=1}^t]}_{op} \right \}\leq {2 \sigma^2\log (2t/\delta)} 
    \end{aligned} 
\end{equation}

Therefore we have

\begin{equation}
    \begin{aligned}
    \nonumber
&\mathbb{P} \left(\lambda_t \geq 2\vertiii{\frac{1}{t} \sum_{i=1}^t X_{i, a_i} \epsilon_i}_{op} \Bigg \vert \{X_{i, a_i}\}_{i=1}^t \right) \\
&= 1 - \mathbb{P} \left(\lambda_t \leq 2\vertiii{\frac{1}{t} \sum_{i=1}^t X_{i, a_i} \epsilon_i}_{op}, \mathcal{B}^c \Bigg \vert \{X_{i, a_i}\}_{i=1}^t \right) -  \mathbb{P} \left(\lambda_t \leq 2\vertiii{\frac{1}{t} \sum_{i=1}^t X_{i, a_i} \epsilon_i}_{op} , \mathcal{B} \Bigg \vert \{X_{i, a_i}\}_{i=1}^t \right)\\
&\geq 1 -  \mathbb{P} \left(\mathcal{B}^c \right) -  \mathbb{P} \left(\lambda_t \leq 2\vertiii{\frac{1}{t} \sum_{i=1}^t X_{i, a_i} \epsilon_i}_{op} , \mathcal{B}  \Bigg \vert \{X_{i, a_i}\}_{i=1}^t  \right) \\
&\geq 1 - \left (\frac{\delta}{2} + (d_1+d_2)\text{exp}(\frac{-\lambda_t^2 t^2/8}{{2t \sigma^2 \log (2t/\delta)} + \lambda_t t \sigma \sqrt{2 \log (2t/\delta)} / 6}) \right) 
    \end{aligned} 
\end{equation}

where the last inequality is by Lemma \ref{lem: matrix_bernstein} with the corresponding upper bound constants computed above. Now we can further bound the probability using a specific choice of $\lambda_t$, i.e., we want that 

\begin{equation}
    \begin{aligned}
    \nonumber
(d_1+d_2)\text{exp}(\frac{-t \lambda_t^2/4}{{4 \sigma^2  \log (2t/\delta)} + \lambda_t \sigma (\sqrt{2 \log (2t/\delta)} / 3}) \leq \delta/2
    \end{aligned} 
\end{equation}

Reversing the above inequality, we get

\begin{equation}
    \begin{aligned}
    \nonumber
t\lambda_t^2 \geq 16 \sigma^2\log (2t/\delta) + \frac{4}{3}\lambda_t \sigma \sqrt{2 \log (2t/\delta)}\log \left(\frac{2(d_1+d_2)}{\delta}\right)
    \end{aligned} 
\end{equation}

Therefore if we take $\lambda_t^2 \geq  \frac{128}{9t^2} \sigma^2  \log (2t/\delta)\log^2 \left(\frac{2(d_1+d_2)}{\delta}\right) + \frac{32}{t} \sigma^2\log (2t/\delta)$, then the above inequality holds. This claim is true because

\begin{equation}
    \begin{aligned}
    \nonumber
\frac{t\lambda_t^2}{2} \geq 16   \sigma^2\log (2t/\delta)  \quad \text{and} \quad \frac{t\lambda_t^2}{2} \geq \frac{4}{3}\lambda_t \sigma \sqrt{2\log (2t/\delta)}\log \left(\frac{2(d_1+d_2)}{\delta}\right)
    \end{aligned} 
\end{equation}

Note that $d_1 + d_2 \geq 2$ and hence $\log(2(d_1+d_2)/\delta)\geq 1$, and thus if we use 
\begin{equation}
    \begin{aligned}
    \nonumber\lambda_t^2 = \frac{48}{t}\sigma^2 \log(2t/\delta) \log^2(2(d_1+d_2)/\delta)    
    \end{aligned} 
\end{equation} 

then we have

\begin{equation}
\tag*{$\square$}
    \begin{aligned}
\mathbb{P} \left(\lambda_t \geq 2\vertiii{\frac{1}{t} \sum_{i=1}^t X_{i, a_i} \epsilon_i}_{op} \right)  
&= \mathbb{E}_{\{X_{i, a_i}\}_{i=1}^t} \left [\mathbb{P} \left(\lambda_t \geq 2\vertiii{\frac{1}{t} \sum_{i=1}^t X_{i, a_i} \epsilon_i}_{op} 
\Bigg \vert \{X_{i, a_i}\}_{i=1}^t   \right) \right] \\
&\geq 1 - (\frac{\delta}{2} + \frac{\delta}{2}) 
\geq 1 - \delta 
    \end{aligned} 
\end{equation}

\subsection{Proof of Lemma \ref{lem: matrix_RSC_condition}}

{
\begin{lemma}
\label{lem: matrix_RSC_condition}
Suppose Assumption \ref{ass: alpha_covariance} is satisfied, then with probability at least $1 -  \text{exp}\left(-\frac{T_0 C(d_1, d_2)}{4}\right) $, we have $B_{T_0}(\Theta, \Theta^*) \geq \frac{\alpha_0}{4} \vertiii{\Theta - \Theta^*}_F^2$, for all $T_0 \geq 2\log(d_1d_2^2+d_1d_2)/C(d_1, d_2)$ and $\Theta - \Theta^* \in \mathbb{C}$, where $C(d_1, d_2) > 0$ is a constant that depends on $d_1, d_2$.
\end{lemma}
}

First, we introduce the following two technical lemmas.

\begin{lemma}
\label{lem: matrix_RSC_difference}
Suppose that $\Sigma_1, \Sigma_2 \in \mathbb{R}^{d_1d_2 \times d_1d_2}$, $\Delta \in \mathbb{C}$, and the matrix $\Sigma_1$ satisfies the strongly convex condition $ vec(\Delta)^T \Sigma_1 vec(\Delta) \geq \alpha  \vertiii{\Delta}_{F}^2$  with $\alpha>0$. Moreover, suppose the two matrices are close enough such that $\vertiii{\Sigma_{2}-\Sigma_{1}}_{op} \leq \delta,$ where $2 \delta \leq \alpha .$ Then $$vec(\Delta)^T \Sigma_2 vec(\Delta) \geq \frac{\alpha}{2}  \vertiii{\Delta}_{F}^2 $$
\end{lemma}
\textbf{Proof}. The proof here is inspired by the proof of Lemma 6.17 in \citet{Buhlmann2011Statistics} and Lemma \ref{lem: lasso_RSC_difference} in Appendix \ref{sec: appendixLASSO}. A similar argument can be made using the nuclear norms $\vertiii{\Delta}_{nuc}$ since $\vertiii{\Delta}_{F} \leq \vertiii{\Delta}_{nuc}$. That is, the RSC condition can also be assumed in terms of the nuclear norm. Interested readers may take a deeper look at them by replicating all the lemmas in this section. Note that

\begin{equation}
    \begin{aligned}
    \nonumber
       \left | vec(\Delta)^T \Sigma_1 vec(\Delta) - vec(\Delta)^T \Sigma_2 vec(\Delta) \right |
      &= \left | vec(\Delta)^T (\Sigma_1 - \Sigma_2) vec(\Delta)\right |\\
      &\leq  \|(\Sigma_1 - \Sigma_2) vec(\Delta)\|_2 \|vec(\Delta)\|_2 \\
      &\leq  \vertiii{\Sigma_1 - \Sigma_2}_{op} \|vec(\Delta)\|_2^2 \\
      &= \vertiii{\Sigma_1 - \Sigma_2}_{op}  \vertiii{\Delta}_F^2 \leq \delta \vertiii{\Delta}_F^2\\
    \end{aligned}
\end{equation}

where the first inequality is by Cauchy-Schwarz and the second inequality is by the definition of the matrix induced norm $\| \cdot \|$. Therefore since we assume that $2\delta   \leq \alpha $, we know that

\begin{equation}
    \begin{aligned}
    \nonumber
       \left | vec(\Delta)^T \Sigma_1 vec(\Delta) - vec(\Delta)^T \Sigma_2 vec(\Delta) \right |  
       \leq \delta \vertiii{\Delta}_{F}^2 \leq  \frac{\delta}{\alpha } vec(\Delta)^T \Sigma_1 vec(\Delta) \leq  \frac{1}{2} vec(\Delta)^T \Sigma_1 vec(\Delta)\\
    \end{aligned}
\end{equation}

Therefore

\begin{equation}
\tag*{$\square$}
    \begin{aligned}
vec(\Delta)^T \Sigma_2 vec(\Delta)\geq  \frac{1}{2}  vec(\Delta)^T \Sigma_1 vec(\Delta) \geq \frac{\alpha}{2}  \vertiii{\Delta}_{F}^2 \\
    \end{aligned}
\end{equation}

\begin{lemma}
\label{lem: matrix_sigma_distance}
\textbf{\upshape (Distance Between Two Matrices)}
Define $C (d_1, d_2) = (\sqrt{ \frac{\alpha_0 }{4\sqrt{d_1 d_2}} + 1} - 1)^2$, where $\alpha_0$ is defined in Assumption \ref{ass: alpha_covariance}. Then for all $T_0 \geq 2 \log(d_1^2 d_2^2+d_1 d_2)/C (d_1, d_2)$,  we have
\begin{equation*}
    \mathbb{P}\left( \vertiii{\Sigma - \hat{\Sigma}_{T_0}}_{op} \geq \frac{\alpha_0}{2} \right)
    \leq \exp \left(-\frac{T_0 C(d_1, d_2)}{2}\right) 
\end{equation*}
\end{lemma}
 \textbf{Proof.} Since we sample uniformly in the exploration stage, we know that the contexts $\{X_{\tau, a_\tau}\}_{\tau=1}^{T_0}$ are i.i.d., and thus for any constant $w$,  by Lemma \ref{lem: T0_bernstein} we have

\begin{equation*}
    \mathbb{P}\left(\frac{1}{2}\vertiii{\hat{\Sigma}_{T_0} - \Sigma}_{\max} 
    \geq w + \sqrt{2w} + \sqrt{\frac{2\log(d_1^2 d_2^2+d_1 d_2)}{T_0}} +  {\frac{\log(d_1^2 d_2^2+d_1 d_2)}{T_0}}  \right) \leq \exp \left(-wT_0\right)
\end{equation*}

Note that by the property of matrix norms, we have $\vertiii{\Sigma - \hat{\Sigma}_{T_0}}_{2} \leq \sqrt{d_1d_2} \vertiii{\Sigma - \hat{\Sigma}_{T_0}}_{\max}$, therefore we get

\begin{equation*}
    \mathbb{P}\left(\frac{1}{2}\vertiii{\Sigma - \hat{\Sigma}_{T_0}}_{\max} 
    \geq \frac{\alpha_0}{4\sqrt{d_1d_2}}\right) 
    \geq \mathbb{P}\left(\frac{1}{2}\vertiii{\Sigma - \hat{\Sigma}_{T_0}}_{op} 
    \geq \frac{\alpha_0}{4} \right)
\end{equation*}

Now our choices of $w$ and $T_0$ is to let 

\begin{equation*}
w + \sqrt{2w} \leq \frac{\alpha_0}{8\sqrt{d_1 d_2}} 
\quad \text{ and } \quad
\sqrt{\frac{2\log(d_1^2 d_2^2+d_1 d_2)}{T_0}} +  {\frac{\log(d_1^2 d_2^2+d_1 d_2)}{T_0}}  \leq \frac{\alpha_0}{8\sqrt{d_1d_2}}
\end{equation*}

which leads to the following choices

\begin{equation*}
w = \frac{1}{2} (\sqrt{ \frac{\alpha_0}{4\sqrt{d_1 d_2}} + 1} - 1)^2 \quad  \text{ and } \quad 
T_0 \geq 2 \log(d_1^2 d_2^2+d_1 d_2) / (\sqrt{ \frac{\alpha_0 }{4\sqrt{d_1d_2}} + 1} - 1)^2
\end{equation*}

Then we have the following inequality

\begin{equation*}
\tag*{$\square$}
    \mathbb{P}\left(\frac{1}{2}\vertiii{\Sigma - \hat{\Sigma}_{T_0}}_{op} 
    \geq \frac{\alpha_0}{4} \right)
    \leq \mathbb{P}\left(\frac{1}{2}\vertiii{\Sigma - \hat{\Sigma}_{T_0}}_{\max} 
    \geq \frac{\alpha_0}{4\sqrt{d_1d_2}}\right) 
    \leq \text{exp} (- T_0 w) = \text{exp} (-\frac{T_0 C(d_1, d_2)}{2})
\end{equation*}

Finally, we provide the proof of Lemma \ref{lem: matrix_RSC_condition}.

\textbf{Proof}.
The proof follows from Lemma \ref{lem: matrix_RSC_difference} and Lemma \ref{lem: matrix_sigma_distance}. Since we have that the restricted eigenvalue condition holds for $\Sigma$ by Assumption \ref{ass: alpha_covariance} , i.e., 

\begin{equation}
    \begin{aligned}
    \nonumber
 vec(\Theta - \Theta^* )^T {\Sigma} ~ vec(\Theta - \Theta^* ) \geq \alpha_0 \|\Theta - \Theta^*\|_F^2, \forall \Theta - \Theta^* \in \mathbb{C}
    \end{aligned} 
\end{equation}

Also, the two matrices  $\Sigma$,  $\hat{\Sigma}_{T_0}$ are close enough when $T_0$ is large by Lemma \ref{lem: matrix_sigma_distance},

\begin{equation}
\nonumber
    \begin{aligned}
    \mathbb{P}\left( \vertiii{\Sigma - \hat{\Sigma}_{T_0}}_{op} \geq \frac{\alpha_0}{2} \right)
    \leq \exp \left(-\frac{T_0 C(d_1, d_2)}{2}\right) 
    \end{aligned}
\end{equation}
where $C(d_1, d_2)$ is the constant defined in  Lemma \ref{lem: matrix_sigma_distance}. By  Lemma \ref{lem: matrix_RSC_difference} we can claim that

\begin{equation}
    \begin{aligned}
    \nonumber
 vec(\Theta - \Theta^*)^T \hat{\Sigma}_{T_0}~ vec(\Theta - \Theta^*) \geq \frac{\alpha_0}{2} \|\Theta - \Theta^*\|_F^2, \forall \Theta - \Theta^* \in \mathbb{C}
    \end{aligned} 
\end{equation}

with high probability $1 - \text{exp}(-\frac{{T_0}C(d_1, d_2)}{2})$ when $T_0 \geq 2\log(d_1^2d_2^2+d_1d_2)/C(d_1, d_2)$. Therefore

\begin{equation}
\tag*{$\square$}
    \begin{aligned}
    B_{T_0}(\Theta, \Theta^*) = \frac{1}{2} (\Theta - \Theta^*)^T \hat{\Sigma}_{T_0} (\Theta - \Theta^*) \geq \frac{\alpha_0}{4} \|\Theta - \Theta^*\|_F^2, \forall \Theta - \Theta^* \in \mathbb{C}
    \end{aligned} 
\end{equation}

\subsection{Proof of Corollary \ref{cor: matrix_regret}}
Now given the probability of $\mathcal{A}_t$ and $\mathcal{E}_t$ in Lemma \ref{lem: matrix_lambda_condition} and Lemma \ref{lem: matrix_RSC_condition}, we can easily derive a regret upper bound for our low rank matrix bandit algorithm. We specify all the constants here, for example, the Lipschitz constants $C_1 = C_2 = 1$, the bound of the norm $k_1 = 1$, and we set $T_0 \geq 2\log(d_1^2 d_2^2+d_1 d_2)/C(d_1, d_2) $, where $C(d_1, d_2)$ is a constant defined in Lemma \ref{lem: matrix_sigma_distance}.  The restricted strong convexity holds with $\alpha = \alpha_0/4$.  The compatibility constant $\phi = \sqrt{2r}$. Also, $\Theta^*_{\mathcal{M}^{\bot}} = 0$ so $R(\Theta^*_{\mathcal{M}^{\bot}}) = 0$.

\textbf{Proof.} Using Theorem \ref{thm: regret_bound}, we can get the following upper bound

\begin{equation*}
    \begin{aligned}
       \mathbb{E}[ \text{Regret}(T) ]
        &\leq 2 T_0 + 2C_1 k_1 \sum_{t = T_0}^T \left[  \mathbb{P} \left( \mathcal{A}_{T_0}^c \right) + \mathbb{P}(\mathcal{E}_{T_0}^c)\right] +  2C_2 \sum_{ t = T_0}^T \sqrt{ \dfrac{144\lambda_{T_0}^2}{\alpha_0^2} \phi^2 + \dfrac{8\lambda_{T_0}}{\alpha_0} Z_{T_0} (\Theta^*)} \\
    \end{aligned} 
\end{equation*}

Since we take  $\delta = 1/T^2$ in Lemma \ref{lem: matrix_lambda_condition}, then $P(\mathcal{A}_t^c) \leq {1}/{t^2}$. Furthermore

\begin{equation*}
    \begin{aligned}
        \sum_{t = T_0}^T \mathbb{P} \left( \mathcal{A}_{T_0}^c \right)
        \leq  \sum_{t = T_0}^T \frac{1}{T^2}
        \leq \sum_{t = 1}^{T} \frac{1}{T^2} 
        \leq \frac{1}{T} 
    \end{aligned} 
\end{equation*}

Also by our choice of $T_0$, we know that

\begin{equation*}
    \begin{aligned}
        \sum_{t = T_0}^T \mathbb{P} \left( \mathcal{E}_{T_0}^c \right)
        \leq \sum_{t = T_0}^T  \text{ exp }( - \frac{T_0 C(d_1, d_2)}{2} )
        \leq T\text{ exp }( - \frac{T_0 C(d_1, d_2)}{2} ) = \mathcal{O}(1)
    \end{aligned} 
\end{equation*}

where the last equality is by the fact that $ x^{3/2} e^{-ax}$ is bounded above by a constant $(\frac{3}{2a})^3 e^{-\frac{9}{4a}}$ when $ a > 0$. The last summation term in the regret can be written as

\begin{equation*}
    \begin{aligned}
\sum_{ t = T_0}^T \sqrt{ \dfrac{144\lambda_{T_0}^2}{\alpha_0^2} \phi^2 + \dfrac{8\lambda_{T_0}}{\alpha_0}Z_{T_0}(\Theta^*)} 
& \leq \sum_{ t = T_0}^T  \frac{48 \sigma \sqrt{6r}}{\alpha_0 } \frac{\sqrt{\log(2T^3)} \log(2(d_1+d_2)T^2)}{\sqrt{T_0}} \\
& \leq \sum_{ t = 1}^T  \frac{48 \sigma \sqrt{6r}}{\alpha_0 }  \frac{\sqrt{\log(2T^3)} \log(2(d_1+d_2)T^2)}{\sqrt{T_0}} \\
& \leq  (T- T_0)\frac{1}{\sqrt{T_0}} \frac{48 \sigma \sqrt{6r}}{\alpha_0 } \sqrt{\log(2T^3)} \log(2(d_1+d_2)T^2)
    \end{aligned} 
\end{equation*}

Therefore we arrive to the following final regret bound

\begin{equation*}
    \begin{aligned}
       \mathbb{E}[ \text{Regret}(T) ]
        &\leq 2 T_0 + 2 \sum_{t = T_0}^T \left[  \mathbb{P} \left( \mathcal{A}_{T_0}^c \right) + \mathbb{P}(\mathcal{E}_{T_0}^c)\right] +  2\sum_{ t = T_0}^T \sqrt{ \dfrac{144\lambda_{T_0}^2}{\alpha_0^2} \phi^2 + \dfrac{8\lambda_{T_0}}{\alpha_0} Z_{T_0} (\Theta^*)} \\
        &\leq 2 T_0 + 2  \sum_{t = T_0}^T \left[  \mathbb{P} \left( \mathcal{A}_{T_0}^c \right) + \mathbb{P}(\mathcal{E}_{T_0}^c)\right] +  \frac{T}{\sqrt{T_0}} \frac{96 \sigma \sqrt{6r}}{\alpha_0 } \sqrt{\log(2T^3)} \log(2(d_1+d_2)T^2)  \\
    \end{aligned} 
\end{equation*}

Matching the two dominating terms, we get the choice of $T_0 = r^{1/3}(\frac{48\sqrt{6} \sigma}{\alpha_0})^{2/3} T^{2/3}$ and the regret upper bound

\begin{equation}
    \tag*{$\square$}
    \begin{aligned}
       \mathbb{E}[ \text{Regret}(T) ]
        & \leq  4 r^{1/3}(\frac{48\sqrt{6} \sigma}{\alpha_0})^{2/3} T^{2/3} \sqrt{2\log(2dT) }\log(2(d_1+d_2)T^2)  + 2 \sum_{t = T_0}^T \left[  \mathbb{P} \left( \mathcal{A}_{T_0}^c \right) + \mathbb{P}(\mathcal{E}_{T_0}^c)\right] \\
        & = \mathcal{O} \left ( r^{1/3} T^{2/3}\sqrt{\log(T)} \log((d_1+d_2)T) \right)\\
    \end{aligned} 
\end{equation}

\section{Proof for the Group-sparse Bandit}
\label{sec: appendixGroupSparse}

\subsection{Notations and Algorithm}

\begin{algorithm}[ht]
   \caption{The ESTC Algorithm for  Group-Sparse Matrix Bandit}
   \label{alg: group-sparse}
\begin{algorithmic}[1]
   \STATE \textbf{Input:} $\lambda_{T_0}, K \in \mathbb{N}, L_t(\theta), R(\theta), f(x, \theta), T_0$
   \FOR{$ t= 1$ to $T_0$}
   \STATE Observe $K$ contexts, $X_{t,1}, X_{t,2}, \cdots, X_{t, K}$ 
    \STATE Choose action $a_t$ uniformly randomly 
    \STATE Receive reward $y_{t} = \vertangl{ X_{t,a_t}, \Theta^* } + \epsilon_{t} $
        \ENDFOR
    \STATE Compute the estimator $\Theta_{T_0}$: 
      \qquad $$\Theta_{T_0} \in \text{argmin}_{\Theta \in \mathbf{\Theta}} \left \{\frac{1}{2T_0} \sum_{i=1}^{T_0}(y_{i}- \vertangl{X_{i, a_i}, \Theta} )^{2} + \lambda_t \vertiii{\Theta}_{1,q} \right \}$$
      \vspace{-10pt}
    \FOR{$ t=  T_0 +1$ to $T$}
      \STATE Choose action $a_t = \text{argmax}_a  \vertangl{ X_{t,a_t}, \Theta_{T_0} }$ 
   \ENDFOR
\end{algorithmic}
\end{algorithm}

We first clarify the notations we use through out this section. As we discuss in Section \ref{sec: applications}, $\Theta^* = [\theta^{(1)*}, \theta^{(2)*} \cdots, \theta^{(d_2)*}]$ is used to denote the matrix whose columns are vectors with similar supports, so that $|S(\Theta^*)| = s \ll d_1$.  Similarly, we use the shorthand vector notations $\mathcal{X}_t(\Theta^*), Y_t, e_t \in \mathbb{R}^t$ such that $[\mathcal{X}_t(\Theta^*)]_i = \vertangl{X_{i, a_i}, \Theta^*}$, $[Y_t]_i = y_i,$ and $[e_t]_i = \epsilon_i$. By our results in the low-rank matrix bandit problem, the derivative with respect to $\Theta^*$ is

\begin{equation}
    \begin{aligned}
    \nonumber
\nabla L_t(\Theta^*) 
&= \dfrac{\partial}{\partial \Theta^*} \frac{1}{2t}(Y_t - \mathcal{X}_t(\Theta^*))^T (Y_t - \mathcal{X}_t(\Theta^*)) \\
&= \frac{1}{2t} \dfrac{\partial}{\partial \Theta^*} ( - 2\mathcal{X}_t(\Theta^*)^T Y_t + \mathcal{X}_t(\Theta^*)^T \mathcal{X}_t(\Theta^*)) \\
&= - \frac{1}{t} \sum_{i=1}^t X_{i, a_i} \epsilon_i \\
    \end{aligned} 
\end{equation}

Therefore the event $\mathcal{A}_t$ is equivalent to
\begin{equation}
    \begin{aligned}
    \nonumber\left\{\lambda_t \geq 2\vertiii{\nabla L_t(\Theta^*)}_{\infty, q/(q-1)} = 2 \max_{i \in [d_1]}( \sum_{j=1}^{d_2}|\nabla L_t(\Theta^*)_{ij}|^{q/(q-1)})^{(q-1)/q}\right\}   
    \end{aligned}
\end{equation}

\subsection{Proof of Lemma \ref{lem: group_sparse_lambda_condition}}
\textbf{Proof}.
The proof here basically follows from Lemma 5 in \citet{Negahban2012A}, except for the function $\eta: \mathbb{N} \times \mathbb{N} \rightarrow \mathbb{N}$ which we introduce later. From the union bound, we know that for any constant $\lambda_t\in \mathbb{R}$, we have

\begin{equation}
    \begin{aligned}
    \nonumber
  \mathbb{P}\left(\lambda_t \geq 2\vertiii{\nabla L_t(\Theta^*)}_{\infty, q/(q-1)} \right)
  &=  \mathbb{P}\left(\forall i \in [d_1], \lambda_t \geq (\sum_{j=1}^{d_2}|\nabla L_t(\Theta^*)_{ij}|^{q/(q-1)})^{(q-1)/q} \right) \\
  &\geq 1 - \sum_{i=1}^{d_1} \mathbb{P} \left( \lambda_t \leq (\sum_{j=1}^{d_2}|\nabla L_t(\Theta^*)_{ij}|^{q/(q-1)})^{(q-1)/q} \right)\\
    \end{aligned} 
\end{equation}

Next, we establish a tail bound for the random variable $(\sum_{j=1}^{d_2}|\nabla L_t(\Theta^*)_{ij}|^{q/(q-1)})^{(q-1)/q}$. Let $X_{a_t}^{(i)}$ be the $i$-th row of $X_{a_t}$ and define $X = [X_{a_1}^{(i)T}, X_{a_2}^{(i)T}, \cdots, X_{a_t}^{(i)T}]$.

For any two $\sigma$-sub-Gaussian vectors $w, w'$, we have

\begin{equation}
    \begin{aligned}
    \nonumber
    \| \frac{1}{t} X w \|_{q/(q-1)} - \| \frac{1}{t} X w' \|_{q/(q-1)} | 
    &\leq \frac{1}{t}  \| X (w  - w') \|_{q/(q-1)}  \\
    &= \frac{1}{t} \sup_{\|\theta\|_q = 1} \langle X^T \theta, w - w' \rangle \\
    &\leq \frac{1}{t} \sup_{\|\theta\|_q = 1} \| X^T \theta \|_2 \|w - w' \|_2 \\
    \end{aligned}.
\end{equation}

Now if $q \in (1,2]$, we have 

$$\sup_{\|\theta\|_q = 1} \| X^T \theta \|_2 \leq  \sup_{\|\theta\|_2 = 1} \| X^T \theta \|_2 \leq \| X^T\|_F \leq \sqrt{t}$$

If $q > 2$, then we have a different inequality 

$$\sup_{\|\theta\|_q = 1} \| X^T \theta \|_2 \leq d_2^{1/2 - 1/q} \sup_{\|\theta\|_2 = 1} \| X^T \theta \|_F \leq d_2^{1/2 - 1/q}\sqrt{t}$$

Therefore if we define $\eta(d_2, m) = \max\{1,  d_2^{m}\}$, we have $|\| \frac{1}{t} X w \|_{q/(q-1)} - \| \frac{1}{t} X w' \|_{q/(q-1)} | \leq \frac{\eta(d_2, 1/2 - 1/q)}{\sqrt{t}} \|w - w' \|_2$. Thus the function is Lipschitz with constant $\eta(d_2, 1/2 - 1/q)/\sqrt{t}$. Based on the concentration inequality of measure for Lipschitz functions \citep{bobkov2015concentration, Negahban2012A}, we know that

\begin{equation}
    \begin{aligned}
    \nonumber
    \mathbb{P} \left ( \frac{1}{t} \| X e_t \|_{q/(q-1)} \geq \mathbb{E}[\frac{1}{t} \|X e_t \|_{q/(q-1)}] + \sigma \delta \right) \leq 2\text{exp} (- \frac{t\delta^2}{2\eta(d_2, 1/2 - 1/q)^2})
    \end{aligned}.
\end{equation}

By Lemma 5 in \citet{Negahban2012A}, we know that the mean is bounded by $2d_2^{1-\frac{1}{q}}\sigma/\sqrt{t}$, therefore

\begin{equation}
    \begin{aligned}
    \nonumber
    \mathbb{P} \left ( \frac{1}{t} \| X e_t \|_{q/(q-1)} \geq 2 \sigma \frac{d_2^{1-1/q}}{\sqrt{t}} + \sigma \eta(d_2, \frac{1}{2} - \frac{1}{q})\delta \right) \leq 2\text{exp} (- \frac{t\delta^2}{2})
    \end{aligned}.
\end{equation}

By change of variables and the union bound, we can get the claim in the lemma. \hfill $\square$.

\subsection{ Proof of Corollary \ref{cor: group_sparse_regret}}
Now given the probability of $\mathcal{A}_t$ and $\mathcal{E}_t$ in Lemma \ref{lem: group_sparse_lambda_condition} and Lemma \ref{lem: matrix_RSC_condition}, we can derive a regret upper bound for our group-sparse matrix bandit algorithm. Similarly, We specify all the constants here, for example, the Lipschitz constants $C_1 = C_2 = 1$, the bound of the norm $k_1 = 1$, and we set $T_0 \geq 2\log(d_1^2 d_2^2+d_1 d_2)/C(d_1, d_2) $, where $C(d_1, d_2)$ is the constant defined in Lemma \ref{lem: matrix_sigma_distance}.  The restricted strong convexity holds with $\alpha = \alpha_0/4$. The compatibility constant $\phi = \eta(d_2, {1/q - 1/2}) \sqrt{s}$.  Also, $\Theta^*_{\mathcal{M}^{\bot}} = 0$ so $R(\Theta^*_{\mathcal{M}^{\bot}}) = 0$.

\textbf{Proof.} Using Theorem \ref{thm: regret_bound}, we can get the following upper bound

\begin{equation*}
    \begin{aligned}
       \mathbb{E}[ \text{Regret}(T) ]
        &\leq 2 T_0 + 2 \sum_{t = T_0}^T \left[  \mathbb{P} \left( \mathcal{A}_{T_0}^c \right) + \mathbb{P}(\mathcal{E}_{T_0}^c)\right] +  2\sum_{ t = T_0}^T \sqrt{ \dfrac{144\lambda_{T_0}^2}{\alpha_0^2} \phi^2 + \dfrac{8\lambda_{T_0}}{\alpha_0}Z_{T_0}(\Theta^*)} \\
    \end{aligned} 
\end{equation*}

Since we take $\delta = 1/{T}^2$ in Lemma \ref{lem: group_sparse_lambda_condition}, $P(\mathcal{A}_{T_0}^c) \leq {1}/{T^2}$. Furthermore,

\begin{equation*}
    \begin{aligned}
        \sum_{t = T_0}^T \mathbb{P} \left( \mathcal{A}_{T_0}^c \right)
        \leq  \sum_{t = T_0}^T \frac{1}{T^2}
        \leq \sum_{t = 1}^{T} \frac{1}{T^2} 
        \leq 1/T
    \end{aligned} 
\end{equation*}

Also by our choice of $T_0 = \Theta(s^{1/3} T^{2/3})$, we know that

\begin{equation*}
    \begin{aligned}
        \sum_{t = T_0}^T \mathbb{P} \left( \mathcal{E}_{T_0}^c \right)
        \leq \sum_{t = T_0}^T  \text{ exp }( - \frac{T_0 C(d_1, d_2)}{2} )
        \leq T\text{ exp }( - \frac{T_0 C(d_1, d_2)}{2} ) = \mathcal{O}(1)
    \end{aligned} 
\end{equation*}

where the last equality is by the fact that $ x^{3/2} e^{-ax}$ is bounded above by a constant $(\frac{3}{2a})^3 e^{-\frac{9}{4a}}$ when $ a > 0$. The last summation term in the regret can be written as

\begin{equation*}
    \begin{aligned}
&\sum_{ t = T_0}^T \sqrt{ \dfrac{144\lambda_{T_0}^2}{\alpha_0^2} \phi^2 + \dfrac{8\lambda_{T_0}}{\alpha_0}Z_t(\Theta^*)} \\
& \leq \sum_{ t = T_0}^T  (\frac{24 \sigma \sqrt{s} \eta(d_2, \frac{1}{q} - \frac{1}{2}) d_2^{1-\frac{1}{q}}}{\alpha_0})\frac{1}{\sqrt{{T_0}}} 
 +  \sum_{ t = T_0}^T  \frac{12 \sigma \sqrt{s}\eta(d_2, \frac{1}{q} - \frac{1}{2})\eta(d_2, \frac{1}{2} - \frac{1}{q})}{\alpha_0}\frac{\sqrt{2\log(2d_1T^2)}}{\sqrt{{T_0}}} \\
&\leq \frac{T - T_0}{\sqrt{T_0}} \left(\frac{24 \sigma \sqrt{s} \eta(d_2, {\frac{1}{q} - \frac{1}{2}})d_2^{1-\frac{1}{q}} }{\alpha_0} +  \frac{12 \sqrt{s} \sigma  \eta(d_2, \frac{1}{q} - \frac{1}{2})\eta(d_2, \frac{1}{2} - \frac{1}{q})}{\alpha_0} \sqrt{2\log(2d_1T^2)} \right)
    \end{aligned} 
\end{equation*}

Therefore we arrive to the following bound

\begin{equation*}
    \begin{aligned}
       \mathbb{E}[ \text{Regret}(T) ]
        &\leq 2 T_0 + 2\sum_{t = T_0}^T \left[  \mathbb{P} \left( \mathcal{A}_{T_0}^c \right) + \mathbb{P}(\mathcal{E}_{T_0}^c)\right] +  2\sum_{ t = T_0}^T \sqrt{ \dfrac{144\lambda_{T_0}^2}{\alpha_0^2} \phi^2 + \dfrac{8\lambda_{T_0}}{\alpha_0}Z_{T_0}(\Theta^*)} \\
        &\leq 2 T_0 + 2 \sum_{t = T_0}^T \left[  \mathbb{P} \left( \mathcal{A}_{T_0}^c \right) + \mathbb{P}(\mathcal{E}_{T_0}^c)\right] \\
        & \quad + \frac{T}{\sqrt{T_0}} \left(\frac{24\sqrt{s}\sigma \eta(d_2, {\frac{1}{q} - \frac{1}{2}})d_2^{1-\frac{1}{q}} }{\alpha_0} +  \frac{12 \sqrt{s} \sigma  \eta(d_2, \frac{1}{q} - \frac{1}{2})\eta(d_2, \frac{1}{2} - \frac{1}{q})}{\alpha_0} \sqrt{2\log(2d_1T^2)} \right) \\
    \end{aligned} 
\end{equation*}

Matching the two dominating terms, we get the choice of 

\begin{equation}
T_0 = s^{1/3} \left(\frac{12 \sigma \eta(d_2, {\frac{1}{q} - \frac{1}{2}})d_2^{1-\frac{1}{q}} }{\alpha_0} +  \frac{6 \sigma  \eta(d_2, \frac{1}{q} - \frac{1}{2})\eta(d_2, \frac{1}{2} - \frac{1}{q})}{\alpha_0}  \sqrt{2\log(2d_1T^2)}   \right)^{2/3} T^{2/3}
\end{equation}
and the following regret upper bound

\begin{equation*}
    \begin{aligned}
       \mathbb{E}[ \text{Regret}(T) ]
       & \leq  2 \sum_{t = T_0}^T \left[  \mathbb{P} \left( \mathcal{A}_{T_0}^c \right) + \mathbb{P}(\mathcal{E}_{T_0}^c)\right] \\
        & \quad + 4 s^{1/3} \left(\frac{12\sqrt{s}\sigma \eta(d_2, {\frac{1}{q} - \frac{1}{2}})d_2^{1-\frac{1}{q}} }{\alpha_0} +  \frac{6 \sqrt{s} \sigma  \eta(d_2, \frac{1}{q} - \frac{1}{2})\eta(d_2, \frac{1}{2} - \frac{1}{q})}{\alpha_0} \sqrt{2\log(2d_1T^2)}  \right)^{2/3} T^{2/3}  \\
        & = \mathcal{O}\left( s^{1/3}\left(C_1(d_2) + C_2(d_2) \right)^{2/3} T^{2/3}  \left( \log(d_1T^2)\right)^{1/3}  \right)
    \end{aligned} 
\end{equation*}

where $C_1(d_2) = \eta(d_2, {\frac{1}{q} - \frac{1}{2}})d_2^{1-\frac{1}{q}}$ and $C_2(d_2) = \eta(d_2, {\frac{1}{q} - \frac{1}{2}})\eta(d_2, \frac{1}{2} - \frac{1}{q})$. \hfill $\square$

\section{Proof for the Multi-agent LASSO}
\label{sec: appendixMultiAgent}

\subsection{Notations and Algorithm}

\begin{algorithm}[ht]
   \caption{The ESTC Algorithm for Multiple Agent LASSO Bandit}
   \label{alg: multi-agent}
\begin{algorithmic}[1]
 \STATE \textbf{Input:} $\{\lambda_t\}_{t=1}^T, K, d_1, d_2 \in \mathbb{N}$, $L_t(\theta), R(\theta), f(x, \theta), T_0$
   \FOR{$ t= 1$ to $T_0$}
   \STATE Observe $K$ contexts, $x_{t,1}, x_{t,2}, \cdots, x_{t, K}$
    \STATE Each Agent $k$ choose action $a_t^{(k)}$ uniformly randomly
    \STATE Each Agent receives reward $y_{t}^{(k)} = f(x_{t, a_t}^{(k)}, \theta^{(k)*}) + \epsilon_{t}^{(k)}$
        \ENDFOR
    \STATE Compute the estimator $\Theta_{T_0} = (\theta_{T_0}^{(1)}, \theta_{T_0}^{(2)}, \cdots, \theta_{T_0}^{(d_2)})$: 
      \qquad $$\Theta_{T_0} \in \text{argmin}_{\Theta \in \mathbf{\Theta}} \{\frac{1}{2T_0} \sum_{i=1}^{T_0} \sum_{k=1}^{d_2} (y_{i}^{(k)}- x_{i,a_i}^{(k)T}\theta^{(k)})^2+ \lambda_t \vertiii{\Theta}_{1,q}\}$$
      \vspace{-10pt}
    \FOR{$ t=  T_0 +1$ to $T$}
      \STATE Each agent $k$ choose action $a_t^{(k)} = \text{argmax}_a f(x_{t,a}^{(k)}, \theta_{T_0}^{(k)})$
   \ENDFOR
\end{algorithmic}
\end{algorithm}

Define $\Theta^* = [\theta^{(1)*}, \theta^{(2)*} \cdots, \theta^{(d_2)*}] \in \mathbb{R}^{d1\times d_2}$ and $S(\Theta^*) = \{i \in [d_1] \mid \Theta^{*}_i = 0\}$ as the set of zero rows, then by our setting $|S(\Theta^*)| = s \ll d_1$. Let the loss be the sum of squared error function and the regularization function be the $l_{1,q}$ norm. More formally,
\begin{equation}
    \nonumber
    L_t(\Theta) =  \frac{1}{2t} \sum_{i=1}^{t} \sum_{k=1}^{d_2} (y_{i}^{(k)}- x_{i,a_i}^{(k)T}\theta^{(k)*})^2, R(\Theta) = \vertiii{\Theta}_{1,q}
\end{equation}
Define $(\mathcal{M}, \overline{\mathcal{M}}^\bot)$ as in Example \ref{example: group_sparse}, then the $l_{1,q}$ norm is decomposable and $\Theta^*_{\overline{\mathcal{M}}^\bot}= 0$. For each agent $j$, we use the notations $X_t^{(j)} \in \mathbb{R}^{t \times d_1}, Y_t^{(j)}$, $e_t^{(j)} \in \mathbb{R}^{t}$ to represent the context matrix, the reward and the error vectors, i.e., $[X_t^{(j)}]_i = x_{i, a_i}^{(j)}, [Y_t^{(j)}]_i = y_{i}^{(j)}, [e_t^{(j)}]_i = \epsilon_i^{(j)}, \forall i \in [t], j \in [d_2]$.  The derivative with respect to $\Theta^*$ can be computed as 

\begin{equation}
    \begin{aligned}
    \nonumber
    \nabla_{\Theta^*} L_t(\Theta^*) =  \frac{1}{2t} \sum_{i=1}^{t} \sum_{k=1}^{d_2} \nabla_{\Theta^*} (y_{i}^{(k)}- x_{i,a_i}^{(k)T}\theta^{(k)*})^2
    \end{aligned} 
\end{equation}

Note that if we take partial derivatives, we get

\begin{equation}
    \begin{aligned}
    \nonumber
        \frac{\partial L_t(\Theta^*)}{\partial \theta^{(j)*}} &= \frac{1}{2t} \sum_{i=1}^{t} \sum_{k=1}^{d_2} \frac{\partial }{\partial \theta^{(j)*}} (y_{i}^{(k)}- x_{i,a_i}^{(k)T}\theta^{(k)*})^2 \\
        &= \frac{1}{2t} \sum_{i=1}^{t}  \frac{\partial }{\partial \theta^{(j)*}} (y_{i}^{(j)}- x_{i,a_i}^{(j)T}\theta^{(j)*})^2 \\
        &= -\frac{1}{t} X_{t}^{(j)T} e_t^{(j)}
    \end{aligned} 
\end{equation}

Therefore $\nabla L_t(\Theta^*) = -\frac{1}{t}[X_{t}^{(1)T} e_t^{(1)}, X_{t}^{(2)T} e_t^{(2)}, \cdots, X_{t}^{(d_2)T} e_t^{(d_2)}]$. Now we can compute the Bregman divergence as follows.

\begin{equation}
    \begin{aligned}
    \nonumber
    B_t(\Theta, \Theta^*)  
    &= L_t(\Theta) - L_t(\Theta^*) - \langle \nabla L_t(\Theta^*), \Theta - \Theta^* \rangle \\
    &= \frac{1}{2t} \sum_{k=1}^{d_2}\|Y_t^{(k)} - X_t^{(k)}\theta^{(k)}\|_2^2 -  \frac{1}{2t}\sum_{k=1}^{d_2}\|Y_t^{(k)} - X_t^{(k)}\theta^{(k)*}\|_2^2 + \frac{1}{t} \sum_{k=1}^{d_2} e_t^{(k)T}X_t^{(k)}(\theta^{(k)} - \theta^{(k)*}) \\
    &= \frac{1}{2t} \sum_{k=1}^{d_2} (\theta^{(k)} - \theta^{(k)*})^TX_t^{(k)T}X_t^{(k)}(\theta^{(k)} - \theta^{(k)*})
    \end{aligned}
\end{equation}

Therefore the event $\mathcal{A}_t$ is equivalent to
\begin{equation}
    \begin{aligned}
    \nonumber\left\{\lambda_t \geq 2\vertiii{\nabla L_t(\Theta^*)}_{\infty, q/(q-1)} = 2 \max_{i \in [d_1]}( \sum_{j=1}^{d_2}|\nabla L_t(\Theta^*)_{ij}|^{q/(q-1)})^{(q-1)/q}\right\}   
    \end{aligned}
\end{equation}

The event $\mathcal{E}_t$ (RSC condition) is equivalent to

\begin{equation}
    \begin{aligned}
    \nonumber
    \left\{ \frac{1}{2t} \sum_{k=1}^{d_2} (\theta^{(k)} - \theta^{(k)*})^TX_t^{(k)T}X_t^{(k)}(\theta^{(k)} - \theta^{(k)*}) \geq \alpha \vertiii{\Theta - \Theta^*}_{F}^2 - Z_t(\Theta^*)\right\}
    \end{aligned}
\end{equation}

\subsection{Useful Lemmas}

\begin{lemma}
\label{lem: multi_agent_lambda_condition} \textbf{\upshape (Good Choice of Lambda)}
Suppose that $\epsilon_t$ is $\sigma$-sub-Gaussian. For any $\delta \in (0,1)$, use 
$$\lambda_t =   2 \sigma \frac{d_2^{1-1/q}}{\sqrt{t}} + \sigma \eta(d_2, \frac{1}{2} - \frac{1}{q}) \sqrt{\frac{2 \log ({2d_1}/{\delta})}{t}}$$ 
at each  round $t$ in Algorithm \ref{alg: multi-agent}, then with probability at least $1-\delta$, we have $\lambda_t \geq R^*(\nabla L_t(\Theta^*))$.
\end{lemma}

\textbf{Proof}. From the union bound, we know that for any constant $\lambda_t\in \mathbb{R}$, we have

\begin{equation}
    \begin{aligned}
    \nonumber
  \mathbb{P}\left(\lambda_t \geq 2\vertiii{\nabla L_t(\Theta^*)}_{\infty, q/(q-1)} \right) 
  &=  \mathbb{P}\left(\forall i \in [d_1], \lambda_t \geq (\sum_{j=1}^{d_2}|\nabla L_t(\Theta^*)_{ij}|^{q/(q-1)})^{(q-1)/q} \right) \\
  &\geq 1 - \sum_{i=1}^{d_1} \mathbb{P} \left( \lambda_t \leq (\sum_{j=1}^{d_2}|\nabla L_t(\Theta^*)_{ij}|^{q/(q-1)})^{(q-1)/q} \right)\\
    \end{aligned} 
\end{equation}

Note that the $j$-th row of $\nabla L_t(\Theta^*)$ consists of $\left[ \sum_i ( x_{i,a_i}^{(1)})_j \epsilon_i^{(1)}, \sum_i ( x_{i,a_i}^{(2)})_j \epsilon_i^{(2)}, \cdots, \sum_i ( x_{i,a_i}^{(d_2)})_j \epsilon_i^{(2)} \right]$, which are all $\sqrt{t}\sigma$-sub-Gaussian random variables.

For any two $\sqrt{t}\sigma$-sub-Gaussian vectors $w, w'$, we have

\begin{equation}
    \begin{aligned}
    \nonumber
    |\| \frac{1}{t} w \|_{q/(q-1)} - \| \frac{1}{t} w' \|_{q/(q-1)} | 
    &\leq \frac{1}{t}  \| (w  - w') \|_{q/(q-1)} \\
    &= \frac{1}{t} \sup_{\|\theta\|_q = 1} \langle \theta, w - w' \rangle \\
    &\leq \frac{1}{t} \sup_{\|\theta\|_q = 1} \| \theta \|_2 \|w - w' \|_2 \\
    \end{aligned}.
\end{equation}

Now if $q \in (1,2]$, we have $\|\theta\|_2 \leq \|\theta\|_q = 1$. If $q > 2$, then we have a different inequality $\|\theta\|_2 \leq d_2^{1/2-1/q} \|\theta\|_q \leq d_2^{1/2-1/q}$. Therefore if we define $\eta(d_2, m) = \max\{1,  d_2^{m}\}$, we have $|\| \frac{1}{t} w \|_{q/(q-1)} - \| \frac{1}{t} w' \|_{q/(q-1)} | \leq \frac{\eta(d_2, 1/2 - 1/q)}{t} \|w - w' \|_2$. Thus the function is Lipschitz with constant $\eta(d_2, 1/2 - 1/q)/t$. Based on the concentration inequality of measure for Lipschitz functions \citep{bobkov2015concentration, Negahban2012A}, we know that

\begin{equation}
    \begin{aligned}
    \nonumber
    \mathbb{P} \left ((\sum_{j=1}^{d_2}|\nabla L_t(\Theta^*)_{ij}|^{q/(q-1)})^{(q-1)/q} \geq \mathbb{E}[(\sum_{j=1}^{d_2}|\nabla L_t(\Theta^*)_{ij}|^{q/(q-1)})^{(q-1)/q} ] + \delta \right) \leq 2\text{exp} (- \frac{t^2\delta^2}{2\eta(d_2, 1/2 - 1/q)^2 t \sigma^2})
    \end{aligned}.
\end{equation}

By similar arguments as in the proof of Lemma \ref{lem: group_sparse_lambda_condition}, we know that 

\begin{equation}
    \begin{aligned}
    \nonumber
    \mathbb{P} \left ( (\sum_{j=1}^{d_2}|\nabla L_t(\Theta^*)_{ij}|^{q/(q-1)})^{(q-1)/q}  \geq 2 \sigma \frac{d_2^{1-1/q}}{\sqrt{t}} + \sigma \eta(d_2, \frac{1}{2} - \frac{1}{q})\delta \right) \leq 2\text{exp} (- \frac{t\delta^2}{2})
    \end{aligned}.
\end{equation}

By change of variables and the union bound, we can get the results in the lemma. \hfill $\square$.

{
\begin{lemma}\textbf{\upshape (RSC condition)}
\label{lem: multi_agent_RSC_condition}
Suppose Assumption \ref{ass: alpha_covariance} is satisfied, then with probability at least $1 -  d_2\text{exp}\left(-\frac{T_0 C(s)}{2}\right) $, we have $B_{T_0}(\Theta, \Theta^*) \geq \frac{\alpha_0}{4} \vertiii{\Theta - \Theta^*}_F^2 $, for all $T_0 \geq 2\log(2d_1^2 + d_1) / C(s)$ and $\Theta - \Theta^* \in \mathbb{C}$, where $C(s)$ is a constant defined in Lemma \ref{lem: lasso_sigma_distance}.
\end{lemma}
}
\textbf{Proof.}
Note that the RSC condition (event $\mathcal{E}_t$ ) is equivalent to 

\begin{equation}
    \begin{aligned}
    \nonumber
    \frac{1}{2t} \sum_{k=1}^{d_2} (\theta^{(k)} - \theta^{(k)*})^TX_t^{(k)T}X_t^{(k)}(\theta^{(k)} - \theta^{(k)*}) 
    & \geq \alpha \vertiii{\Theta - \Theta^*}_{F}^2 - Z_t(\Theta^*) \\
    &
     = \sum_{k=1}^{d_2} \alpha \|\theta^{(k)} - \theta^{(k)*}\|_{2}^2 - Z_t(\Theta^*) \\
    \end{aligned}
\end{equation}

By the results in Lemma \ref{lem: lasso_RSC_condition} in the LASSO bandit problem, we know that with high probability $1 - \text{exp}(-\frac{{T_0}C(s)}{2})$ 

\begin{equation}
\nonumber
    \begin{aligned}
    \frac{1}{2} (\theta^{(k)} - \theta^{(k)*})^T \hat{\Sigma}_{T_0}^{(k)} (\theta^{(k)} - \theta^{(k)*}) \geq \frac{\alpha_0}{4} \|\theta^{(k)} - \theta^{(k)*}\|_2^2
    \end{aligned} 
\end{equation}
when $T_0 \geq 2\log(d_1^2+d_1)/C(s)$, where $C(s)$ is defined in Lemma \ref{lem: lasso_sigma_distance}. Therefore by the Frechet inequality, we know that

\begin{equation}
    \tag*{$\square$}
    \begin{aligned}
    &P \left(\sum_{k=1}^{d_2} \frac{1}{2t} (\theta^{(k)} - \theta^{(k)*})^TX_{T_0}^{(k)T}X_{T_0}^{(k)}(\theta^{(k)} - \theta^{(k)*}) \geq \sum_{k=1}^{d_2} \frac{\alpha_0}{2} \|\theta^{(k)} - \theta^{(k)*}\|_{2}^2 \right) \\
    &\geq P \left(\frac{1}{2t} (\theta^{(k)} - \theta^{(k)*})^TX_{T_0}^{(k)T}X_{T_0}^{(k)}(\theta^{(k)} - \theta^{(k)*}) \geq \frac{\alpha_0}{2} \|\theta^{(k)} - \theta^{(k)*}\|_{2}^2, \forall k\in [d_2] \right) \\
    & \geq \sum_{k=1}^{d_2} P \left(\frac{1}{2t} (\theta^{(k)} - \theta^{(k)*})^TX_{T_0}^{(k)T}X_{T_0}^{(k)}(\theta^{(k)} - \theta^{(k)*}) \geq \frac{\alpha_0}{2} \|\theta^{(k)} - \theta^{(k)*}\|_{2}^2 \right) - (d_2-1) \\
    &\geq 1 -  d_2\text{exp}\left(-\frac{T_0 C(s)}{2}\right) 
    \end{aligned}
\end{equation}

\subsection{Proof of Theorem \ref{thm: regret_bound_multi_agent}}

{
\textbf{Proof}.
By the boundedness of $\|x\|_2$ and $\|\theta\|_2$, we know that

\begin{equation}
    \begin{aligned}
    \nonumber
        f(x_{t, a_t^*}^{(k)}, \theta^{(k)*}) - f(x_{t, a_t}^{(k)}, \theta^{(k)*}) 
    = x_{t, a_t^*}^{(k)T}\theta^{(k)*} - x_{t, a_t}^{(k)T}\theta^{(k)*} \leq (x_{t, a_t^*}^{(k)T}  - x_{t, a_t}^{(k)T}) \theta^{(k)*} \leq 2
    \end{aligned} 
\end{equation}

Then we can decompose the one-step regret from round $t$ across different agents into three parts as follows

\begin{equation}
    \begin{aligned}
    \nonumber
        &  \sum_{k=1}^{d_2} f(x_{t, a_t^*}^{(k)}, \theta^{(k)*}) - f(x_{t, a_t}^{(k)}, \theta^{(k)*}) \\
                      &=  \left [\sum_{k=1}^{d_2} f(x_{t, a_t^*}^{(k)}, \theta^{(k)*}) - f(x_{t, a_t}^{(k)}, \theta^{(k)*})  \right] \mathbb{I}(t \leq T_0) + \left [\sum_{k=1}^{d_2} f(x_{t, a_t^*}^{(k)}, \theta^{(k)*}) - f(x_{t, a_t}^{(k)}, \theta^{(k)*}) \right] \mathbb{I}(t > T_0, \mathcal{E}_{T_0}) \\
                     & \qquad  + \left [\sum_{k=1}^{d_2} f(x_{t, a_t^*}^{(k)}, \theta^{(k)*}) - f(x_{t, a_t}^{(k)}, \theta^{(k)*})  \right] \mathbb{I}(t \geq T_0, \mathcal{E}_{T_0}^c) \\
                     &\leq 2d_2 \mathbb{I}(t \leq T_0) +   \left [\sum_{k=1}^{d_2} f(x_{t, a_t^*}^{(k)}, \theta^{(k)*}) - f(x_{t, a_t}^{(k)}, \theta^{(k)*}) \right]  \mathbb{I}(t >  T_0, \mathcal{E}_{T_0}) + 2d_2\mathbb{I}(t > T_0, \mathcal{E}_{T_0}^c) \\
                    &= 2d_2 \mathbb{I}(t \leq T_0) +   \left [\sum_{k=1}^{d_2} f(x_{t, a_t^*}^{(k)}, \theta^{(k)*}) - f(x_{t, a_t}^{(k)}, \theta^{(k)*}) \right]  \mathbb{I}(t >  T_0, f(x_{t, a_t}^{(k)}, \theta_{T_0}^{(k)}) \geq f(x_{t, a_t^*}^{(k)}, \theta_{T_0}^{(k)}), \forall k \in [d_2],  \mathcal{E}_{T_0})\\
                    & \quad + 2d_2\mathbb{I}(t > T_0, \mathcal{E}_{T_0}^c)
    \end{aligned} 
\end{equation}

where $\mathbb{I}(\cdot)$ is the indicator function. The last equality is due to the choice of $a_t^{(k)} =  \text{argmax}_a f(x_{t,a}^{(k)}, \theta_{T_0}^{(k)})$, and thus we know that $f(x_{t,a}^{(k)}, \theta_{T_0}^{(k)}) \geq  f(x_{t,a_t^*}, \theta_{T_0}^{(k)})$. We focus on the second indicator function now. By the Lipschitzness of $f$ over $\theta$, we have

\begin{equation}
    \begin{aligned}
    \nonumber
        &\mathbb{I}\left(t >  T_0, \mathcal{E}_{T_0}\right)  \\
        &= \mathbb{I}\left(t >  T_0,  f(x_{t, a_t}^{(k)}, \theta_{T_0}^{(k)}) \geq f(x_{t, a_t^*}^{(k)}, \theta_{T_0}^{(k)}), \forall k \in [d_2], \mathcal{E}_{T_0}\right)  \\
        &= \mathbb{I}\left(t >  T_0,  f(x_{t, a_t}^{(k)}, \theta_{T_0}^{(k)}) - f(x_{t, a_t^*}^{(k)}, \theta_{T_0}^{(k)}) +  f(x_{t, a_t^*}^{(k)}, \theta_{T_0}^{(k)*}) -  f(x_{t, a_t}^{(k)}, \theta_{T_0}^{(k)*}) \right. \\
        & \qquad \left. \geq f(x_{t, a_t^*}^{(k)}, \theta_{T_0}^{(k)*}) -  f(x_{t, a_t}^{(k)}, \theta_{T_0}^{(k)*}) , \forall k, \mathcal{E}_{T_0}\right) \\
        &= \mathbb{I}\left(t >  T_0,  [f(x_{t, a_t}^{(k)}, \theta_{T_0}^{(k)}) -  f(x_{t, a_t}^{(k)}, \theta_{T_0}^{(k)*}) ]+  [f(x_{t, a_t^*}^{(k)}, \theta_{T_0}^{(k)*}) - f(x_{t, a_t^*}^{(k)}, \theta_{T_0}^{(k)})]  \geq \right. \\
        & \left. \qquad f(x_{t, a_t^*}^{(k)}, \theta_{T_0}^{(k)*}) -  f(x_{t, a_t}^{(k)}, \theta_{T_0}^{(k)*}) , \forall k \in [d_2], \mathcal{E}_{T_0}\right) \\
        & \leq \mathbb{I}\left(t >  T_0, 2  \|\theta_{T_0}^{(k)} - \theta^{(k)*}\|_2 \geq f(x_{t, a_t^*}^{(k)}, \theta_{T_0}^{(k)*}) -  f(x_{t, a_t}^{(k)}, \theta_{T_0}^{(k)*}), \forall k \in [d_2], \mathcal{E}_{T_0} \right) \\
        & \leq \mathbb{I}\left(t >  T_0, 2\sum_{k=1}^{d_2} \|\theta_{T_0}^{(k)} - \theta^{(k)*}\|_2 \geq \sum_{k=1}^{d_2}  f(x_{t, a_t^*}^{(k)}, \theta_{T_0}^{(k)*}) -  f(x_{t, a_t}^{(k)}, \theta_{T_0}^{(k)*}), \mathcal{E}_{T_0} \right) \\
        & \leq \mathbb{I}\left(t >  T_0, 2 \sqrt{d_2} \vertiii{\Theta_{T_0} - \Theta^{*}}_F \geq \sum_{k=1}^{d_2}  f(x_{t, a_t^*}^{(k)}, \theta_{T_0}^{(k)*}) -  f(x_{t, a_t}^{(k)}, \theta_{T_0}^{(k)*}), \mathcal{E}_{T_0} \right) \\
    \end{aligned} 
\end{equation}

where the last inequality is by the Cauchy-Schwarz Inequality. Substitute the above inequality back and take expectation on both sides of the one-step regret from round $t$, we get

\begin{equation}
    \begin{aligned}
    \nonumber
        \mathbb{E}\left[ \sum_{k=1}^{d_2} f(x_{t, a_t^*}^{(k)}, \theta^{(k)*}) - f(x_{t, a_t}^{(k)}, \theta^{(k)*}) \right] \leq 2d_2 \text{ \hfill  for }t \leq T_0
    \end{aligned} 
\end{equation}

For $t > T_0$ and any constant $v \in \mathbb{R}$, the expectation is bounded by

\begin{equation}
    \begin{aligned}
    \nonumber
        &  \mathbb{E}\left[ \sum_{k=1}^{d_2} f(x_{t, a_t^*}^{(k)}, \theta^{(k)*}) - f(x_{t, a_t}^{(k)}, \theta^{(k)*}) \right] \\
        &\leq \mathbb{E} \left [ \left(\sum_{k=1}^{d_2} f(x_{t, a_t^*}^{(k)}, \theta^{(k)*}) - f(x_{t, a_t}^{(k)}, \theta^{(k)*}) \right)\mathbb{I}\left( 2 \sqrt{d_2} \vertiii{\Theta_{T_0} - \Theta^{*}}_F\geq   \sum_{k=1}^{d_2} f(x_{t, a_t^*}^{(k)}, \theta^{(k)*}) - f(x_{t, a_t}^{(k)}, \theta^{(k)*}), \mathcal{E}_{T_0} \right) \right] \\
        & \quad + 2d_2 \mathbb{P}(\mathcal{E}_{T_0}^c) \\
        & \leq \mathbb{E}  \left [ \left(\sum_{k=1}^{d_2} f(x_{t, a_t^*}^{(k)}, \theta^{(k)*}) - f(x_{t, a_t}^{(k)}, \theta^{(k)*}) \right) \right. \\
        & \qquad \left. \mathbb{I}\left( 2 \sqrt{d_2}  \vertiii{\Theta_{T_0} - \Theta^{*}}_F\geq   \sum_{k=1}^{d_2} f(x_{t, a_t^*}^{(k)}, \theta^{(k)*}) - f(x_{t, a_t}^{(k)}, \theta^{(k)*})  \geq 2 \sqrt{d_2} v,  \mathcal{E}_{T_0}   \right)\right] +  2d_2 \mathbb{P}(\mathcal{E}_{T_0}^c) +2\sqrt{d_2} v \\
        & \leq\mathbb{E}  \left [ \left(\sum_{k=1}^{d_2} f(x_{t, a_t^*}^{(k)}, \theta^{(k)*}) - f(x_{t, a_t}^{(k)}, \theta^{(k)*}) \right)\mathbb{I}\left( \vertiii{\Theta_{T_0} - \Theta^{*}}_F\geq  v  \right), \mathcal{E}_{T_0} \right] + 2d_2 \mathbb{P}(\mathcal{E}_{T_0}^c) + 2 \sqrt{d_2} v \\ 
        & \leq 2 d_2 \mathbb{P} \left( \vertiii{\Theta_{T_0} - \Theta^{*}}_F  \geq  v , \mathcal{E}_{T_0} \right) + 2d_2 \mathbb{P}(\mathcal{E}_{T_0}^c) + 2\sqrt{d_2} v \\  \\
    \end{aligned} 
\end{equation}

By Lemma \ref{lem: multi_agent_RSC_condition}, the RSC condition is satisfied when $T_0 \geq 2\log(2d_1^2 + d_1)/C(s)$, where $C(s)$ is defined in Lemma \ref{lem: lasso_sigma_distance}.
Now take $v$ to be the upper bound of $\vertiii{\Theta_{T_0} - \Theta^{*}}_F$ in Lemma \ref{lem: oracle_inequality}, i.e., $v^2 = 9 \dfrac{\lambda_{T_0}^2}{\alpha^2} \phi^2 + \dfrac{1}{\alpha}[2Z_{T_0}(\theta^*) + 4\lambda_{T_0} R(\theta^*_{\mathcal{M}^{\bot}})$. We know by Lemma \ref{lem: oracle_inequality}, the expected cumulative regret becomes

\begin{equation}
\nonumber
    \begin{aligned}
       \mathbb{E}[ \text{Regret}(T) ]
        &\leq 2d_2T_0 + \sum_{t=T_0}^T \left[ 2d_2\mathbb{P} \left( \mathcal{A}_{T_0}^c, \mathcal{E}_{T_0} \right) + 2d_2 \mathbb{P}(\mathcal{E}_{T_0}^c) + 2 \sqrt{d_2}v \right] \\
        &\leq 2d_2 T_0 + 2d_2 \sum_{t=T_0}^T \left[  \mathbb{P} \left( \mathcal{A}_{T_0}^c \right) + \mathbb{P}(\mathcal{E}_{T_0}^c)\right] +  2 \sqrt{d_2} \sum_{t=1}^T \sqrt{9 \dfrac{\lambda_{T_0}^2}{\alpha^2} \phi^2 + \dfrac{\lambda_{T_0}}{\alpha}[2Z_{T_0}(\theta^*) + 4 R(\theta^*_{\mathcal{M}^{\bot}})}] \\
    \end{aligned} 
\end{equation}

By setting $\delta = 1/T^2$ in Lemma \ref{lem: multi_agent_lambda_condition} and Lemma \ref{lem: multi_agent_RSC_condition}, the second term can be bounded as 

\begin{equation}
\nonumber
    \begin{aligned}
        2d_2 \sum_{t=T_0}^T \left[  \mathbb{P} \left( \mathcal{A}_t^c \right) + \mathbb{P}(\mathcal{E}_t^c)\right] \leq \sum_{t=1}^T \left[  \frac{1}{T^2} + d_2 \exp(-\frac{T_0C(s)}{2})\right] = \mathcal{O}(1)
    \end{aligned} 
\end{equation}

The last term can be bounded as 

\begin{equation}
\nonumber
    \begin{aligned}
        &2 \sqrt{d_2} \sum_{t=1}^T \sqrt{9 \dfrac{\lambda_{T_0}^2}{\alpha^2} \phi^2 + \dfrac{1}{\alpha}[2Z_{T_0}(\theta^*) + 4 \lambda_{T_0} R(\theta^*_{\mathcal{M}^{\bot}})}]  \\
        & \leq 2 \sqrt{d_2} \sum_{t=1}^T \sqrt{144 \dfrac{\lambda_{T_0}^2}{\alpha_0^2} \phi^2 + \dfrac{8\lambda_t}{\alpha_0}Z_{T_0}(\theta^*)}\\
        &\leq 2 \sqrt{d_2 \frac{144\phi^2}{\alpha_0^2}} \sum_{t=1}^T 2 \left(\sigma \frac{d_2^{1-1/q}}{\sqrt{T_0}} + \sigma \eta(d_2, \frac{1}{2} - \frac{1}{q}) \sqrt{\frac{2 \log 2d_1T^2}{T_0}} \right) \\
        & \leq 2 \eta(d_2, \frac{1}{q} - \frac{1}{2})\sqrt{\frac{144 }{\alpha_0^2} d_2 s  } \sum_{t=1}^T 2 \left(\sigma \frac{d_2^{1-1/q}}{\sqrt{T_0}} + \sigma \eta(d_2, \frac{1}{2} - \frac{1}{q}) \sqrt{\frac{2 \log 2d_1T^2}{T_0}} \right) \\
        & \leq \frac{48}{\alpha_0} \eta(d_2, \frac{1}{q} - \frac{1}{2})\sqrt{d_2 s  }  \frac{T}{\sqrt{T_0}} \left(\sigma {d_2^{1-1/q}} + \sigma \eta(d_2, \frac{1}{2} - \frac{1}{q}) \sqrt{{2 \log 2d_1T^2}} \right) \\
    \end{aligned} 
\end{equation}

Take $q =2 $ and match the two dominating terms, we get the choice of $T_0 = \Theta(d_2  s^{1/3} T^{2/3})$, Therefore the final regret bound is of size

\begin{equation}
\tag*{$\square$}
    \begin{aligned}
        & 2d_2 T_0 + 2d_2 \sum_{t=T_0}^T \left[  \mathbb{P} \left( \mathcal{A}_{T_0}^c \right) + \mathbb{P}(\mathcal{E}_{T_0}^c)\right] +  2 \sqrt{d_2} \sum_{t=1}^T \sqrt{9 \dfrac{\lambda_{T_0}^2}{\alpha^2} \phi^2 + \dfrac{1}{\alpha}[2Z_{T_0}(\theta^*) + 4 \lambda_{T_0} R(\theta^*_{\mathcal{M}^{\bot}})}] \\
        &= \mathcal{O}\left( d_2  s^{1/3} T^{2/3} \sqrt{\log 2d_1T}\right)
    \end{aligned} 
\end{equation}
}

\section{Proofs Related to the Oracle Inequalities}
\label{sec: appendixOracle}

Recall the definition of the constraint set $\mathbb{C}$. We define the subset with bounded norm $\mathbb{K}(\delta) := \mathbb{C}\cap \{\Delta | \Delta \leq \delta \}$. Also define the function  $F_t(\Delta)$ as

\begin{equation}
    \begin{aligned}
    \nonumber
  F_t(\Delta) = L_t(\theta^* + \Delta) -  L_t(\theta^*) + \lambda_t (R(\theta^* + \Delta) - R(\theta^*)) \\ 
    \end{aligned} 
\end{equation}

\subsection{Proof of Lemma \ref{lem: oracle_inequality}}

First, we introduce the following two lemmas from \citet{Negahban2012A}

\begin{lemma}
\label{lem: deviation lemma}
\textbf{\upshape(Lemma 3 of \citet{Negahban2012A}) }For any vectors $\theta^*, \Delta$, and a decomposable norm $R$ on $\mathcal{M}, \overline{\mathcal{M}}^{\bot}$, we have the following inequality
\begin{equation}
     R(\theta^* + \Delta) - R(\theta^*)
              \geq  R(\Delta_{ \overline{\mathcal{M}}^\bot}) - R(\Delta_{\overline{\mathcal{M}}} ) - 2R ( \theta^*_{\mathcal{M^{\bot}}})\\ \nonumber
\end{equation}
\end{lemma}

\begin{lemma}
\label{lem: error_bound_delta}\textbf{\upshape (Lemma 4 of \citet{Negahban2012A})}
If $F_t(\Delta) > 0$ for all vectors $\Delta \in \mathbb{K}(\delta)$ for a constant $\delta$, then $\|\theta_t - \theta^*\|\leq \delta$
\end{lemma}

Now, we provide the proof of Lemma \ref{lem: oracle_inequality}

\textbf{Proof.}  Note that $\delta^2 \leq r_t^2$, by the restricted strong convexity of $L_t(\theta^*)$ on $\mathbb{K}(\delta)$, we have                                                                                           
\begin{equation}
    \begin{aligned}
    \nonumber
  F_t(\Delta) &= L_t(\theta^* + \Delta) -  L_t(\theta^*) + \lambda_t (R(\theta^* + \Delta) - R(\theta^*)) \\ 
              &\geq \langle \nabla L_t(\theta^*), \Delta \rangle + \alpha \| \Delta \|^2 - Z_t(\theta^*) + \lambda_t (R(\theta^* + \Delta) - R(\theta^*)) \\
              &\geq \langle \nabla L_t(\theta^*), \Delta \rangle + \alpha \| \Delta \|^2 - Z_t(\theta^*) + \lambda_t (R(\Delta_{ \overline{\mathcal{M}}^\bot}) - R(\Delta_{\overline{\mathcal{M}}} ) - 2R ( \theta^*_{\mathcal{M^{\bot}}})) \\
              &\geq - |\langle \nabla L_t(\theta^*), \Delta \rangle| + \alpha \| \Delta \|^2 - Z_t(\theta^*) + \lambda_t (R(\Delta_{ \overline{\mathcal{M}}^\bot}) - R(\Delta_{\overline{\mathcal{M}}} ) - 2R ( \theta^*_{\mathcal{M^{\bot}}})) \\
              &\geq - R^*(\nabla L_t(\theta^*)) R(\Delta) + \alpha \| \Delta \|^2 - Z_t(\theta^*) + \lambda_t (R(\Delta_{ \overline{\mathcal{M}}^\bot}) - R(\Delta_{\overline{\mathcal{M}}} ) - 2R ( \theta^*_{\mathcal{M^{\bot}}})) \\
              &\geq \alpha \| \Delta \|^2 - Z_t(\theta^*) + \lambda_t (R(\Delta_{ \overline{\mathcal{M}}^\bot}) - R(\Delta_{\overline{\mathcal{M}}} ) - 2R ( \theta^*_{\mathcal{M^{\bot}}})) - \frac{\lambda_t}{2} R(\Delta)  \\
             &\geq \alpha \| \Delta \|^2 - Z_t(\theta^*) + \lambda_t (R(\Delta_{ \overline{\mathcal{M}}^\bot}) - R(\Delta_{\overline{\mathcal{M}}} ) - 2R ( \theta^*_{\mathcal{M^{\bot}}})) - \frac{\lambda_t}{2} R(\Delta_{\overline{\mathcal{M}}^{\bot}}) - \frac{\lambda_t}{2} R(\Delta_{\overline{\mathcal{M}}})  \\
             &= \alpha \| \Delta \|^2 - Z_t(\theta^*) + \frac{\lambda_t }{2}\left(R(\Delta_{ \overline{\mathcal{M}}^\bot}) - 3R(\Delta_{\overline{\mathcal{M}}} ) - 4R ( \theta^*_{\mathcal{M^{\bot}}})\right) \\
    \end{aligned} 
\end{equation}
where the second inequality is by {Lemma \ref{lem: deviation lemma}}. The fourth inequality is by the generalized Cauchy Schwarz inequality. The fifth inequality is because of our setting $\lambda_t \geq 2 R(\nabla L_t(\theta^*))$. The last inequality is because of the triangle inequality $R(\Delta) = R(\Delta_{\overline{\mathcal{M}}^{\bot}} + \Delta_{\overline{\mathcal{M}}})  \leq R(\Delta_{\overline{\mathcal{M}}^{\bot}}) +  R(\Delta_{\overline{\mathcal{M}}})$. Now by the subspace compatibility constant, we know that 

\begin{equation}
    \begin{aligned}
    \nonumber
    R(\Delta_{\overline{\mathcal{M}}}) \leq \phi \|\Delta_{\overline{\mathcal{M}}}\| = \phi \|\Pi_{\overline{\mathcal{M}}}(\Delta) - \Pi_{\overline{\mathcal{M}}}(0)\| \leq \phi \|\Delta - 0\|
    \end{aligned} 
\end{equation}

where the last inequality is because $0 \in \overline{\mathcal{M}}$ and because the projection operation is non-expansive. Therefore we can continue to lower bound $F_t(\Delta)$ in the following way. 

\begin{equation}
    \begin{aligned}
    \nonumber
  F_t(\Delta)&\geq \alpha \| \Delta \|^2 - Z_t(\theta^*) - \frac{\lambda_t }{2}\left( 3R(\Delta_{\overline{\mathcal{M}}} )+ 4R ( \theta^*_{\mathcal{M^{\bot}}})\right) \\
            &\geq \alpha \| \Delta \|^2 - Z_t(\theta^*) - \frac{3\lambda_t \phi}{2} \| \Delta \| - 2\lambda_t R ( \theta^*_{\mathcal{M^{\bot}}})\\
    \end{aligned} 
\end{equation}

Now since we take $\|\Delta\|^2 = \delta^2 = 9 \frac{\lambda_t^2}{\alpha^2} \phi^2 + \frac{1}{\alpha}[2 Z_t(\theta^*) + 4 \lambda_t R(\theta^*_{\mathcal{M}^{\bot}})]$, by the same algebraic manipulations in \citet{Negahban2012A}, we have $F_t(\Delta) > 0$. Now by {Lemma \ref{lem: error_bound_delta}}, we know that 

\begin{equation}
    \begin{aligned}
    \nonumber\|\theta_t - \theta^*\|^2 \leq 9 \dfrac{\lambda_t^2}{\alpha^2} \phi^2 + \dfrac{1}{\alpha}[2 Z_t(\theta^*) + 4 \lambda_t R(\theta^*_{\mathcal{M}^{\bot}})]    
    \end{aligned} 
\end{equation}

The second oracle inequality in Lemma \ref{lem: oracle_inequality_regularizer_norm} can be proved easily by the triangle inequality decomposition and the definition of $\mathbb{C}$. That is

\begin{equation}
    \begin{aligned}
    \nonumber
  R(\theta_t - \theta^*) &= R((\theta_t - \theta^*)_{\overline{\mathcal{M}}^{\bot}} + (\theta_t - \theta^*)_{\overline{\mathcal{M}}})  \leq R((\theta_t - \theta^*)_{\overline{\mathcal{M}}^{\bot}}) +  R((\theta_t - \theta^*)_{\overline{\mathcal{M}}}) \\
            &\leq 4 R((\theta_t - \theta^*)_{\overline{\mathcal{M}}}) + 4 R(\theta^*_{\mathcal{M}^{\bot}})\\
            &\leq 4 \phi \|(\theta_t - \theta^*)_{\overline{\mathcal{M}}}\| + 4 R(\theta^*_{\mathcal{M}^{\bot}}) \\
            &\leq 4 \phi \|\theta_t - \theta^*\| + 4 R(\theta^*_{\mathcal{M}^{\bot}}) \\
    \end{aligned} 
\end{equation}
If $\theta^* \in \mathcal{M}$, then $R(\theta^*_{\mathcal{M}^{\bot}}) = 0$. Therefore we know that 

\begin{equation}
    \begin{aligned}
    \nonumber
  R(\theta_t - \theta^*)
            &\leq 4\phi \sqrt{9 \dfrac{\lambda_t^2}{\alpha^2} \phi^2 + \dfrac{2}{\alpha} Z_t(\theta^*)} \\
    \end{aligned} 
\end{equation}
\hfill $\square$

\section{Validation Experiments}
\label{sec: appendixExp}

\begin{figure}[ht]
     \centering
          \hspace*{-15pt}%
          \begin{subfigure}[b]{0.4\linewidth}
         \centering
         \includegraphics[width=\textwidth]{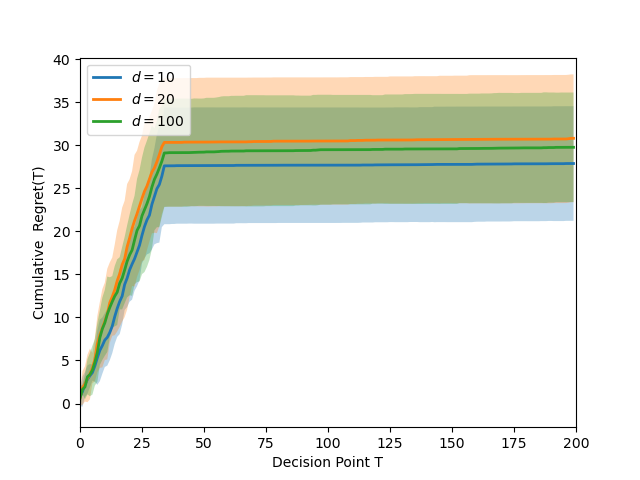}
         \caption{Cumulative $\text{Regret}(T), T=200$}
         \label{fig: lasso_1}
     \end{subfigure}
     \hspace*{-15pt}%
          \begin{subfigure}[b]{0.4\linewidth}
         \centering
         \includegraphics[width=\textwidth]{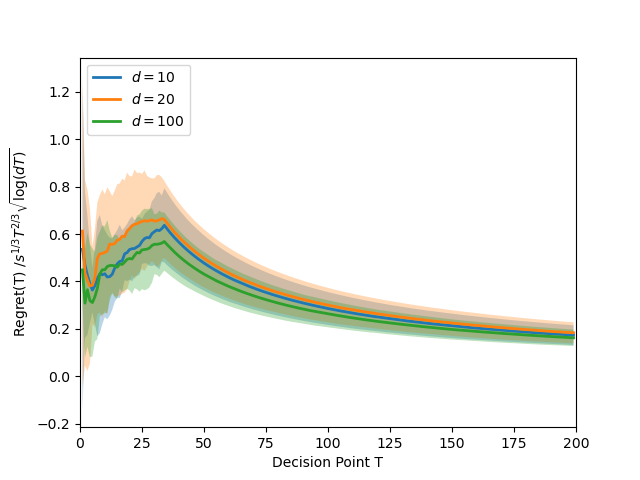}
         \caption{$\text{Regret}(T) / s^{1/3}T^{2/3}\sqrt{\log(dT)}$ }
         \label{fig: lasso_2}
     \end{subfigure}
     \hspace*{-15pt}%
     \begin{subfigure}[b]{0.4\linewidth}
         \centering
         \includegraphics[width=\textwidth]{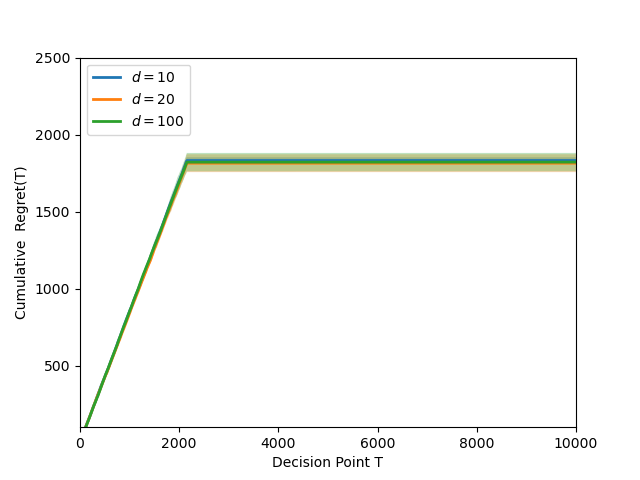}
         \caption{Cumulative $\text{Regret}(T), T=10000 $}
         \label{fig: lasso_3}
     \end{subfigure}
     \hspace*{-15pt}%
     \begin{subfigure}[b]{0.4\textwidth}
         \centering
         \includegraphics[width=\textwidth]{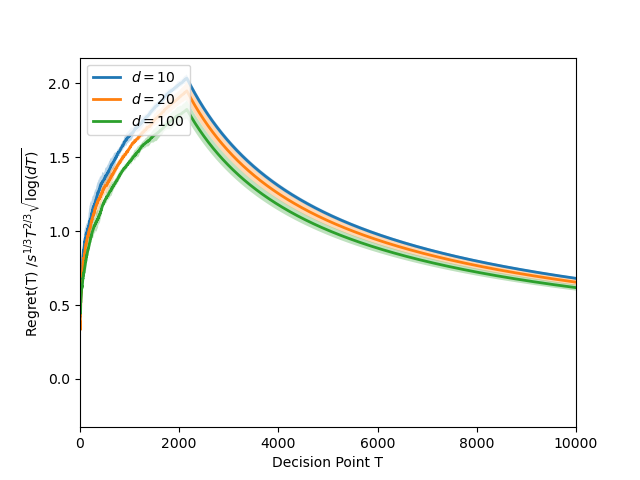}
         \caption{$\text{Regret}(T) / s^{1/3}T^{2/3}\sqrt{\log(dT)}$ }
         \label{fig: lasso_4}
     \end{subfigure}
     \caption{ The experimental results in LASSO Bandit problem when $T=200$ and $T=10000$}
     \label{fig: lasso}
\end{figure}

\begin{figure}[ht]
     \centering
          \hspace*{-15pt}%
        \begin{subfigure}[b]{0.4\linewidth}
         \centering
         \includegraphics[width=\textwidth]{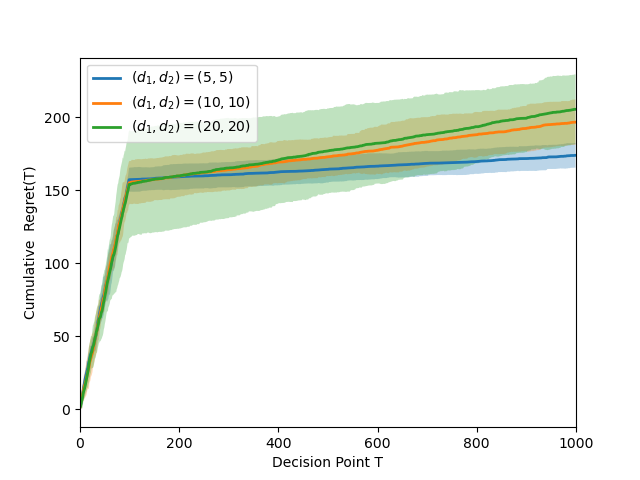}
         \caption{Cumulative $\text{Regret}(T), T = 1000 $}
         \label{fig: low_rank_1}
     \end{subfigure}
     \hspace*{-15pt}%
          \begin{subfigure}[b]{0.4\linewidth}
         \centering
         \includegraphics[width=\textwidth]{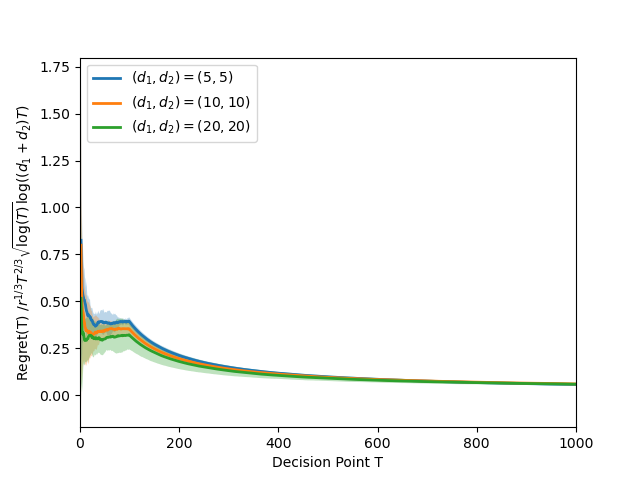}
         \caption{Cumulative $\text{Regret}(T) $}
         \label{fig: low_rank_2}
     \end{subfigure}
     \hspace*{-15pt}%
     \begin{subfigure}[b]{0.4\linewidth}
         \centering
         \includegraphics[width=\textwidth]{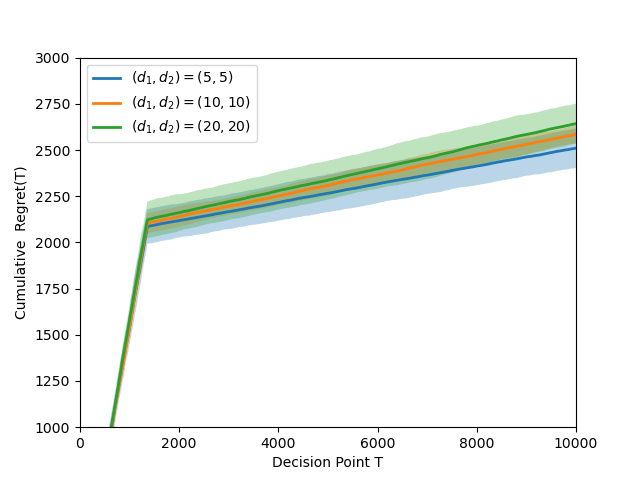}
         \caption{Cumulative $\text{Regret}(T), T = 10000 $}
         \label{fig: low_rank_3}
     \end{subfigure}
     \hspace*{-15pt}%
     \begin{subfigure}[b]{0.4\textwidth}
         \centering
         \includegraphics[width=\textwidth]{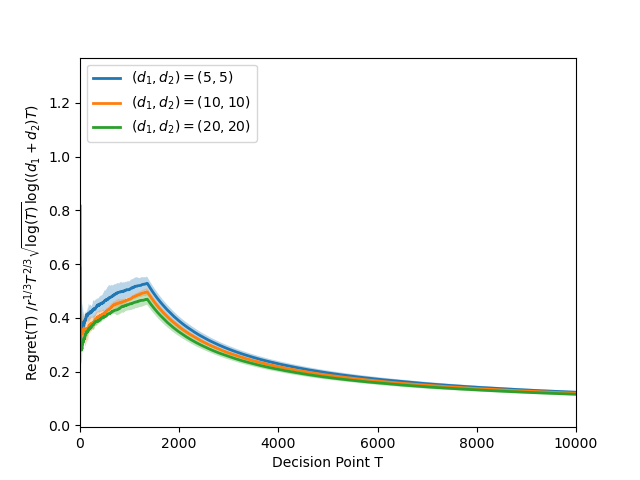}
         \caption{$\text{Regret}(T) / r^{1/3}T^{2/3}\sqrt{\log(T)} \log((d_1 + d_2)T)$ }
         \label{fig: low_rank_4}
     \end{subfigure}
     \caption{ The experimental results in Low-rank Matrix Bandit problem when $T=1000$ and $T=10000$}
     \label{fig: low_rank}
\end{figure}

\begin{figure}[ht]
     \centering
          \hspace*{-15pt}%
        \begin{subfigure}[b]{0.4\linewidth}
         \centering
         \includegraphics[width=\textwidth]{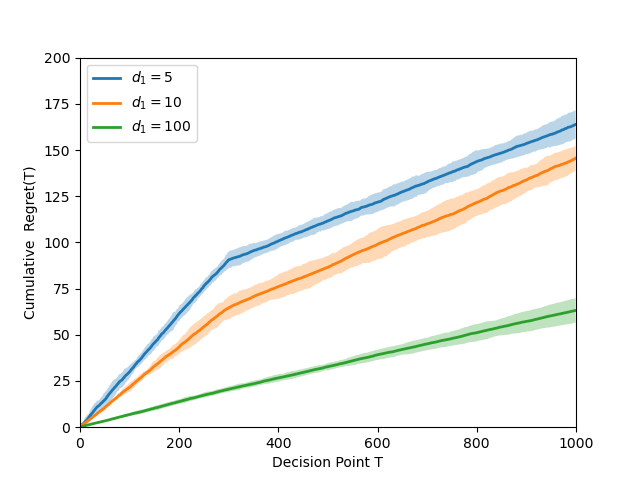}
         \caption{Cumulative $\text{Regret}(T) , T=1000$ }
         \label{fig: group_sparse_1}
     \end{subfigure}
          \hspace*{-15pt}%
     \begin{subfigure}[b]{0.4\textwidth}
         \centering
         \includegraphics[width=\textwidth]{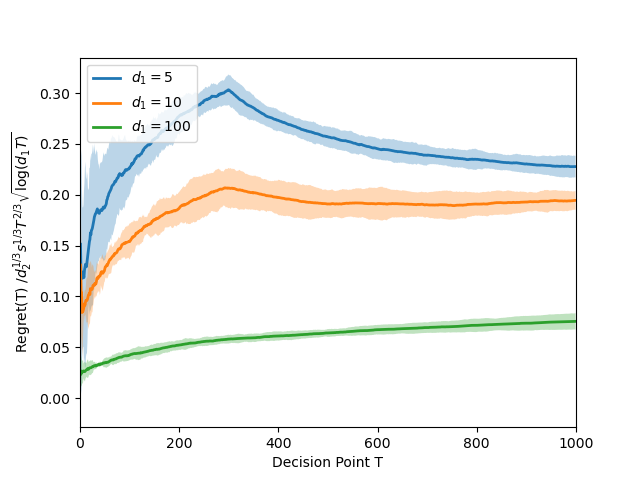}
         \caption{$\text{Regret}(T) / d_2^{1/3}s^{1/3}T^{2/3}\sqrt{ \log(d_1 T)} $ }
         \label{fig: group_sparse_2}
     \end{subfigure}
          \hspace*{-15pt}%
     \begin{subfigure}[b]{0.4\linewidth}
         \centering
         \includegraphics[width=\textwidth]{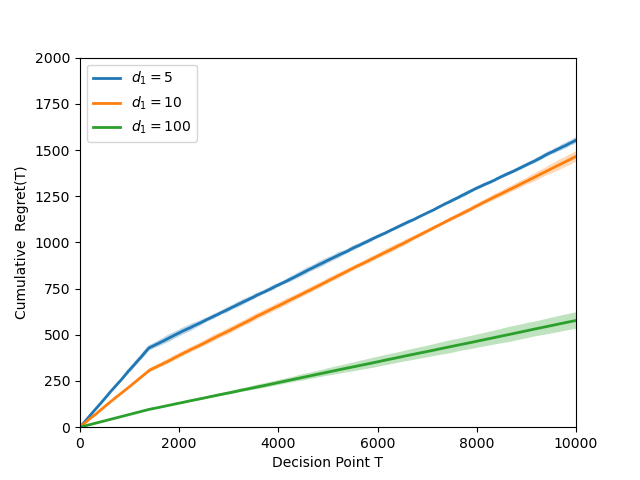}
         \caption{Cumulative $\text{Regret}(T), T=10000 $}
         \label{fig: group_sparse_3}
     \end{subfigure}
         \hspace*{-15pt}%
     \begin{subfigure}[b]{0.4\textwidth}
         \centering
         \includegraphics[width=\textwidth]{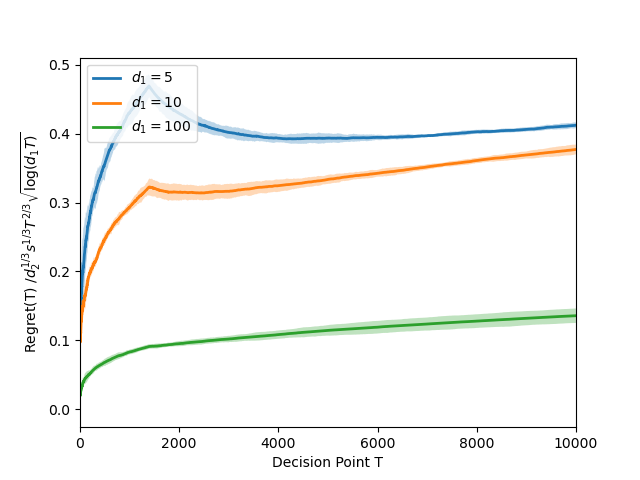}
         \caption{$\text{Regret}(T) / d_2^{1/3}s^{1/3}T^{2/3}\sqrt{ \log(d_1 T)} $ }
         \label{fig: group_sparse_4}
     \end{subfigure}
     \caption{ The experimental results in Group-sparse Matrix Bandit problem}
     \label{fig: group_sparse}
\end{figure}

In this section, we provide some experiments in order to validate our claims in the theoretical results. We choose to run our algorithms in the different high dimensional bandit problems and validate the corresponding regret upper bounds in Corollary \ref{cor: lasso_regret}, Corollary \ref{cor: matrix_regret}, Corollary \ref{cor: group_sparse_regret} and Theorem \ref{thm: regret_bound_multi_agent}.  For each problem, we plot the cumulative regret $R(T)$ as well as $R(T)/B(T)$, where $B(T)$ is the upper bound derived in the specific application. For each setting in each problem, we run the algorithm for 10 independent runs and plot the mean results with one standard deviation error bars.

\begin{itemize}
    \item \textbf{LASSO Bandit}. We generate our true $\theta^*$ by randomly choosing its non-zero indices, and then generate each of its non-zero values uniformly randomly from [0,1] and then perform the normalization. We set $K=10$ so that there are ten different contexts available at each round. The contexts $\{x_{t, a_i}\}_{i=1}^K$ are generated from the zero-mean and identity-covariance normal distribution.  We choose three different dimension sizes $d = 10, 20, 100$. The sparsity level is chosen to be $s = 5$ in all three settings.  The experiment results are shown in Figure \ref{fig: lasso}.  As we can observe, the figures show that the regret is at most a constant times $s^{1/3}T^{2/3}\sqrt{\log(dT)}$, which only depends on the dimension $d$ logarithmically and thus satisfies our requirement. Therefore,  our bound correctly delineate the order of the regret.
    \item \textbf{Low-rank Matrix Bandit}. We first randomly generate $r$ vectors of dimension $d_2$,  and then we generate our true $\Theta^*$ by randomly choosing its rows from the generated $r$ vectors, and thus the matrix is low-rank. We set $K=10$ so that there are ten different contexts available at each round. The contexts $\{X_{t, a_i}\}_{i=1}^K$ are generated from the zero-mean and identity-covariance normal distribution.  We choose three different dimension sizes $(d_1, d_2) = (5, 5), (10, 10), (20, 20)$. The rank is chosen to be $r = 2$ in all three settings.  The experiment results are shown in Figure \ref{fig: low_rank}.  As we can observe, the figures show that the regret is at most a constant times $r^{1/3}T^{2/3}\sqrt{\log(T)} \log((d_1 + d_2)T)$, which only depends on the dimensions $d_1 + d_2$ logarithmically and thus satisfies our requirement. Therefore,  our bound correctly delineate the order of the regret.
    
    \item \textbf{Group-sparse Matrix Bandit}. We first randomly generate $s$ row indices from the set $[d_2]$, where $d_2 = 5$ and then we generate our true $\Theta^*$ by setting the selected rows in $\Theta^*$ to be non-zero and generated uniform randomly from $(0,1)$. The other rows are set to be zero. We then perform the normalization. We set $K=10$ so that there are ten different contexts available at each round. The contexts $\{X_{t, a_i}\}_{i=1}^K$ are generated from the zero-mean and identity-covariance normal distribution. We choose three different row dimensions $d_1 = 5, 10, 100$. The group-sparsity is chosen to be $s = 3$ in all three settings. The experiment results are shown in Figure \ref{fig: group_sparse}.  As we can observe, the figures show that the regret is at most a constant times $d_2^{1/3}s^{1/3}T^{2/3}\sqrt{ \log(d_1 T)}$, which only depends on the dimension $d_1$ logarithmically and thus satisfies our requirement. Therefore, our bound correctly delineate the order of the regret.
\end{itemize}

In conclusions, all the experimental results validate all our claims in Section \ref{sec: applications}.

\end{document}